\documentclass[11pt]{article}

\usepackage[final]{acl}

\usepackage{times}
\usepackage{latexsym}

\usepackage[T1]{fontenc}

\usepackage[utf8]{inputenc}

\usepackage{microtype}

\usepackage{inconsolata}

\usepackage{graphicx} 
\usepackage{algorithm}
\usepackage{algpseudocode}
\usepackage{amsmath,amssymb,amsfonts}
\usepackage{tikz}
\usepackage{subcaption}
\usepackage{framed}
\usepackage{textcomp}
\usepackage{tabularx}
\usepackage{listings}
\usepackage{array}
\usepackage{adjustbox}
\usepackage{float}
\usepackage{makecell}
\usepackage{multirow}
\usepackage{xltabular}
\usepackage{siunitx}
\usepackage{ragged2e}
\usepackage{wrapfig}
\usepackage{hyperref}
\usepackage{booktabs} 
\usepackage{svg}
\usepackage{cuted}
\usepackage{balance}
\usepackage{lipsum}
\usepackage[most]{tcolorbox}
\usepackage{xcolor}

\newcommand{\Input}{\Require}
\newcommand{\Output}{\Ensure}

\tcbset{
  compactprompt/.style={
    enhanced,
    colback=white,
    colframe=black,
    boxrule=0.6pt,
    arc=1mm,
    outer arc=1mm,
    left=3pt,
    right=3pt,
    top=2pt,
    bottom=2pt,
    before skip=2pt,
    after skip=4pt
  }
}

\algrenewcommand\algorithmicindent{1em}
\newcommand{\AlgState}[1]{%
  \State
  \parbox[t]{\dimexpr\linewidth-0.8em\relax}{%
    \raggedright #1%
  }%
}

\newcommand{\qset}[1]{\mathbf{q}.{#1}}

\newcommand{\clamp}{\operatorname{clamp}}
\newcommand{\oplow}{\operatorname{lower}}        
\newcommand{\age}{\operatorname{age}}
\newcommand{\topk}{\operatorname{TopK}}
\newcommand{\jac}{J}                             
\newcommand{\mpipat}{\texttt{MPI\_\textbackslash w+}}

\def\BibTeX{{\rm B\kern-.05em{\sc i\kern-.025em b}\kern-.08em
    T\kern-.1667em\lower.7ex\hbox{E}\kern-.125emX}}

\algrenewcomment[1]{\hfill$\triangleright$ \textit{\small #1}}

\hypersetup{
  pdftitle={MetaFORGE: A Meta-Learning Framework for Harness Optimization in Automated Code Generation for Parallel Computing},
  pdfauthor={Scott Piersall, Yang Gao, Shenyang Liu, Liqiang Wang}
}

\usetikzlibrary{arrows.meta, positioning, shapes.geometric, fit}

\title{MetaFORGE: A Meta-Learning Framework for Harness Optimization in Automated Code Generation for Parallel Computing}

\author{
Scott Piersall,
Yang Gao,
Shenyang Liu,
Liqiang Wang \\
Department of Computer Science, University of Central Florida \\
Orlando, FL, USA \\
\texttt{\{spiersall,Yang.Gao,Shenyang.Liu,Liqiang.Wang\}@ucf.edu}
}

\begin{document}

\maketitle

\begin{abstract}
Large language models (LLMs) can generate parallel programs, but their
correctness and performance depend strongly on the instructions that guide
generation. We present \textsc{MetaFORGE}, an execution-grounded meta-learning
framework that iteratively optimizes persistent natural-language skill
specifications for MPI C code generation. \textsc{MetaFORGE} combines
correctness testing, compiler diagnostics, static-analysis findings, runtime
measurements, and rank-level load-imbalance profiles into a scalar reward that
drives textual-gradient refinement, while an external memory distills and
retrieves reward-supported optimization knowledge across runs.

Across six MPI tasks and three workload sizes, \textsc{MetaFORGE} improves
static-analysis quality, runtime, and load balance relative to zero-shot
generation and generally exceeds the evaluated \textsc{TextGrad} iterative
baseline. A model-matched refinement ablation shows that the MetaFORGE
advantage persists when its refinement model is reduced to \texttt{gpt-4o}.
An expanded retrieval ablation across all 18 task--workload configurations
shows consistent runtime benefit and broad static-analysis and load-balance
benefit from retrieved context. Leakage-controlled held-out and zero-update
experiments further provide evidence that optimization knowledge persists in
both the transferred skill and external memory under the evaluated protocol.
\end{abstract}

\begin{figure*}[!t]
    \centering
    \includegraphics[width=0.9\linewidth]{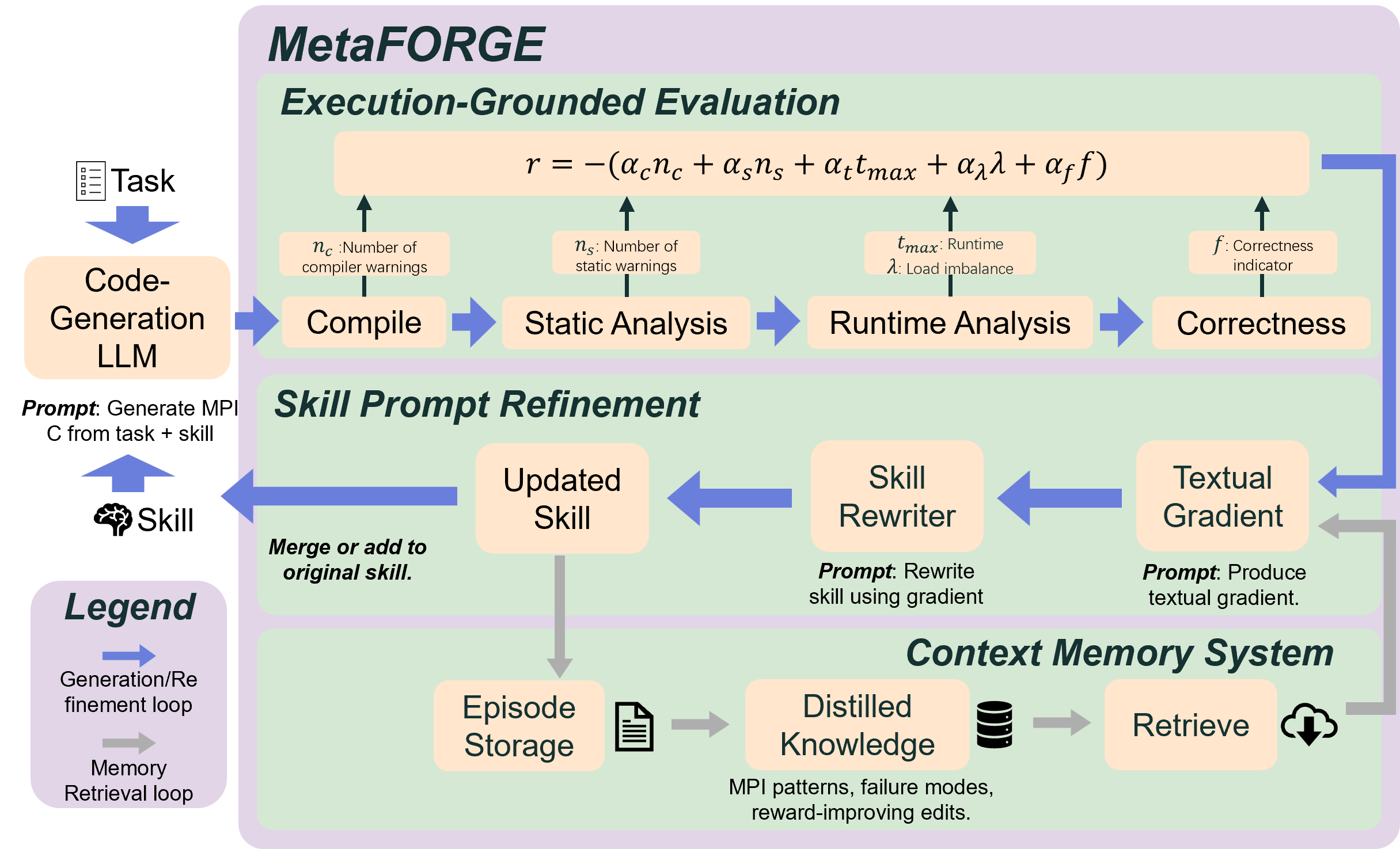}
    \caption{\textbf{MetaFORGE: An automated meta-learning skill-refinement architecture.}
    MetaFORGE optimizes LLM coding skills through execution-grounded feedback.
    Generated MPI C code is compiled, statically analyzed, executed, and scored using
    Eq.~\eqref{eq:reward}. Diagnostics and reward trends are converted into a textual
    gradient for skill-prompt rewriting, while iteration artifacts are stored,
    distilled, indexed, and retrieved as long-term memory, enabling future
    updates to reuse prior reward-improving strategies.}  
    \label{fig:frame}
\end{figure*}

\section{Introduction}
LLM-based code generation remains difficult because informal or semi-formal requirements often contain ambiguities in intent, assumptions, constraints, performance expectations, and correctness criteria. Requirements engineering has long recognized that software requirements are frequently incomplete or open to multiple interpretations, and that translating them into working systems typically requires trained professionals and iterative processes of analysis, design, implementation, and validation~\cite{zadenoori2025LLMRequirementsEngineering}. Although code-trained LLMs have demonstrated strong synthesis capabilities~\cite{chen2021evaluating}, recent work shows that ambiguous or incompletely specified requirements can substantially degrade generated code quality, and that models often fail to detect or resolve such ambiguity without external guidance~\cite{yang2026orchid,fang2026clarifycodebench}.

These challenges are especially amplified in high-performance parallel programming. MPI programs coordinate distributed processes through explicit communication, so correctness and performance depend on how computation, data, synchronization, and communication are distributed across processes. Errors in decomposition, message ordering, collectives, boundary exchange, or memory layout can cause incorrect results and deadlocks~\cite{Li_2019_Efficient_Concolic,Chen_2020_MPI_SV,Hilbrich_2012_MUST_DeadlockDetection,HUang_2020_Predictive_Deadlock,Droste_2015_MPI_CHECKER_STATIC,Li_2018_COMPI_Concolic,Yu_Hengbiao_Sym_Model_MPI_2018}. Many performance issues are also execution-dependent: code may compile and run correctly yet still suffer from communication overhead, root bottlenecks, inefficient decomposition, or rank-level imbalance that emerges only at runtime and scale~\cite{adhianto2010hpctoolkit,knupfer2012score}. Thus, MPI code generation is better viewed as an execution-grounded optimization problem than as one-shot synthesis.

Iterative refinement approaches used by LLMs primarily optimize for functional correctness or unit-test success and rarely incorporate compiler diagnostics, static analysis, runtime profiling, or utilization metrics into a unified optimization objective. Furthermore, they typically discard optimization experience after each task rather than accumulating reusable domain knowledge across tasks.

We present \textsc{MetaFORGE}, an execution-grounded meta-learning
framework for MPI C code generation. Rather than optimizing only an
individual generated program, \textsc{MetaFORGE} treats the persistent
\emph{skill specification}---the natural-language instructions that
condition generation---as an adaptive policy refined within and across
tasks. Each candidate program is compiled, tested, statically analyzed,
executed, and profiled; the resulting evidence is combined into a reward
that penalizes correctness failures, compiler warnings, static-analysis
findings, externally measured whole-job runtime, and load imbalance. \textsc{MetaFORGE}
then converts the diagnostics and reward trends into a \emph{textual
gradient}: a natural-language update signal that identifies measured
deficiencies and proposes revisions to the skill. A second model call
applies the gradient while preserving unrelated instructions.
Figure~\ref{fig:frame} summarizes this closed-loop process.

Beyond within-task skill refinement, \textsc{MetaFORGE} integrates
automated execution, MPI-aware diagnostics, per-rank runtime profiling,
deterministic skill versioning, and persistent external memory. The
memory system stores iteration-level evidence and distills
reward-improving episodes into compact guidance indexed by MPI
constructs, parallel patterns, and observed correctness or performance
symptoms. The skill captures broadly applicable generation behavior,
whereas memory selectively retrieves more specific optimization
experience from prior runs. Because both artifacts are external to the
model, the framework supports inspectable cross-task transfer without
parameter fine-tuning. MPI therefore serves as a demanding testbed for
execution-grounded skill refinement and reusable optimization knowledge
in language-model agents.

Our central hypothesis is that execution-grounded refinement of
persistent coding skills can improve program correctness,
static-analysis quality, runtime performance, and rank-level load
balance beyond zero-shot generation and general-purpose
textual-gradient refinement. This work makes three contributions.
\textbf{(1)} We introduce \textsc{MetaFORGE}, an MPI-aware
meta-learning harness that optimizes persistent LLM skill
specifications using correctness tests, compiler diagnostics,
static-analysis findings, externally measured whole-job runtime, and rank-level
load-imbalance measurements. Its two-stage refinement procedure
converts this execution evidence into an explicit textual gradient
and then applies the gradient while preserving unrelated skill
instructions.
\textbf{(2)} We develop a persistent cross-task memory mechanism
that separates detailed iteration episodes from compact,
reward-supported optimization lessons. The mechanism distills,
indexes, and retrieves prior lessons using MPI constructs, parallel
patterns, task characteristics, and observed correctness or
performance symptoms.
\textbf{(3)} We evaluate \textsc{MetaFORGE} across six MPI tasks,
three workload sizes, and five independent trials per configuration
against zero-shot generation and a \textsc{TextGrad}-based iterative
baseline. We additionally conduct retrieval, cross-model,
refinement-model, and leave-one-task-out studies to assess memory
contribution, model sensitivity, and cross-task persistence.

\section{Related Work}

MetaFORGE builds upon prior research in execution-guided code optimization, memory-augmented LLM agents, and HPC performance engineering. This section briefly reviews these areas and highlights the gaps addressed by our framework.

\subsection{Execution-Grounded Code Optimization}
Recent language-model agents have demonstrated the effectiveness of iterative refinement using external feedback. Reflexion~\cite{shinn2023reflexion}, SWE-Agent~\cite{yang2024swe}, and AutoCodeRover~\cite{zhang2024autocoderover} leverage execution results, tool use, and program analysis to improve generated code, while optimization-oriented systems such as Language Agents as Optimizable Graphs~\cite{zhuge2024language} and AlphaEvolve~\cite{novikov2025alphaevolve} further demonstrate the value of performance-driven refinement. TextGrad~\cite{yuksekgonul2024textgradautomaticdifferentiationtext} and related work~\cite{lee2026metaharnessendtoendoptimizationmodel,yuksekgonul2025optimizing} show that textual feedback can optimize agent behavior without modifying model parameters. However, these approaches primarily target general software-engineering tasks and do not focus on distributed-memory performance optimization. This paper~\cite{wei2024improving} incorporate execution feedback into parallel-program optimization, but optimization experiences remain task-specific rather than reusable across tasks.

\subsection{Memory-Augmented Agents}

External memory and retrieval mechanisms enable language-model agents to accumulate and reuse knowledge across tasks. MemoryBank~\cite{zhong2024memorybank}, MemGPT~\cite{packer2023memgpt}, Generative Agents~\cite{park2023generative}, MemInsight~\cite{salama2025meminsight}, and Mem0~\cite{chhikara2025mem0} demonstrate the benefits of persistent memory for long-horizon reasoning. Retrieval-augmented generation (RAG)~\cite{lewis2020retrieval} and software-engineering systems such as RepoCoder~\cite{zhang2023repocoder} further show that incorporating external knowledge can improve generation quality. Existing approaches primarily retrieve conversational history, documentation, or source code, whereas our work focuses on storing and retrieving optimization knowledge derived from execution feedback, profiling results, and compiler diagnostics.

\subsection{Performance Analysis and Optimization in HPC}
Performance engineering has long relied on execution-grounded analysis to identify bottlenecks in distributed-memory applications. Profiling frameworks such as HPCToolkit~\cite{adhianto2010hpctoolkit} and Score-P~\cite{knupfer2012score} provide detailed measurements of communication overhead, synchronization behavior, and load imbalance, while correctness-analysis tools such as MUST~\cite{Hilbrich_2012_MUST_DeadlockDetection},  PARCOACH~\cite{PARCOACH2020}, and LLM-based analysis  ~\cite{piersall2026improving} assist developers in detecting communication and synchronization defects. Although these tools effectively diagnose performance and correctness issues, the resulting optimization experiences are rarely consolidated into reusable knowledge for future optimization tasks. ParaCodex \cite{kaplan-etal-2026-paracodex} uses a Codex-based autonomous workflow that integrates correctness checking, profiling, and refinement for translating serial CPU kernels to OpenMP GPU-offload kernels and also evaluates CUDA-to-OpenMP offload translation; it is limited to GPU-offload execution scopes and does not reuse previously learned optimization knowledge. 

\section{MetaFORGE Harness Design}

Figure~\ref{fig:frame} presents \textsc{MetaFORGE} as two coupled optimization
loops. The within-run loop repeatedly generates, evaluates, and refines a
natural-language skill for the current task. The cross-run loop converts the
resulting trajectory into persistent optimization knowledge that can guide
later tasks. The framework maintains a current skill $\mathcal{S}$, a
best-observed skill $\mathcal{S}^*$, an episode store $\mathcal{E}$ containing
iteration-level evidence, a distilled memory store $\mathcal{M}$, and an
inverted index $\mathcal{I}$ used for retrieval. This separation between
transient episodes and persistent memories allows the harness to preserve
detailed execution evidence during refinement while exposing only concise,
reward-supported guidance to subsequent model calls.

The central design principle is that execution evidence should modify the
persistent instructions that condition future generation rather than only
repairing the current program. Each candidate is therefore evaluated using
correctness, compiler, static-analysis, runtime, and rank-level utilization
signals. These signals guide a two-stage textual-gradient update, after which
successful optimization guidance is distilled and indexed for reuse. The
following subsections describe the evaluator, skill-refinement process, and
cross-run memory mechanism.

\subsection{Framework State and Control Flow}

MetaFORGE iteratively optimizes a \emph{skill specification}, a structured
natural-language prompt that guides an LLM-based MPI C code generator.
Algorithm~\ref{alg:harness_loop_inline} summarizes the closed-loop refinement
process. 
Given a task $\mathcal{T}$ and an initial skill $\mathcal{S}_0$, each
iteration generates and evaluates a candidate MPI C program, extracts an
MPI-relevant tuple, retrieves matching optimization memories, and
updates the skill using execution-grounded textual feedback augmented with
the retrieved context. The best
skill observed so far is retained, and refinement stops after $P$ consecutive iterations without reward improvement.

\subsection{Execution-Grounded Evaluation}
\label{subsection:execution_grounded_evaluation}

Let $\mathcal{S}$ denote the skill specification, $\mathcal{T}$ the fixed task
description, and $K$ the number of MPI processes. Let $t_{\mathrm{job}}$
denote the externally measured whole-job runtime used in the reward, and let
$\boldsymbol{\tau}=(\tau_0,\ldots,\tau_{K-1})$ denote the per-rank elapsed
times used to compute load imbalance. \textsc{MetaFORGE} defines the scalar
reward as
\begin{equation}
r = -\bigl(\alpha_c n_c+\alpha_s n_s+\alpha_t t_{\mathrm{job}}
+\alpha_\lambda \lambda+\alpha_f f\bigr).
\label{eq:reward}
\end{equation}
Here, $n_c$ and $n_s$ are compiler-warning and static-analysis-warning counts,
and $\bar{\tau}=\frac{1}{K}\sum_{k=0}^{K-1}\tau_k$ is the mean per-rank elapsed
time. The normalized load-imbalance coefficient is
\[
\lambda=
\frac{\max_k\tau_k-\min_k\tau_k}{\bar{\tau}}.
\]
The binary
variable $f$ indicates correctness failure, with $f=1$ if the generated program
fails task-specific correctness testing and $f=0$ otherwise. The coefficient
$\alpha_f$ is set large enough that incorrect programs receive a high negative
reward and are not selected merely because they reduce warnings, runtime, or
imbalance. The remaining coefficients
$\alpha_c,\alpha_s,\alpha_t,\alpha_\lambda>0$ control the relative importance
of code quality, runtime, and utilization. Because $r\leq0$, values closer to
zero indicate better candidates.

For each candidate, the harness executed the compiled program 20 times
sequentially without recompilation and retained all repetitions; no initial run
was discarded as a warm-up. For execution $j$, the harness measured the elapsed
time of the complete \texttt{mpirun -n 4 <binary>} invocation externally. The
candidate runtime used for reward computation and reported runtime improvement is
the arithmetic mean $t_{\mathrm{job}}=
\frac{1}{20}\sum_{j=1}^{20}t_{\mathrm{mpirun}}^{(j)}$.

Rank-level measurements $\tau_k$ were collected separately for load-imbalance
evaluation. All timing was collected by the harness rather than self-reported by
the generated program, preventing generated code from directly manipulating the
runtime signal used for evaluation.

The reward serves as a within-task candidate-selection signal rather than as a
universal task-normalized measure of program quality. Its terms capture
complementary failure modes. Correctness testing prevents invalid programs from
being preferred because they execute quickly; compiler and static-analysis
findings expose defects that may not appear in task outputs; whole-job runtime
measures end-to-end execution cost; and $\lambda$ distinguishes poor rank
utilization from uniformly slow execution. Absolute raw reward values are not
interpreted as directly comparable across tasks because runtime and
warning-count scales differ.

\subsection{Reward-Guided Skill Refinement}
\label{subsection:reward_guided_refinement}

\begin{algorithm}[t]
\caption{MetaFORGE Skill Meta-Learning Harness}
\label{alg:harness_loop_inline}
\footnotesize
\begin{algorithmic}[1]

\Input Task $\mathcal{T}$, initial skill $\mathcal{S}_0$, limits $(N,P)$,
process count $K$, retrieval budget $k$, stores
$(\mathcal{E},\mathcal{M},\mathcal{I})$, weights
$\boldsymbol{\alpha}$, and rewriter $\rho$

\Output Optimized skill $\mathcal{S}^*$

\AlgState{Initialize $\mathcal{S}\gets\mathcal{S}_0$,
$\mathcal{S}^*\gets\mathcal{S}_0$, $r^*\gets-\infty$, $p\gets0$,
$\nabla_{\mathrm{prev}}\gets\varepsilon$, and
$\mathbf{ids}\gets[\,]$.}

\For{$i=1$ \textbf{to} $N$}

  \State $C_i\gets\textsc{Generate}(\mathcal{S},\mathcal{T})$
  \Comment{Prompt~\ref{prompt_code_generation}}


\AlgState{$(\mathbf{m}_i,\ell_i,r_i)\gets
\textsc{Evaluate}(C_i,K,\boldsymbol{\alpha})$, which compiles, tests,
analyzes, executes, and scores $C_i$.}

  \AlgState{If $r_i>r^*$, set
  $(\mathcal{S}^*,r^*,p)\gets(\mathcal{S},r_i,0)$;
  otherwise set $p\gets p+1$.}

  \AlgState{$\mathbf{q}_i\gets
  \textsc{ExtractTuples}(\mathcal{T},C_i,\mathbf{m}_i,
  \nabla_{\mathrm{prev}},\ell_i)$.
  \textit{(Appendix Alg.~\ref{alg:extract})}}

  \AlgState{$\mathrm{id}_i\gets
  \mathcal{E}.\mathrm{store}
  (\mathcal{S},\mathcal{T},C_i,\mathbf{q}_i,
  \mathbf{m}_i,r_i,r^*)$;
  append $\mathrm{id}_i$ to $\mathbf{ids}$.}

  \State \textbf{break if} $p\geq P$

  \AlgState{$\boldsymbol{\xi}_i\gets
  \textsc{FormatContextBlock}
  \bigl(\textsc{RetrieveContext}(\mathbf{q}_i,k)\bigr)$.
  \textit{(Appendix Alg.~\ref{alg:retrieve_inline})}}

  \AlgState{$\mathcal{X}_i\gets
  [\mathcal{S},\mathcal{T},C_i,\mathbf{m}_i,
  \ell_i,r_i,r^*,\boldsymbol{\xi}_i]$.}

  \State $(\mathcal{S}',\nabla_i)\gets
  \textsc{RewriteViaGradient}(\mathcal{X}_i,\rho)$
  \Comment{Alg.~\ref{alg:rewrite_via_gradient_inline}}

  \AlgState{$\mathcal{E}.\mathrm{updateGradient}
  (\mathrm{id}_i,\nabla_i)$;
  $\mathcal{S}\gets\mathcal{S}'$;
  $\nabla_{\mathrm{prev}}\gets\nabla_i$.}

\EndFor

\AlgState{\textsc{CompressandIndex(}\textbf{ids}\textsc{)} \Comment{Section ~\ref{subsection:long_term_context_memory_system} and Algorithm~\ref{alg:compress_and_Index}}}

\State \Return $\mathcal{S}^*$

\end{algorithmic}
\end{algorithm}

Algorithm~\ref{alg:harness_loop_inline} organizes the \textsc{MetaFORGE}
refinement process into four stages while maintaining the current skill
$\mathcal{S}$ and the best-performing skill $\mathcal{S}^*$. Lines~1--2
initialize $\mathcal{S}$, $\mathcal{S}^*$, the best observed reward $r^*$,
the patience counter $p$, the previous textual gradient
$\nabla_{\mathrm{prev}}$, and the episode-identifier list $\mathbf{ids}$
before entering the bounded refinement loop. The counter $p$ records
consecutive non-improving iterations, while $\mathbf{ids}$ retains the
episodes required for subsequent retrieval and trajectory compression.
Lines~3--5 call \textsc{Generate} to produce MPI C candidate $C_i$ from the
current skill and \textsc{Evaluate} to obtain the metric vector
$\mathbf{m}_i$, diagnostic feedback $\ell_i$, and scalar reward $r_i$; the
algorithm then updates $\mathcal{S}^*$, $r^*$, and $p$ according to whether
$r_i$ improves the best reward observed so far, thereby grounding candidate
selection in measured correctness, code quality, runtime, and load balance
rather than model self-assessment. Lines~6--10 construct the
memory-augmented rewrite context: \textsc{ExtractTuples} produces the tuple
$\mathbf{q}_i$ from the current task, candidate, metrics, diagnostics, and
previous gradient; $\mathcal{E}.\textsc{store}$ preserves the complete
episode-level evidence; and \textsc{RetrieveContext} and
\textsc{FormatContextBlock} retrieve and format relevant prior optimization
experience as context $\boldsymbol{\xi}_i$. Line~8 terminates the loop when
the patience threshold is reached, avoiding an unnecessary rewrite, whereas
Line~10 combines the current execution evidence and retrieved memory into the
complete rewrite context $\mathcal{X}_i$. Lines~11--12 call
\textsc{RewriteViaGradient}, through rewriter $\rho$, to diagnose the
candidate's shortcomings and produce an updated skill $\mathcal{S}'$ together
with textual gradient $\nabla_i$; the algorithm assigns
$\mathcal{S}\gets\mathcal{S}'$, attaches $\nabla_i$ to the stored episode,
and retains it as the rationale for the update and as
$\nabla_{\mathrm{prev}}$ for the next iteration. Finally, Lines~14--15
compress and index the completed trajectory for future retrieval and return
the best-performing skill $\mathcal{S}^*$ observed during the run.

Algorithm~\ref{alg:rewrite_via_gradient_inline} converts
this context into a textual gradient and applies the gradient to rewrite the
skill; if no actionable gradient is produced, the current skill is retained.
The returned gradient is stored with the episode for later lesson
distillation, and after refinement terminates, the accumulated episodes are
compressed and indexed for use in subsequent runs. The scalar reward indicates whether the current candidate improved, but it does not specify how the natural-language skill should change.
Algorithm~\ref{alg:rewrite_via_gradient_inline} addresses this limitation by
separating evidence-based diagnosis from skill modification.

\begin{algorithm}[t]
\caption{Skill Rewriting via Textual Gradient: \textsc{RewriteViaGradient} }
\label{alg:rewrite_via_gradient_inline}
\footnotesize
\begin{algorithmic}[1]

\Require Refinement context
$\mathcal{X}=
[\mathcal{S},\mathcal{T},C,\mathbf{m},
\ell,r_i,r^*,\boldsymbol{\xi}]$
and rewriter $\rho$

\Ensure Updated skill $\mathcal{S}'$ and textual gradient
$\nabla_{\mathrm{text}}$

\AlgState{Unpack $\mathcal{X}$ into the current skill $\mathcal{S}$,
task $\mathcal{T}$, generated code $C$, metric summary $\mathbf{m}$,
diagnostics $\ell$, current and best rewards $(r_i,r^*)$, and
retrieved context $\boldsymbol{\xi}$.}

\Statex \textbf{Step 1: Compute textual gradient}

\State $\nabla_{\mathrm{text}}\gets
\textsc{TextGradient}(\mathcal{X})$
\Comment{Prompt~\ref{prompt_text_gradient_inline}}

\State \textbf{return}
$(\mathcal{S},\nabla_{\mathrm{text}})$
\textbf{if} $\nabla_{\mathrm{text}}=\varepsilon$

\Statex \textbf{Step 2: Apply textual gradient}

\State $\mathcal{S}'\gets
\textsc{RewriteSkill}
(\mathcal{S},\nabla_{\mathrm{text}},\boldsymbol{\xi};\rho)$
\Comment{Prompt~\ref{skillrewrite}}

\State \Return $(\mathcal{S}',\nabla_{\mathrm{text}})$

\end{algorithmic}
\end{algorithm}

Algorithm~\ref{alg:rewrite_via_gradient_inline} implements the skill update
in two intentional stages. Line~1 unpacks the refinement context
$\mathcal{X}$ so that the current skill, task, generated program, diagnostic
evidence, metric summary, reward comparison, and retrieved memories remain
available to the update procedure. In the diagnostic stage,
\textsc{TextGradient} uses this evidence in Line~2 to identify which skill
instructions should change and why. Prompt~\ref{prompt_text_gradient_inline} in Appendix
directs the model to restore correctness when testing fails and otherwise
prioritize the largest weighted reward terms. Line~3 provides a conservative
stopping condition: if \textsc{TextGradient} produces no actionable guidance,
the current skill is returned unchanged. In the modification stage,
\textsc{RewriteSkill} applies the textual gradient and retrieved context in
Line~4 using Prompt~\ref{skillrewrite} in Appendix. Separating diagnosis from rewriting
makes the proposed change explicit and inspectable before it modifies the
persistent skill. Finally, Line~5 returns both the revised skill and the
textual gradient; retaining the gradient preserves the rationale for the
update so that it can later be attached to the episode and distilled into
cross-run memory.

The separation between gradient generation and skill rewriting is deliberate.
The first model call performs diagnosis: it compares the current and
best-observed metrics, identifies the largest weighted reward terms, and
proposes evidence-supported changes. The second call edits the persistent
skill using that critique while preserving unrelated instructions. This
division prevents diagnostic reasoning from being conflated with unrestricted
prompt rewriting and creates an explicit textual artifact that can later be
stored and distilled. The refinement context also includes retrieved
optimization memories, but those memories remain advisory: the current
execution evidence and reward comparison determine whether a retrieved
strategy is applicable. The complete textual-gradient and skill-rewrite
templates are provided in Prompts~\ref{prompt_text_gradient_inline}
and~\ref{skillrewrite} in Appendix \ref{appendix:prompts}.

\subsection{Retrieval-Augmented Optimization Memory}
\label{subsection:long_term_context_memory_system}

\textsc{MetaFORGE} augments the within-run refinement loop with persistent
cross-task memory. In Algorithm~\ref{alg:harness_loop_inline}, the
\textsc{ExtractTuples} step summarizes the current task, generated program,
evaluation metrics, diagnostics, and preceding textual gradient using tuples
for MPI constructs, parallel patterns, task characteristics, and observed
performance or correctness symptoms. The harness stores these tuples together
with the skill, generated code, measurements, reward, and refinement evidence
as an iteration-level episode.

Before the next skill update, \textsc{RetrieveContext} uses the current tuple
bundle to identify related, reward-supported lessons from prior runs
(see Appendix Algorithm~\ref{alg:retrieve_inline}), and
\textsc{FormatContextBlock} converts the selected lessons into a compact
context block. Algorithm~\ref{alg:rewrite_via_gradient_inline} combines this
retrieved context with the current execution evidence when producing and
applying the textual gradient. Retrieved guidance is therefore advisory:
current correctness results, diagnostics, runtime measurements, and reward
trends determine whether a previously learned strategy is appropriate.

After the refinement loop terminates, the final step of
Algorithm~\ref{alg:harness_loop_inline} compresses the accumulated trajectory
into a persistent memory record (see Appendix Algorithm~\ref{alg:compress_and_Index} for details). Reward-improving episodes are prioritized,
their textual gradients are distilled into concise, action-oriented guidance,
and the resulting lesson is indexed by its aggregated tuples for later
retrieval. This separation between detailed episodes and compact memories
preserves an auditable execution history without inserting complete programs
or execution logs into later prompts.

The \textsc{CompressAndIndex} procedure, detailed in Appendix Algorithm~\ref{alg:compress_and_Index}, converts a completed refinement trajectory into compact cross-task memory. It loads the recorded episodes, prioritizes those associated with reward improvement, aggregates their structural, task, performance, and correctness tags, and distills the corresponding textual gradients into concise, reusable guidance. The resulting lesson is stored with its reward improvement and supporting metadata, then indexed by its aggregated tags so that it can be retrieved during subsequent iterations and tasks. When no episode improves the reward, the procedure uses the complete trajectory to ensure that a memory record can still be formed.

\subsection{Cross-Task Knowledge Persistence}
\label{subsection:cross_task_persistence}

Meta-learning occurs through two complementary persistent artifacts. First,
the best-observed natural-language skill can initialize later tasks, allowing
general coding and optimization instructions learned during prior refinement
to influence generation without additional task-specific updates. Second, the
memory store preserves execution-derived lessons independently of the skill,
allowing later tasks to retrieve strategies associated with similar MPI
constructs, parallel patterns, or performance symptoms. The skill therefore
encodes broadly applicable generation behavior, whereas memory provides
selective access to more specific optimization experience. Because both
artifacts are represented externally in natural language and structured
metadata, they can be inspected, transferred, or reused without modifying the
underlying model parameters. Section~\ref{sec:cross_task_transfer_results}
evaluates these two transfer channels separately.

\section{Evaluation}
\subsection{Experimental Setup}
We evaluated \textsc{MetaFORGE} on six MPI C tasks, each at small, medium,
and large workload sizes, using five independent trials per task--workload
configuration. Primary experiments used
\texttt{gpt-5.1-codex-max} through the OpenAI Python API (v2.36.0).
Candidate programs were evaluated on a four-node OpenMPI~4.1 cluster.
Complete model, trial, compilation, static-analysis, and execution-environment
details are provided in
Appendices~\ref{appendix:primary_model_configuration}
and~\ref{appendix:execution_environment}.

We compared \textsc{MetaFORGE} against two baselines. The first was zero-shot
Codex generation, in which \texttt{gpt-5.1-codex-max} generated code using an
empty skill specification. The second was \textsc{TextGrad}, an iterative
skill-refinement baseline based on the unmodified published procedure
~\cite{yuksekgonul2024textgradautomaticdifferentiationtext}. Zero-shot Codex
provides the reference point for percentage improvements, whereas
\textsc{TextGrad} provides the primary iterative-refinement comparison.
The \textsc{MetaFORGE} and \textsc{TextGrad} conditions shared the initial
skill, task specification, code-generation model, retrieval context, iteration
budget, patience threshold, and compilation and evaluation pipeline. Each
retained its native refinement procedure: \textsc{MetaFORGE} used two
refinement calls in addition to code generation, whereas \textsc{TextGrad}
used its standard three-stage refinement sequence and default backward model.
The comparison should therefore be interpreted as an end-to-end method
comparison rather than a model-call-matched update-rule ablation. Complete
baseline controls, refinement-model settings, and model-call sequences are
provided in Appendix~\ref{appendix:baseline_configuration}.
We distinguish the controlled variables from the treatment and its measured
consequences. The shared generator, initial skill, evaluator, reward function,
retrieval configuration, iteration budget, and stopping rule are controlled.
The feedback representation and native refinement procedure constitute the
algorithmic treatment, whereas realized model-call count and token consumption
are measured consequences of those procedures. Refinement-model sensitivity is
evaluated separately in Section~\ref{sec:refiner_sensitivity}.

We evaluated correctness using task-specific output checks and measured
compiler and static-analysis warnings, externally measured whole-job runtime,
rank-level load imbalance, and token consumption. The reward weights were
$\alpha_c=1.0$, $\alpha_s=1.0$, $\alpha_t=2.0$,
$\alpha_\lambda=1.5$, and $\alpha_f=10000$. Static-analysis, runtime, and
imbalance results report improvement relative to the corresponding no-skill
zero-shot program. Trial-level results and complete-run token consumption are
reported in Appendix~\ref{appendix:experimental_trials}.

Task selection was criterion-driven: the six tasks were selected to span
communication topology, load distribution, and dominant performance
bottlenecks rather than to maximize application-domain count. The suite
includes terminal reductions and gathers, repeated point-to-point halo
exchange, repeated collectives, statically balanced and data-dependent
workloads, and bottlenecks in floating-point throughput, memory bandwidth,
cache locality, load imbalance, and communication latency. Structural overlap
is intentional: for example, $\pi$ estimation and Dot Product both terminate
in a reduction but stress different computational resources. We therefore
characterize the suite as a \emph{purposeful but bounded sample of MPI program
structures}, not a comprehensive survey of MPI applications. The mapping from
each task to its dominant parallel structure, communication pattern,
performance-limiting resource, and distinctive challenge is summarized in
Appendix Table~\ref{tab:task_selection_criteria}.

\subsection{Overall Results}

\begin{table*}[t]
\centering
\footnotesize
\setlength{\tabcolsep}{3.5pt}
\begin{tabular}{l|rrrr|rrrr}
\toprule
& \multicolumn{4}{c|}{\textbf{MetaFORGE}}
& \multicolumn{4}{c}{\textbf{TextGrad}} \\
\textbf{Task}
& \textbf{Static} & \textbf{Runtime} & \textbf{Balance} & \textbf{Tokens}
& \textbf{Static} & \textbf{Runtime} & \textbf{Balance} & \textbf{Tokens} \\
\midrule
$\pi$ calculation
& \textbf{100.00} & \textbf{16.45} & \textbf{96.41} & \textbf{8232}
& 8.89 & 7.30 & 35.38 & 12409 \\
Dot Product
& \textbf{100.00} & \textbf{27.00} & \textbf{50.74} & \textbf{9106}
& 5.18 & 14.84 & 23.59 & 14362 \\
Matrix Multiplication
& \textbf{57.16} & \textbf{87.99} & \textbf{47.04} & \textbf{15641}
& -13.14 & 55.46 & 16.91 & 18032 \\
Mandelbrot Set
& \textbf{78.47} & \textbf{9.32} & \textbf{25.29} & 17435
& 6.30 & -0.19 & 0.15 & \textbf{16604} \\
Heat Transfer
& \textbf{63.08} & \textbf{82.46} & \textbf{51.41} & 21350
& 10.38 & 14.87 & 22.94 & \textbf{19375} \\
Distributed PCA
& \textbf{94.44} & \textbf{86.36} & \textbf{38.27} & \textbf{17966}
& 22.32 & 31.75 & 11.64 & 24299 \\
\bottomrule
\end{tabular}
\caption{Task-average results over 15 trials per row (three workloads and five
trials per workload). Static-analysis, runtime, and load-balance values are
percentage improvements over zero-shot Codex; higher is better. Tokens are
mean per-iteration consumption; lower is better. Bold indicates the better
value within each method pair.}
\label{tab:overall_results}
\end{table*}

Table~\ref{tab:overall_results} summarizes results across six
MPI tasks and three workload sizes. Relative to zero-shot Codex,
\textsc{MetaFORGE} achieves positive task-average improvements in
static-analysis findings, runtime, and load balance for every task.
At the task-average level, it exceeds \textsc{TextGrad} on all 18
outcome comparisons---six tasks across three metrics. At the workload
level, \textsc{MetaFORGE} achieves the larger improvement in 52 of 54
static-analysis, runtime, and load-balance comparisons. The two exceptions
are runtime for medium Dot Product and load balance for large Matrix
Multiplication.

The results also reveal task-dependent optimization patterns. The largest
runtime improvements occur for Matrix Multiplication, Heat Transfer, and
Distributed PCA, whereas $\pi$ calculation shows the strongest
static-analysis and load-balance improvements. Dot Product adds a
memory-bandwidth-sensitive reduction workload: \textsc{MetaFORGE} achieves
100.00\% task-average static-analysis improvement, 27.00\% runtime
improvement, and 50.74\% load-balance improvement, compared with 5.18\%,
14.84\%, and 23.59\% for \textsc{TextGrad}. Mandelbrot generation produces
smaller runtime gains, but \textsc{MetaFORGE} remains positive while
\textsc{TextGrad} is approximately neutral.

Token consumption presents a mixed tradeoff. \textsc{MetaFORGE} uses fewer
per-iteration tokens for $\pi$ calculation, Dot Product, Matrix Multiplication,
and Distributed PCA, but more for Heat Transfer and Mandelbrot. Because the
methods differ in native model-call count and stopping behavior, per-iteration
consumption should not be interpreted as complete-run cost. Including Dot
Product, \textsc{MetaFORGE} uses fewer total tokens in eight of the 18
workload-level comparisons. Detailed workload, trial-level, and complete-run
results are provided in Appendix~\ref{appendix:experimental_trials}.

\subsection{Refinement-Model Sensitivity}
\label{sec:refiner_sensitivity}

To test whether the primary comparison is explained by refinement-model
capability, we conducted a six-task model-matched ablation. All conditions use the
same \texttt{gpt-5.3-codex} generator, seed skill, evaluator, reward, stopping
rule, and frozen read-only memory snapshot. \textsc{MetaFORGE}-Strong uses
\texttt{gpt-5.3-codex} for refinement, \textsc{MetaFORGE}-Weak uses
\texttt{gpt-4o}, and \textsc{TextGrad}-Native retains its published
\texttt{gpt-4o} refinement configuration. Mean within-task ranks were 5.30,
7.30, and 11.40, respectively (lower is better).

\begin{table}[t]
\centering
\scriptsize
\begin{tabular}{@{}lrrr@{}}
\toprule
\textbf{Comparison} &
$\boldsymbol{\Delta}$ \textbf{rank} &
\textbf{Exact $p$} &
\textbf{Holm $p$} \\
\midrule
Strong vs. Weak
& -1.533 & 0.054 & 0.054 \\
Weak vs. TextGrad
& -2.933 & $1.28\times10^{-4}$ & $2.55\times10^{-4}$ \\
Strong vs. TextGrad
& -3.867 & $1.38\times10^{-7}$ & $4.15\times10^{-7}$ \\
\bottomrule
\end{tabular}
\caption{Exact task-stratified pairwise comparisons of best reward in the
refinement-model ablation. For each comparison, ranks are recomputed
using only the two methods in that comparison. Therefore, $\Delta$ rank is not
the arithmetic difference between the three-method mean ranks reported in the
text. Negative $\Delta$ rank favors the first method. Holm adjustment covers
the three comparisons.}
\label{tab:refiner_posthoc_main}
\end{table}

Both MetaFORGE variants retain significant best-reward advantages over
\textsc{TextGrad}-Native after multiplicity correction, while Strong versus
Weak does not reach the corrected $\alpha=0.05$ threshold. Thus, the advantage
persists when MetaFORGE's refiner is reduced to \texttt{gpt-4o}; the experiment
does not establish equivalence between the two MetaFORGE refiners. Full
per-task descriptive statistics and experimental controls appear in
Appendix~\ref{appendix:refiner_strength_ablation}.

\subsection{Knowledge Transfer}
\label{sec:cross_task_transfer_results}

Our retrieval-context ablation covers all six tasks and all three workload
sizes. Across the 18 task--workload configurations, retrieval increases mean
static-analysis improvement from 63.34\% to 82.19\%, runtime improvement from
35.99\% to 51.59\%, and load-imbalance improvement from 42.27\% to 51.03\%.
Mean per-iteration token consumption decreases from 15,225 to 14,955 tokens
($-1.8\%$). Runtime improves in all 18 configurations; static-analysis quality
improves in 15 with three ties, and load balance improves in 16. Because these
are configuration-level means rather than independent trial-level observations,
we treat the cross-configuration summaries as descriptive. Full results are
reported in Appendix Table~\ref{tab:context_ablation_all}.

A separate cross-provider ablation on Distributed PCA contrasts Codex and Claude
within \textsc{MetaFORGE}. Both improve code quality, runtime, and load balance,
but with different tradeoffs: Codex achieves larger runtime gains and lower
token use, whereas Claude achieves greater load-balance improvement. We treat
these as empirical model-specific behaviors rather than evidence for a hidden
mechanism and complete results appear in Appendix Table~\ref{tab:model_ablation_pca}.

The primary six-task evaluation establishes within-task optimization; our
evidence for cross-task persistence comes from the held-out and zero-update
studies described next, which use \texttt{gpt-5.3-codex} after the primary
model became unavailable. A three-task leave-one-task-out study separately
evaluates the two persistence mechanisms described in
Section~\ref{subsection:cross_task_persistence}.  Read-only memory
learned exclusively from the two non-held-out tasks improves mean best reward
from $-7.24$ to $-1.21$ for Calculate~$\pi$, from $-141.73$ to $-106.63$ for
Matrix Multiplication, and from $-38.27$ to $-17.35$ for Mandelbrot Sets
($p=0.0039$ for each task). Under zero-update evaluation, the transferred skill
alone also improves reward for all three tasks ($p\leq0.0078$), and every
generated program passes correctness testing. Together, these results provide
evidence that reusable optimization knowledge is retained in both the skill and
external memory under the evaluated protocol. Full protocols, component
metrics, and statistical qualifications are provided in
Appendix~\ref{held_out_cross_task_transfer_study} and
Appendix~\ref{zero_update_study}.

\section{Conclusions}
We presented \textsc{MetaFORGE}, a meta-learning harness that optimizes
persistent natural-language skills for MPI C code generation using
execution-grounded feedback and retrieval-augmented memory. Across six tasks,
it achieves larger task-average static-analysis, runtime, and load-balance
improvements than the evaluated \textsc{TextGrad} baseline, with mixed token-cost
tradeoffs. Retrieval improves runtime in all 18 evaluated task--workload
configurations, and the model-matched ablation shows that MetaFORGE retains a
best-reward advantage when its refiner is reduced to \texttt{gpt-4o}. Held-out
and zero-update studies further support cross-task persistence in both the skill
and external memory under the evaluated protocol.

Future work will extend \textsc{MetaFORGE} beyond MPI C to OpenMP, CUDA, SYCL,
and hybrid MPI+X programming, and will investigate task-normalized retrieval
signals, automatic reward-weight learning, and task-specification refinement.

\section*{Limitations}
This study has several limitations. First, the evaluation covers a purposeful
but bounded six-task sample of MPI C program structures at one fixed four-rank
configuration. It does not cover irregular sparse or graph workloads,
unstructured meshes, other process counts, other parallel-programming models,
or other hardware platforms. The results therefore should not be interpreted as
establishing general performance improvements across MPI or parallel computing
broadly.

Second, the primary \textsc{MetaFORGE}--\textsc{TextGrad} comparison evaluates
the complete end-to-end methods. The model-matched ablation reduces the
refinement-model confound, but the methods still differ in native model-call
structure and feedback representation. We therefore attribute the primary
comparison to the evaluated systems as configured rather than to a single
isolated algorithmic component.

Third, the fixed-weight scalar reward is a within-task optimization signal and
is not task-normalized. Raw reward magnitudes are not interpreted as directly
comparable across tasks. In addition, the retrieval score assigns weight 0.10
to $\clamp(\Delta r_m/2,0,1)$; because $\Delta r_m$ is scale dependent, this
term can saturate for large-runtime tasks and underweight useful memories from
smaller-scale tasks. This is a limitation of the current cross-task retrieval
ranking and motivates task-normalized confidence signals in future work.

Fourth, the held-out transfer study covers three tasks, and Phase~1 produces
one transferred skill and memory store per fold that are reused across the
evaluation trials. The reported replication therefore estimates variability
in held-out generation and adaptation rather than variability from independently
repeated meta-training pipelines. Transfer conclusions are consequently
conditional on the shared Phase~1 training outcome and the evaluated three-task
protocol.

Finally, Codex, Claude, and the evaluated APIs are closed-source commercial
systems. Their training data, internal implementations, and future availability
are outside researchers' control. The original model was also deprecated during
the study, requiring the later transfer and sensitivity experiments to use the
replacement model documented in the appendix. Those later experiments are
internally consistent within their own conditions but are not strictly
commensurable with the primary experiments conducted using the original model.

\bibliography{ref}

\appendix
\newpage 

\clearpage

\section{Reproducibility Artifacts}
\label{appendix:reproducibility_artifacts}

\subsection{Primary Model and Trial Protocol}
\label{appendix:primary_model_configuration}
The primary experiments configured \textsc{MetaFORGE} to use OpenAI's Codex
agent with the \texttt{gpt-5.1-codex-max} model through the OpenAI Python API
(v2.36.0). We selected \texttt{gpt-5.1-codex-max} because it is designed for
agentic coding and long-running software-engineering tasks, making it
well suited to \textsc{MetaFORGE}'s iterative generation, evaluation,
textual-gradient, and skill-rewrite workflow
~\cite{openai2025gpt51codexmax,openai2025gpt51codexmax_release}.
This selection is also consistent with prior evidence that Codex-style models
support program synthesis and iterative code refinement
~\cite{chen2021evaluating,CODEX_openai2025codex}.

Because code generation and skill rewriting are nondeterministic, each
task--workload configuration was evaluated using five independent trials.
Each trial used a separate API conversation so that conversational state was
not shared across repetitions. The trials therefore represent independent
generation and refinement trajectories executed with the same task,
workload, model, iteration budget, stopping criterion, and evaluation
configuration.

\subsection{Criterion-Driven Task Selection}
\label{appendix:task_selection}

The evaluation suite was constructed to vary communication topology, load
distribution, and dominant performance bottlenecks rather than to accumulate
application domains. Table~\ref{tab:task_selection_criteria} summarizes the
structural criterion contributed by each task. The suite is intended as a
purposeful but bounded sample of MPI program structures; it is not intended to
represent the full space of MPI applications.

\begin{table*}[t]
\centering

\scriptsize
\setlength{\tabcolsep}{3pt}
\renewcommand{\arraystretch}{1.10}
\begin{tabularx}{\textwidth}{
    >{\raggedright\arraybackslash}p{0.11\textwidth}
    >{\raggedright\arraybackslash}p{0.19\textwidth}
    >{\raggedright\arraybackslash}p{0.19\textwidth}
    >{\raggedright\arraybackslash}p{0.15\textwidth}
    >{\raggedright\arraybackslash}X}
\toprule
\textbf{Task} &
\textbf{Dominant parallel structure} &
\textbf{Communication pattern} &
\textbf{Limiting factor} &
\textbf{Distinctive challenge} \\
\midrule

$\pi$ estimation &
Naturally parallel integration &
Single terminal reduction &
Floating-point throughput &
Accuracy to $10^{-9}$ under partitioned summation \\

Dot Product &
Data-parallel reduction over distributed operands &
Single terminal reduction &
Memory bandwidth &
Distributed initialization without full-vector broadcast \\

Dense Matrix Multiplication &
Dense linear algebra, $\mathcal{O}(MKN)$ &
Operand distribution and result reduction &
Cache locality / loop order &
Non-divisible $2513\times7999\times3857$ dimensions; BLAS prohibited \\

Mandelbrot Set &
Irregular, data-dependent workload &
Terminal gather of image tiles &
Load imbalance &
Per-pixel cost varies substantially; static decomposition is ineffective \\

Heat Transfer (Jacobi) &
Iterative stencil relaxation &
Nearest-neighbor halo exchange every iteration &
Communication latency &
Ghost-cell correctness and communication/computation behavior over
30,000 iterations \\

Distributed PCA &
Multi-phase iterative linear algebra &
Repeated collectives nested within power iterations &
Collective frequency &
Multi-phase pipeline; $N=1{,}000{,}003$ samples; eigenvector stability \\

\bottomrule
\end{tabularx}
\caption{Criterion-driven selection of the six MPI evaluation tasks. Each task
was selected to exercise a parallel-programming property not fully represented
by the others.}
\label{tab:task_selection_criteria}
\end{table*}

\subsection{Compilation and Execution Environment}
\label{appendix:execution_environment}

Generated MPI C programs were compiled using
\texttt{mpicc -Wall -Wextra -Wpedantic -O2 -std=c11} with
Clang~18.1.3. Static analysis used Clang-Tidy~18.1.3 with the
\texttt{clang-diagnostic-*}, \texttt{clang-analyzer-*},
\texttt{clang-analyzer-optin.mpi.MPI-Checker}, \texttt{mpi-*},
\texttt{bugprone-*}, and \texttt{performance-*} check groups.

Runtime evaluation was performed on a four-node OpenMPI~4.1 cluster running
Ubuntu Linux. Each node contained an Intel Core i7-8700K processor and 16\,GB
RAM, connected through switched full-duplex Gigabit Ethernet. Four MPI ranks
were used. Rank placement was not explicitly configured, but empirical
verification showed one rank per node. CPU affinity was not explicitly
controlled, and CPU governor and turbo settings remained at system defaults. 

Each candidate program was executed 20 times sequentially without recompilation and
no initial execution was discarded as a warm-up. The reported runtime is the
mean across the 20 executions. Runtime and rank-level measurements were collected
externally by the harness rather than self-reported by generated programs.
Compilation, correctness testing, static analysis, runtime measurement, and
load-imbalance evaluation were implemented independently of the active
refinement backend, ensuring that all evaluated methods used the same
program-assessment pipeline. All runtime values reported in the primary results were obtained under this measurement protocol. 

\subsection{Controlled Baseline Configuration}
\label{appendix:baseline_configuration}

We compared \textsc{MetaFORGE} against two baselines:
(1) Codex generation using an empty skill specification and
(2) the unmodified \textsc{TextGrad} implementation for iterative skill
refinement~\cite{yuksekgonul2024textgradautomaticdifferentiationtext}.
The \textsc{MetaFORGE} and \textsc{TextGrad} conditions used matched
code-generation and evaluation configurations. Both began from the same
initial skill specification, received the same task specification for each
workload, and used \texttt{gpt-5.1-codex-max} to generate candidate programs.
Both methods were executed through the same outer optimization loop with the
same maximum-iteration budget and a patience-based stopping criterion of eight
consecutive non-improving iterations.

Both methods received the same retrieval context. Before each refinement
step, the harness queried the shared cross-task memory store and supplied the
same top-$k$ retrieved records to the active refinement mechanism. The task
description, generated source code, compiler diagnostics, and static-analysis
findings were likewise provided to both methods. The feedback representations
differed according to the native refinement procedures. With
\texttt{--inject-reward-to-gradient} enabled, \textsc{MetaFORGE} additionally
received a numeric comparison of the current and preceding iteration metrics
and an explicit decomposition of the weighted reward. The \textsc{TextGrad}
loss-evaluation path used qualitative evidence and did not consume either
numeric artifact.

The refinement procedures differed in their native sequence of model calls.
Each \textsc{MetaFORGE} iteration used one call to generate an
execution-grounded textual critique and a second call to apply that critique
to the skill, resulting in three LLM calls per iteration when code generation
is included. In contrast, the unmodified \textsc{TextGrad} implementation
performed its standard forward loss-evaluation call, backward textual-gradient
call, and textual gradient-descent optimizer call, resulting in four LLM calls
per iteration. These differing call counts arise from the algorithms'
respective refinement procedures rather than from different iteration budgets
or implementation-specific modifications.

The refinement models also followed the native configuration of each method.
\textsc{MetaFORGE} used \texttt{gpt-5.1-codex-max} for its critique and
skill-rewrite calls, whereas the unmodified \textsc{TextGrad} backward engine
used \texttt{gpt-4o}, its default configured model. Thus, the comparison
controls the initial skill, code-generation model, task inputs, retrieval
context, iteration budget, stopping criterion, and evaluation environment
while comparing the complete end-to-end refinement procedures as implemented.
The resulting differences in refinement-model configuration and model-call
count are properties of the evaluated methods rather than experimentally
assigned resource differences.

\section{Experimental Trial Results}
\label{appendix:experimental_trials}

\begin{figure*}[h]
    \centering
    \includegraphics[width=1.0\linewidth]{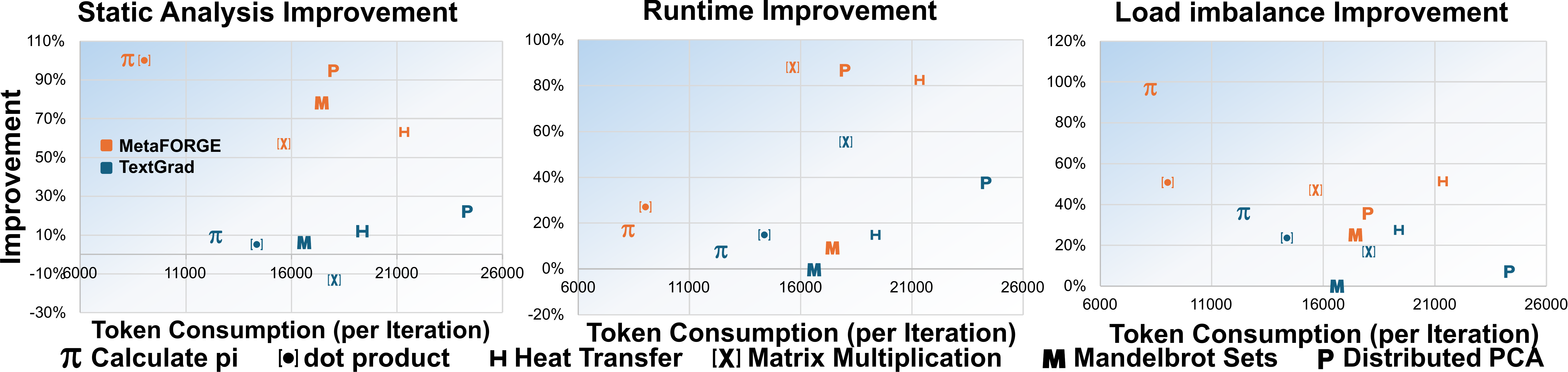}
    \caption{Average optimization improvement versus per-iteration token
consumption for MetaFORGE and TextGrad across six MPI programming tasks.
Across static-analysis quality, runtime, and load-imbalance metrics,
MetaFORGE achieves larger improvements at lower per-iteration token cost for
most tasks, indicating a favorable per-iteration cost--performance tradeoff
relative to TextGrad. MetaFORGE consumes more tokens per iteration for Heat
Transfer and Mandelbrot while achieving larger performance improvements.
Complete-run token consumption is reported separately in the trial-level
results.}

    \label{fig:average}
\end{figure*}

\subsection{Calculate $\pi$}
\label{appendix:pi_results}
\begin{figure*}[ht]
    \centering
    \includegraphics[width=1.0\linewidth]{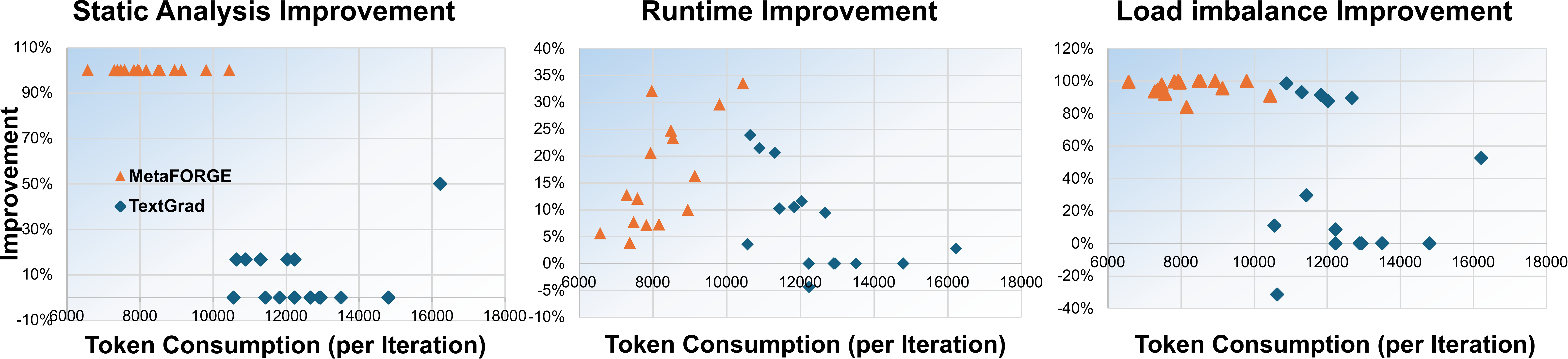}
    \caption{MPI $\pi$ calculation results comparing MetaFORGE and TextGrad~\cite{yuksekgonul2024textgradautomaticdifferentiationtext}. MetaFORGE achieves higher code-quality, runtime, and load-balance improvements with lower token consumption across small, medium, and large workloads over 15 trials (five trials per workload).} 
    \label{fig:compute_pi}
\end{figure*}

The calculate $\pi$ task directed Codex to approximate $\pi$ using the summation-of-integrals method. We evaluated three workload sizes, \texttt{pi\_small}, \texttt{pi\_medium}, and \texttt{pi\_large}, using 134,217,728, 268,435,456, and 536,870,912 intervals, respectively. Figure~\ref{fig:compute_pi} summarizes the effectiveness--cost tradeoff between MetaFORGE and TextGrad, while Table~\ref{tab:overall_results} reports the aggregate results across all $\pi$ workloads.

For the $\pi$ calculation task, MetaFORGE outperforms TextGrad across all measured dimensions. MetaFORGE achieves a 100.00\% static-analysis improvement compared with 8.89\% for TextGrad, indicating a much stronger reduction in static-analysis warnings. It also improves runtime by 16.45\%, more than twice TextGrad's 7.30\% improvement, and reduces load imbalance by 96.41\% compared with 35.38\% for TextGrad. These results show that MetaFORGE produces higher-quality, faster, and better-balanced MPI implementations for numerical integration.

MetaFORGE also achieves these improvements with lower token consumption, consuming 8,232.34 tokens per iteration compared with 12,409.15 for TextGrad. Together, Figure~\ref{fig:compute_pi} and Table~\ref{tab:overall_results} show that MetaFORGE provides a stronger effectiveness--cost tradeoff for MPI $\pi$ computation.

\subsubsection{Calculate-$\pi$ Trial-Level Results}
\label{appendix:pi_trial_level_results}

Tables~\ref{tab:pi_small_trial_summary}--%
\ref{tab:pi_large_trial_summary} report the five independent trials for each
$\pi$-calculation workload. Static, runtime, and imbalance columns report
percentage improvement over the corresponding no-skill zero-shot baseline;
higher values are better, and negative values denote regression. The
``Exec. iter.'' column counts iterations for which the experimental record
contains a measured runtime. The best iteration and reward identify the
reward-maximizing candidate observed within each refinement trajectory. Total
tokens are summed across the complete trajectory. The final two rows report the
arithmetic mean and sample standard deviation of the five trial-level values.
Because the tables report the arithmetic mean of trial-level percentages, these values may differ from aggregate percentages computed from pooled raw measurements.

\begin{table*}[t]
\centering

\scriptsize
\setlength{\tabcolsep}{2.2pt}
\begin{adjustbox}{max width=\textwidth}
\begin{tabular}{ccccccccccc}
\toprule
\textbf{Trial} & \textbf{Method} & \makecell{\textbf{Iter.}} & \makecell{\textbf{Exec.}\\\textbf{iter.}} & \makecell{\textbf{Best}\\\textbf{iter.}} & \makecell{\textbf{Best}\\\textbf{reward}} & \makecell{\textbf{Static}\\\textbf{improv.} \textbf{(\%)}} & \makecell{\textbf{Runtime}\\\textbf{improv.} \textbf{(\%)}} & \makecell{\textbf{Balance}\\\textbf{improv.} \textbf{(\%)}} & \makecell{\textbf{Total}\\\textbf{tokens}} & \makecell{\textbf{Tokens/}\\\textbf{iteration}} \\
\midrule
\multirow{2}{*}{1} & \textsc{MetaFORGE} & 12 & 11 & 6 & -1.072 & 100.00 & 7.70 & 97.91 & 89,657 & 7,471.42 \\
 & \textsc{TextGrad} & 12 & 12 & 6 & -6.082 & 16.67 & 23.88 & -31.46 & 127,622 & 10,635.17 \\
\addlinespace[1.5pt]
\multirow{2}{*}{2} & \textsc{MetaFORGE} & 16 & 15 & 10 & -1.068 & 100.00 & 20.56 & 99.83 & 126,852 & 7,928.25 \\
 & \textsc{TextGrad} & 9 & 9 & 3 & -6.083 & 0.00 & 3.55 & 11.03 & 95,055 & 10,561.67 \\
\addlinespace[1.5pt]
\multirow{2}{*}{3} & \textsc{MetaFORGE} & 20 & 18 & 14 & -1.070 & 100.00 & 9.98 & 99.89 & 178,765 & 8,938.25 \\
 & \textsc{TextGrad} & 8 & 8 & 2 & -6.078 & 16.67 & 21.42 & 98.45 & 87,075 & 10,884.38 \\
\addlinespace[1.5pt]
\multirow{2}{*}{4} & \textsc{MetaFORGE} & 20 & 18 & 18 & -1.071 & 100.00 & 33.55 & 90.87 & 208,787 & 10,439.35 \\
 & \textsc{TextGrad} & 11 & 11 & 5 & -6.075 & 0.00 & 10.19 & 29.54 & 125,670 & 11,424.55 \\
\addlinespace[1.5pt]
\multirow{2}{*}{5} & \textsc{MetaFORGE} & 9 & 9 & 3 & -1.074 & 100.00 & 32.08 & 98.94 & 71,694 & 7,966.00 \\
 & \textsc{TextGrad} & 12 & 12 & 6 & -6.079 & 16.67 & -4.40 & 8.60 & 146,711 & 12,225.92 \\
\midrule
\multirow{2}{*}{\makecell{\textbf{Mean}\\$\boldsymbol{\pm}$ \textbf{SD}}}
& \textsc{MetaFORGE} & $15.4 \pm 4.9$ & $14.2 \pm 4.1$ & $10.2 \pm 6.0$ & $-1.071 \pm 0.002$ & $100.00 \pm 0.00$ & $20.77 \pm 12.03$ & $97.49 \pm 3.79$ & $135,151 \pm 58,073$ & $8,548.65 \pm 1,184.41$ \\
& \textsc{TextGrad} & $10.4 \pm 1.8$ & $10.4 \pm 1.8$ & $4.4 \pm 1.8$ & $-6.079 \pm 0.003$ & $10.00 \pm 9.13$ & $10.93 \pm 11.92$ & $23.23 \pm 47.57$ & $116,427 \pm 24,729$ & $11,146.33 \pm 691.85$ \\
\bottomrule
\end{tabular}
\end{adjustbox}
\caption{Per-trial $\pi$-calculation results for the small workload.}
\label{tab:pi_small_trial_summary}
\end{table*}

\begin{table*}[t]
\centering
\scriptsize
\setlength{\tabcolsep}{2.2pt}
\begin{adjustbox}{max width=\textwidth}
\begin{tabular}{ccccccccccc}
\toprule
\textbf{Trial} & \textbf{Method} & \makecell{\textbf{Iter.}} & \makecell{\textbf{Exec.}\\\textbf{iter.}} & \makecell{\textbf{Best}\\\textbf{iter.}} & \makecell{\textbf{Best}\\\textbf{reward}} & \makecell{\textbf{Static}\\\textbf{improv.} \textbf{(\%)}} & \makecell{\textbf{Runtime}\\\textbf{improv.} \textbf{(\%)}} & \makecell{\textbf{Balance}\\\textbf{improv.} \textbf{(\%)}} & \makecell{\textbf{Total}\\\textbf{tokens}} & \makecell{\textbf{Tokens/}\\\textbf{iteration}} \\
\midrule
\multirow{2}{*}{1} & \textsc{MetaFORGE} & 10 & 9 & 4 & -1.145 & 100.00 & 12.75 & 93.55 & 72,877 & 7,287.70 \\
 & \textsc{TextGrad} & 12 & 11 & 6 & -6.159 & 0.00 & 10.53 & 91.50 & 141,907 & 11,825.58 \\
\addlinespace[1.5pt]
\multirow{2}{*}{2} & \textsc{MetaFORGE} & 10 & 9 & 4 & -1.141 & 100.00 & 12.10 & 92.19 & 75,734 & 7,573.40 \\
 & \textsc{TextGrad} & 14 & 14 & 8 & -6.160 & 16.67 & 20.58 & 93.19 & 158,231 & 11,302.21 \\
\addlinespace[1.5pt]
\multirow{2}{*}{3} & \textsc{MetaFORGE} & 19 & 19 & 19 & -1.139 & 100.00 & 16.30 & 95.37 & 173,519 & 9,132.58 \\
 & \textsc{TextGrad} & 17 & 17 & 11 & -6.174 & 16.67 & 11.55 & 87.58 & 204,478 & 12,028.12 \\
\addlinespace[1.5pt]
\multirow{2}{*}{4} & \textsc{MetaFORGE} & 12 & 12 & 6 & -1.135 & 100.00 & 7.26 & 83.96 & 97,943 & 8,161.92 \\
 & \textsc{TextGrad} & 12 & 12 & 6 & -6.175 & 0.00 & 9.43 & 89.61 & 152,069 & 12,672.42 \\
\addlinespace[1.5pt]
\multirow{2}{*}{5} & \textsc{MetaFORGE} & 20 & 18 & 14 & -1.135 & 100.00 & 29.58 & 99.94 & 196,072 & 9,803.60 \\
 & \textsc{TextGrad} & 7 & 6 & 1 & -6.345 & 0.00 & 0.00 & 0.00 & 90,611 & 12,944.43 \\
\midrule
\multirow{2}{*}{\makecell{\textbf{Mean}\\$\boldsymbol{\pm}$ \textbf{SD}}}
& \textsc{MetaFORGE} & $14.2 \pm 4.9$ & $13.4 \pm 4.8$ & $9.4 \pm 6.8$ & $-1.139 \pm 0.004$ & $100.00 \pm 0.00$ & $15.60 \pm 8.45$ & $93.00 \pm 5.84$ & $123,229 \pm 57,588$ & $8,391.84 \pm 1,058.72$ \\
& \textsc{TextGrad} & $12.4 \pm 3.6$ & $12.0 \pm 4.1$ & $6.4 \pm 3.6$ & $-6.202 \pm 0.080$ & $6.67 \pm 9.13$ & $10.42 \pm 7.32$ & $72.38 \pm 40.51$ & $149,459 \pm 40,715$ & $12,154.55 \pm 660.07$ \\
\bottomrule
\end{tabular}
\end{adjustbox}
\caption{Per-trial $\pi$-calculation results for the medium workload.}
\label{tab:pi_medium_trial_summary}
\end{table*}

\begin{table*}[t]
\centering

\scriptsize
\setlength{\tabcolsep}{2.2pt}
\begin{adjustbox}{max width=\textwidth}
\begin{tabular}{ccccccccccc}
\toprule
\textbf{Trial} & \textbf{Method} & \makecell{\textbf{Iter.}} & \makecell{\textbf{Exec.}\\\textbf{iter.}} & \makecell{\textbf{Best}\\\textbf{iter.}} & \makecell{\textbf{Best}\\\textbf{reward}} & \makecell{\textbf{Static}\\\textbf{improv.} \textbf{(\%)}} & \makecell{\textbf{Runtime}\\\textbf{improv.} \textbf{(\%)}} & \makecell{\textbf{Balance}\\\textbf{improv.} \textbf{(\%)}} & \makecell{\textbf{Total}\\\textbf{tokens}} & \makecell{\textbf{Tokens/}\\\textbf{iteration}} \\
\midrule
\multirow{2}{*}{1} & \textsc{MetaFORGE} & 10 & 9 & 4 & -1.281 & 100.00 & 5.66 & 99.49 & 65,642 & 6,564.20 \\
 & \textsc{TextGrad} & 7 & 7 & 1 & -7.346 & 0.00 & 0.00 & 0.00 & 90,282 & 12,897.43 \\
\addlinespace[1.5pt]
\multirow{2}{*}{2} & \textsc{MetaFORGE} & 12 & 11 & 7 & -1.279 & 100.00 & 23.41 & 99.99 & 102,491 & 8,540.92 \\
 & \textsc{TextGrad} & 7 & 7 & 1 & -6.477 & 0.00 & 0.00 & 0.00 & 85,579 & 12,225.57 \\
\addlinespace[1.5pt]
\multirow{2}{*}{3} & \textsc{MetaFORGE} & 12 & 11 & 6 & -1.330 & 100.00 & 7.16 & 99.70 & 93,851 & 7,820.92 \\
 & \textsc{TextGrad} & 7 & 7 & 1 & -6.755 & 0.00 & 0.00 & 0.00 & 94,509 & 13,501.29 \\
\addlinespace[1.5pt]
\multirow{2}{*}{4} & \textsc{MetaFORGE} & 11 & 10 & 5 & -1.287 & 100.00 & 3.89 & 94.72 & 81,061 & 7,369.18 \\
 & \textsc{TextGrad} & 7 & 5 & 1 & -7.377 & 0.00 & 0.00 & 0.00 & 103,551 & 14,793.00 \\
\addlinespace[1.5pt]
\multirow{2}{*}{5} & \textsc{MetaFORGE} & 14 & 14 & 8 & -1.263 & 100.00 & 24.78 & 99.86 & 118,823 & 8,487.36 \\
 & \textsc{TextGrad} & 20 & 19 & 15 & -4.435 & 50.00 & 2.75 & 52.69 & 324,311 & 16,215.55 \\
\midrule
\multirow{2}{*}{\makecell{\textbf{Mean}\\$\boldsymbol{\pm}$ \textbf{SD}}}
& \textsc{MetaFORGE} & $11.8 \pm 1.5$ & $11.0 \pm 1.9$ & $6.0 \pm 1.6$ & $-1.288 \pm 0.025$ & $100.00 \pm 0.00$ & $12.98 \pm 10.22$ & $98.75 \pm 2.26$ & $92,374 \pm 20,290$ & $7,756.51 \pm 825.41$ \\
& \textsc{TextGrad} & $9.6 \pm 5.8$ & $9.0 \pm 5.7$ & $3.8 \pm 6.3$ & $-6.478 \pm 1.206$ & $10.00 \pm 22.36$ & $0.55 \pm 1.23$ & $10.54 \pm 23.56$ & $139,646 \pm 103,443$ & $13,926.57 \pm 1,590.87$ \\
\bottomrule
\end{tabular}
\end{adjustbox}
\caption{Per-trial $\pi$-calculation results for the large workload.}
\label{tab:pi_large_trial_summary}
\end{table*}

\subsection{Dot Product}

We selected the MPI dot product as a canonical data-parallel reduction
workload widely used in numerical and scientific computing. Efficient
implementations must partition input vectors, compute local partial products, and reduce partial sums while minimizing synchronization and load imbalance. We evaluated three dot product workload sizes:
\texttt{dot\_product\_small}, \texttt{dot\_product\_medium}, and
\texttt{dot\_product\_large} across five trials each.
Table~\ref{tab:overall_results}
compares \textsc{MetaFORGE} with \textsc{TextGrad}.

MetaFORGE achieved 100.00\% static-analysis improvement for all three workloads, compared with 0.00\%, 11.11\%, and 4.44\% for TextGrad. MetaFORGE also produced stronger load-balance improvements of 46.47\%, 45.67\%, and 60.08\%, versus 38.64\%, 16.23\%, and 15.92\% for TextGrad. Runtime improvements favored MetaFORGE on the small and large workloads (19.27\% and 40.08\% versus 2.54\% and 18.96\%), while TextGrad slightly exceeded MetaFORGE on the medium workload (23.03\% versus 21.65\%).

MetaFORGE additionally required fewer tokens per iteration across all workload
sizes: 9,506 versus 13,673 tokens for small, 8,050 versus 15,714 for medium,
and 9,760 versus 13,699 for large. Overall, MetaFORGE provides a favorable
effectiveness--cost tradeoff, achieving substantially better code quality and
load balance, competitive or superior runtime performance, and lower token
consumption.

\subsubsection{Dot-Product Trial-Level Results}
\label{appendix:dot_product_trial_level_results}

Tables~\ref{tab:dot_small_trial_summary}--\ref{tab:dot_large_trial_summary} report the five independent trials for each dot-product workload. Static, runtime, and imbalance columns report percentage improvement over the corresponding no-skill zero-shot baseline; higher values are better, and negative values denote regression. The best iteration and reward identify the reward-maximizing candidate observed within each refinement trajectory. Total tokens are summed across the complete trajectory, and tokens per iteration are the corresponding run-normalized values. The final two rows report the arithmetic mean and sample standard deviation of the five trial-level values.

\begin{table*}[t]
\centering

\scriptsize
\setlength{\tabcolsep}{2.6pt}
\begin{adjustbox}{max width=\textwidth}
\begin{tabular}{cccccccccc}
\toprule
\textbf{Trial} & \textbf{Method} & \textbf{Iter.} & \makecell{\textbf{Best}\\\textbf{iter.}} & \makecell{\textbf{Best}\\\textbf{reward}} & \makecell{\textbf{Static}\\\textbf{improv.} \textbf{(\%)}} & \makecell{\textbf{Runtime}\\\textbf{improv.} \textbf{(\%)}} & \makecell{\textbf{Balance}\\\textbf{improv.} \textbf{(\%)}} & \makecell{\textbf{Total}\\\textbf{tokens}} & \makecell{\textbf{Tokens/}\\\textbf{iteration}} \\
\midrule
\multirow{2}{*}{1} & \textsc{MetaFORGE} & 15 & 9 & -1.010 & 100.00 & 92.59 & -20.97 & 137,699 & 9,179.93 \\
 & \textsc{TextGrad} & 9 & 1 & -6.086 & 0.00 & 0.00 & 0.00 & 121,053 & 13,450.33 \\
\addlinespace[1.5pt]
\multirow{2}{*}{2} & \textsc{MetaFORGE} & 12 & 6 & -1.033 & 100.00 & 15.75 & 84.72 & 106,887 & 8,907.25 \\
 & \textsc{TextGrad} & 11 & 3 & -6.091 & 0.00 & 5.19 & 40.50 & 147,317 & 13,392.45 \\
\addlinespace[1.5pt]
\multirow{2}{*}{3} & \textsc{MetaFORGE} & 9 & 3 & -1.067 & 100.00 & -2.42 & -4.22 & 78,891 & 8,765.67 \\
 & \textsc{TextGrad} & 10 & 1 & -9.091 & 0.00 & 0.00 & 0.00 & 138,189 & 13,818.90 \\
\addlinespace[1.5pt]
\multirow{2}{*}{4} & \textsc{MetaFORGE} & 20 & 16 & -1.008 & 100.00 & -19.72 & 93.13 & 210,205 & 10,510.25 \\
 & \textsc{TextGrad} & 15 & 7 & -6.058 & 0.00 & 3.66 & 85.72 & 203,936 & 13,595.73 \\
\addlinespace[1.5pt]
\multirow{2}{*}{5} & \textsc{MetaFORGE} & 9 & 3 & -1.017 & 100.00 & 10.14 & 79.67 & 91,541 & 10,171.22 \\
 & \textsc{TextGrad} & 12 & 4 & -6.038 & 0.00 & 3.84 & 67.00 & 169,306 & 14,108.83 \\
\midrule
\multirow{2}{*}{\makecell{\textbf{Mean}\\$\boldsymbol{\pm}$ \textbf{SD}}} & \textsc{MetaFORGE} & $13.0 \pm 4.6$ & $7.4 \pm 5.4$ & $-1.027 \pm 0.024$ & $100.00 \pm 0.00$ & $19.27 \pm 43.19$ & $46.47 \pm 54.45$ & $125,045 \pm 52,428$ & $9,506.86 \pm 784.85$ \\
& \textsc{TextGrad} & $11.4 \pm 2.3$ & $3.2 \pm 2.5$ & $-6.673 \pm 1.352$ & $0.00 \pm 0.00$ & $2.54 \pm 2.39$ & $38.64 \pm 38.76$ & $155,960 \pm 31,972$ & $13,673.25 \pm 293.89$ \\
\bottomrule
\end{tabular}
\end{adjustbox}

\caption{Per-trial MPI dot-product results for the small workload.}
\label{tab:dot_small_trial_summary}
\end{table*}

\begin{table*}[t]
\centering

\scriptsize
\setlength{\tabcolsep}{2.6pt}
\begin{adjustbox}{max width=\textwidth}
\begin{tabular}{cccccccccc}
\toprule
\textbf{Trial} & \textbf{Method} & \textbf{Iter.} & \makecell{\textbf{Best}\\\textbf{iter.}} & \makecell{\textbf{Best}\\\textbf{reward}} & \makecell{\textbf{Static}\\\textbf{improv.} \textbf{(\%)}} & \makecell{\textbf{Runtime}\\\textbf{improv.} \textbf{(\%)}} & \makecell{\textbf{Balance}\\\textbf{improv.} \textbf{(\%)}} & \makecell{\textbf{Total}\\\textbf{tokens}} & \makecell{\textbf{Tokens/}\\\textbf{iteration}} \\
\midrule
\multirow{2}{*}{1} & \textsc{MetaFORGE} & 9 & 3 & -1.006 & 100.00 & 11.75 & 99.46 & 76,718 & 8,524.22 \\
 & \textsc{TextGrad} & 9 & 1 & -6.109 & 0.00 & 0.00 & 0.00 & 135,284 & 15,031.56 \\
\addlinespace[1.5pt]
\multirow{2}{*}{2} & \textsc{MetaFORGE} & 10 & 4 & -1.060 & 100.00 & 95.31 & 16.61 & 85,668 & 8,566.80 \\
 & \textsc{TextGrad} & 12 & 4 & -6.119 & 44.44 & 91.51 & 6.12 & 180,756 & 15,063.00 \\
\addlinespace[1.5pt]
\multirow{2}{*}{3} & \textsc{MetaFORGE} & 11 & 3 & -1.041 & 100.00 & -7.96 & -25.54 & 70,794 & 6,435.82 \\
 & \textsc{TextGrad} & 16 & 8 & -9.071 & 11.11 & 23.66 & 75.02 & 257,762 & 16,110.12 \\
\addlinespace[1.5pt]
\multirow{2}{*}{4} & \textsc{MetaFORGE} & 12 & 6 & -1.076 & 100.00 & 21.56 & 64.68 & 108,961 & 9,080.08 \\
 & \textsc{TextGrad} & 9 & 1 & -6.071 & 0.00 & 0.00 & 0.00 & 152,952 & 16,994.67 \\
\addlinespace[1.5pt]
\multirow{2}{*}{5} & \textsc{MetaFORGE} & 10 & 4 & -1.022 & 100.00 & -12.40 & 73.14 & 76,476 & 7,647.60 \\
 & \textsc{TextGrad} & 9 & 1 & -7.085 & 0.00 & 0.00 & 0.00 & 138,342 & 15,371.33 \\
\midrule
\multirow{2}{*}{\makecell{\textbf{Mean}\\$\boldsymbol{\pm}$ \textbf{SD}}} & \textsc{MetaFORGE} & $10.4 \pm 1.1$ & $4.0 \pm 1.2$ & $-1.041 \pm 0.028$ & $100.00 \pm 0.00$ & $21.65 \pm 43.47$ & $45.67 \pm 49.81$ & $83,723 \pm 15,079$ & $8,050.91 \pm 1,039.30$ \\
& \textsc{TextGrad} & $11.0 \pm 3.1$ & $3.0 \pm 3.1$ & $-6.891 \pm 1.291$ & $11.11 \pm 19.25$ & $23.03 \pm 39.63$ & $16.23 \pm 32.97$ & $173,019 \pm 50,671$ & $15,714.14 \pm 837.25$ \\
\bottomrule
\end{tabular}
\end{adjustbox}
\caption{Per-trial MPI dot-product results for the medium workload.}
\label{tab:dot_medium_trial_summary}
\end{table*}

\begin{table*}[t]
\centering

\scriptsize
\setlength{\tabcolsep}{2.6pt}
\begin{adjustbox}{max width=\textwidth}
\begin{tabular}{cccccccccc}
\toprule
\textbf{Trial} & \textbf{Method} & \textbf{Iter.} & \makecell{\textbf{Best}\\\textbf{iter.}} & \makecell{\textbf{Best}\\\textbf{reward}} & \makecell{\textbf{Static}\\\textbf{improv.} \textbf{(\%)}} & \makecell{\textbf{Runtime}\\\textbf{improv.} \textbf{(\%)}} & \makecell{\textbf{Balance}\\\textbf{improv.} \textbf{(\%)}} & \makecell{\textbf{Total}\\\textbf{tokens}} & \makecell{\textbf{Tokens/}\\\textbf{iteration}} \\
\midrule
\multirow{2}{*}{1} & \textsc{MetaFORGE} & 18 & 12 & -1.009 & 100.00 & 94.19 & 97.88 & 170,530 & 9,473.89 \\
 & \textsc{TextGrad} & 9 & 1 & -6.024 & 0.00 & 0.00 & 0.00 & 118,939 & 13,215.44 \\
\addlinespace[1.5pt]
\multirow{2}{*}{2} & \textsc{MetaFORGE} & 9 & 3 & -1.042 & 100.00 & -4.23 & -43.91 & 73,766 & 8,196.22 \\
 & \textsc{TextGrad} & 9 & 1 & -10.161 & 0.00 & 0.00 & 0.00 & 113,276 & 12,586.22 \\
\addlinespace[1.5pt]
\multirow{2}{*}{3} & \textsc{MetaFORGE} & 20 & 20 & -1.007 & 100.00 & 38.16 & 99.45 & 187,252 & 9,362.60 \\
 & \textsc{TextGrad} & 9 & 1 & -6.050 & 0.00 & 0.00 & 0.00 & 126,684 & 14,076.00 \\
\addlinespace[1.5pt]
\multirow{2}{*}{4} & \textsc{MetaFORGE} & 20 & 20 & -1.008 & 100.00 & 30.97 & 88.60 & 222,474 & 11,123.70 \\
 & \textsc{TextGrad} & 9 & 1 & -6.038 & 0.00 & 0.00 & 0.00 & 129,981 & 14,442.33 \\
\addlinespace[1.5pt]
\multirow{2}{*}{5} & \textsc{MetaFORGE} & 19 & 13 & -1.122 & 100.00 & 41.31 & 58.38 & 202,290 & 10,646.84 \\
 & \textsc{TextGrad} & 13 & 5 & -8.022 & 22.22 & 94.79 & 79.61 & 184,273 & 14,174.85 \\
\midrule
\multirow{2}{*}{\makecell{\textbf{Mean}\\$\boldsymbol{\pm}$ \textbf{SD}}} & \textsc{MetaFORGE} & $17.2 \pm 4.7$ & $13.6 \pm 7.0$ & $-1.038 \pm 0.049$ & $100.00 \pm 0.00$ & $40.08 \pm 35.28$ & $60.08 \pm 60.43$ & $171,262 \pm 57,765$ & $9,760.65 \pm 1,154.49$ \\
& \textsc{TextGrad} & $9.8 \pm 1.8$ & $1.8 \pm 1.8$ & $-7.259 \pm 1.836$ & $4.44 \pm 9.94$ & $18.96 \pm 42.39$ & $15.92 \pm 35.60$ & $134,631 \pm 28,510$ & $13,698.97 \pm 773.50$ \\
\bottomrule
\end{tabular}
\end{adjustbox}
\caption{Per-trial MPI dot-product results for the large workload.}
\label{tab:dot_large_trial_summary}
\end{table*}

\subsection{Matrix Multiplication}

\begin{figure*}[ht]
    \centering
    \includegraphics[width=1.0\linewidth]{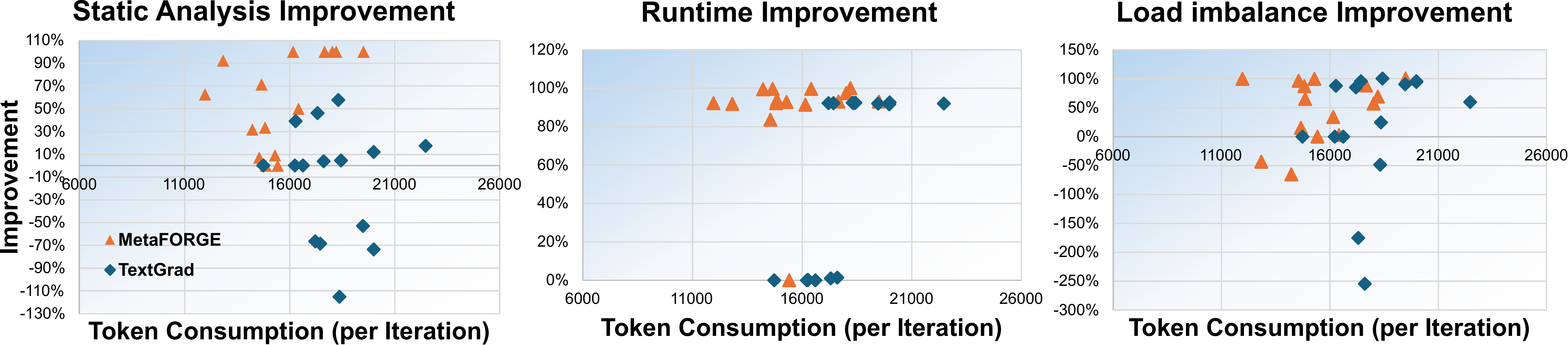}
    \caption{MPI matrix multiplication results comparing MetaFORGE and TextGrad across 15 trials (five trials per workload). MetaFORGE achieves higher code-quality, runtime, and load-balance improvements.}   
    \label{fig:matrix_multiplication}
\end{figure*}

We selected matrix multiplication because matrix-based linear-algebra kernels are widely used in scientific computing and machine-learning workloads, including privacy-preserving computation. For example, this paper~\cite{gao2026efficient} study efficient homomorphic-encryption support for sparse matrix--vector multiplication. Dense matrix multiplication also provides a strong MPI benchmark because efficient execution requires numerical correctness, communication-aware decomposition, and balanced work distribution. We evaluated small, medium, and large workloads using non-square matrix dimensions that were not evenly divisible across the four-node execution environment, preventing trivial uniform partitioning. Figure~\ref{fig:matrix_multiplication} summarizes the effectiveness--cost tradeoff, while Table~\ref{tab:overall_results} reports the aggregate results across matrix multiplication workloads.

MetaFORGE outperforms TextGrad across all aggregate metrics. It improves static-analysis quality by 57.16\%, whereas TextGrad shows a 13.14\% degradation, indicating that MetaFORGE generates cleaner MPI implementations according to static analysis. MetaFORGE also achieves stronger runtime improvement, 87.99\% compared with 55.46\% for TextGrad, and improves load-balance utilization by 47.04\% compared with 16.91\%. These gains are achieved with lower per-iteration token consumption, 15,641.22 tokens for MetaFORGE versus 18,031.71 for TextGrad. Together, Figure~\ref{fig:matrix_multiplication} and Table~\ref{tab:overall_results} show that MetaFORGE provides a stronger effectiveness--cost tradeoff for MPI matrix multiplication.

\subsubsection{Matrix-Multiplication Trial-Level Results}
\label{appendix:matmul_trial_level_results}

Tables~\ref{tab:matmul_small_trial_summary}--%
\ref{tab:matmul_large_trial_summary} report the five independent trials for
each matrix-multiplication workload. Static, runtime, and imbalance columns
report percentage improvement over the corresponding no-skill zero-shot
baseline; higher values are better, and negative values denote regression.
The ``Exec. iter.'' column counts iterations for which the experimental record
contains a measured runtime. The best iteration and reward identify the
reward-maximizing candidate observed within each refinement trajectory. Total
tokens are summed across the complete trajectory, whereas tokens per iteration
are taken from the run-level summary. The final two rows report the
arithmetic mean and sample standard deviation of the five trial-level values.

\begin{table*}[t]
\centering

\scriptsize
\setlength{\tabcolsep}{2.2pt}
\begin{adjustbox}{max width=\textwidth}
\begin{tabular}{ccccccccccc}
\toprule
\textbf{Trial} & \textbf{Method} & \makecell{\textbf{Iter.}} & \makecell{\textbf{Exec.}\\\textbf{iter.}} & \makecell{\textbf{Best}\\\textbf{iter.}} & \makecell{\textbf{Best}\\\textbf{reward}} & \makecell{\textbf{Static}\\\textbf{improv.} \textbf{(\%)}} & \makecell{\textbf{Runtime}\\\textbf{improv.} \textbf{(\%)}} & \makecell{\textbf{Balance}\\\textbf{improv.} \textbf{(\%)}} & \makecell{\textbf{Total}\\\textbf{tokens}} & \makecell{\textbf{Tokens/}\\\textbf{iteration}} \\
\midrule
\multirow{2}{*}{1} & \textsc{MetaFORGE} & 12 & 4 & 6 & -18.581 & 62.50 & 92.24 & 99.92 & 143,595 & 11,966.25 \\
 & \textsc{TextGrad} & 9 & 8 & 1 & -20.634 & 0.00 & 0.00 & 0.00 & 132,627 & 14,736.33 \\
\addlinespace[1.5pt]
\multirow{2}{*}{2} & \textsc{MetaFORGE} & 18 & 17 & 12 & -1.000 & 100.00 & 97.45 & 57.28 & 324,435 & 18,024.17 \\
 & \textsc{TextGrad} & 10 & 10 & 2 & -132.581 & 3.70 & 1.34 & -254.74 & 176,071 & 17,607.10 \\
\addlinespace[1.5pt]
\multirow{2}{*}{3} & \textsc{MetaFORGE} & 20 & 20 & 17 & -9.007 & 100.00 & 92.95 & 88.44 & 353,321 & 17,666.05 \\
 & \textsc{TextGrad} & 20 & 19 & 14 & -32.262 & 12.00 & 91.61 & 94.21 & 399,772 & 19,988.60 \\
\addlinespace[1.5pt]
\multirow{2}{*}{4} & \textsc{MetaFORGE} & 20 & 20 & 19 & -9.002 & 100.00 & 92.99 & 99.99 & 389,984 & 19,499.20 \\
 & \textsc{TextGrad} & 17 & 17 & 9 & -42.036 & -73.68 & 92.59 & 95.55 & 339,845 & 19,990.88 \\
\addlinespace[1.5pt]
\multirow{2}{*}{5} & \textsc{MetaFORGE} & 8 & 8 & 2 & -11.722 & 92.31 & 91.84 & -43.81 & 102,665 & 12,833.12 \\
 & \textsc{TextGrad} & 15 & 14 & 7 & -20.468 & 57.69 & 92.44 & -48.66 & 274,746 & 18,316.40 \\
\midrule
\multirow{2}{*}{\makecell{\textbf{Mean}\\$\boldsymbol{\pm}$ \textbf{SD}}}
& \textsc{MetaFORGE} & $15.6 \pm 5.4$ & $13.8 \pm 7.4$ & $11.2 \pm 7.2$ & $-9.862 \pm 6.314$ & $90.96 \pm 16.26$ & $93.49 \pm 2.26$ & $60.37 \pm 60.80$ & $262,800 \pm 130,405$ & $15,997.76 \pm 3,369.63$ \\
& \textsc{TextGrad} & $14.2 \pm 4.7$ & $13.6 \pm 4.6$ & $6.6 \pm 5.3$ & $-49.596 \pm 47.253$ & $-0.06 \pm 47.21$ & $55.60 \pm 50.15$ & $-22.73 \pm 143.77$ & $264,612 \pm 111,007$ & $18,127.86 \pm 2,164.62$ \\
\bottomrule
\end{tabular}
\end{adjustbox}
\caption{Per-trial matrix-multiplication results for the small workload.}
\label{tab:matmul_small_trial_summary}
\end{table*}

\begin{table*}[t]
\centering

\scriptsize
\setlength{\tabcolsep}{2.2pt}
\begin{adjustbox}{max width=\textwidth}
\begin{tabular}{ccccccccccc}
\toprule
\textbf{Trial} & \textbf{Method} & \makecell{\textbf{Iter.}} & \makecell{\textbf{Exec.}\\\textbf{iter.}} & \makecell{\textbf{Best}\\\textbf{iter.}} & \makecell{\textbf{Best}\\\textbf{reward}} & \makecell{\textbf{Static}\\\textbf{improv.} \textbf{(\%)}} & \makecell{\textbf{Runtime}\\\textbf{improv.} \textbf{(\%)}} & \makecell{\textbf{Balance}\\\textbf{improv.} \textbf{(\%)}} & \makecell{\textbf{Total}\\\textbf{tokens}} & \makecell{\textbf{Tokens/}\\\textbf{iteration}} \\
\midrule
\multirow{2}{*}{1} & \textsc{MetaFORGE} & 15 & 14 & 9 & -32.672 & 33.33 & 92.21 & 87.00 & 222,329 & 14,821.93 \\
 & \textsc{TextGrad} & 9 & 8 & 1 & -228.006 & 0.00 & 0.00 & 0.00 & 149,584 & 16,620.44 \\
\addlinespace[1.5pt]
\multirow{2}{*}{2} & \textsc{MetaFORGE} & 9 & 9 & 3 & -54.924 & 7.14 & 83.50 & 96.25 & 130,999 & 14,555.44 \\
 & \textsc{TextGrad} & 11 & 10 & 3 & -49.596 & -68.42 & 92.15 & 95.64 & 191,867 & 17,442.45 \\
\addlinespace[1.5pt]
\multirow{2}{*}{3} & \textsc{MetaFORGE} & 15 & 15 & 9 & -26.961 & 9.09 & 92.76 & 99.89 & 229,347 & 15,289.80 \\
 & \textsc{TextGrad} & 9 & 8 & 1 & -52.536 & 0.00 & 0.00 & 0.00 & 146,097 & 16,233.00 \\
\addlinespace[1.5pt]
\multirow{2}{*}{4} & \textsc{MetaFORGE} & 7 & 7 & 1 & -39.490 & 0.00 & 0.00 & 0.00 & 107,983 & 15,426.14 \\
 & \textsc{TextGrad} & 20 & 18 & 19 & -40.718 & 17.24 & 92.00 & 59.65 & 449,399 & 22,469.95 \\
\addlinespace[1.5pt]
\multirow{2}{*}{5} & \textsc{MetaFORGE} & 20 & 19 & 20 & -19.478 & 100.00 & 91.30 & 33.93 & 323,138 & 16,156.90 \\
 & \textsc{TextGrad} & 10 & 6 & 2 & -217.332 & 38.89 & 0.10 & 87.48 & 162,808 & 16,280.80 \\
\midrule
\multirow{2}{*}{\makecell{\textbf{Mean}\\$\boldsymbol{\pm}$ \textbf{SD}}}
& \textsc{MetaFORGE} & $13.2 \pm 5.2$ & $12.8 \pm 4.8$ & $8.4 \pm 7.4$ & $-34.705 \pm 13.488$ & $29.91 \pm 41.14$ & $71.95 \pm 40.40$ & $63.41 \pm 44.31$ & $202,759 \pm 86,187$ & $15,250.04 \pm 616.59$ \\
& \textsc{TextGrad} & $11.8 \pm 4.7$ & $10.0 \pm 4.7$ & $5.2 \pm 7.8$ & $-117.638 \pm 96.053$ & $-2.46 \pm 40.19$ & $36.85 \pm 50.41$ & $48.55 \pm 46.29$ & $219,951 \pm 129,524$ & $17,809.33 \pm 2,650.03$ \\
\bottomrule
\end{tabular}
\end{adjustbox}
\caption{Per-trial matrix-multiplication results for the medium workload.}
\label{tab:matmul_medium_trial_summary}
\end{table*}

\begin{table*}[t]
\centering

\scriptsize
\setlength{\tabcolsep}{2.2pt}
\begin{adjustbox}{max width=\textwidth}
\begin{tabular}{ccccccccccc}
\toprule
\textbf{Trial} & \textbf{Method} & \makecell{\textbf{Iter.}} & \makecell{\textbf{Exec.}\\\textbf{iter.}} & \makecell{\textbf{Best}\\\textbf{iter.}} & \makecell{\textbf{Best}\\\textbf{reward}} & \makecell{\textbf{Static}\\\textbf{improv.} \textbf{(\%)}} & \makecell{\textbf{Runtime}\\\textbf{improv.} \textbf{(\%)}} & \makecell{\textbf{Balance}\\\textbf{improv.} \textbf{(\%)}} & \makecell{\textbf{Total}\\\textbf{tokens}} & \makecell{\textbf{Tokens/}\\\textbf{iteration}} \\
\midrule
\multirow{2}{*}{1} & \textsc{MetaFORGE} & 9 & 9 & 3 & -8.713 & 71.43 & 99.68 & 15.40 & 131,976 & 14,664.00 \\
 & \textsc{TextGrad} & 16 & 14 & 8 & -52.336 & 45.95 & 0.97 & -175.08 & 276,971 & 17,310.69 \\
\addlinespace[1.5pt]
\multirow{2}{*}{2} & \textsc{MetaFORGE} & 20 & 20 & 20 & -1.021 & 100.00 & 100.00 & 69.32 & 364,294 & 18,214.70 \\
 & \textsc{TextGrad} & 18 & 18 & 10 & -67.685 & -66.67 & 92.13 & 85.04 & 309,841 & 17,213.39 \\
\addlinespace[1.5pt]
\multirow{2}{*}{3} & \textsc{MetaFORGE} & 9 & 8 & 3 & -20.216 & 50.00 & 99.54 & 2.37 & 147,783 & 16,420.33 \\
 & \textsc{TextGrad} & 16 & 16 & 8 & -59.502 & -115.38 & 92.22 & 24.48 & 293,795 & 18,362.19 \\
\addlinespace[1.5pt]
\multirow{2}{*}{4} & \textsc{MetaFORGE} & 9 & 9 & 3 & -39.804 & 0.00 & 93.86 & 65.23 & 133,718 & 14,857.56 \\
 & \textsc{TextGrad} & 16 & 16 & 8 & -52.789 & 4.55 & 92.33 & 100.00 & 294,897 & 18,431.06 \\
\addlinespace[1.5pt]
\multirow{2}{*}{5} & \textsc{MetaFORGE} & 11 & 10 & 5 & -16.347 & 31.58 & 99.50 & -65.64 & 156,450 & 14,222.73 \\
 & \textsc{TextGrad} & 18 & 16 & 10 & -57.577 & -52.94 & 92.06 & 90.01 & 350,503 & 19,472.39 \\
\midrule
\multirow{2}{*}{\makecell{\textbf{Mean}\\$\boldsymbol{\pm}$ \textbf{SD}}}
& \textsc{MetaFORGE} & $11.6 \pm 4.8$ & $11.2 \pm 5.0$ & $6.8 \pm 7.4$ & $-17.220 \pm 14.617$ & $50.60 \pm 38.07$ & $98.52 \pm 2.61$ & $17.34 \pm 55.02$ & $186,844 \pm 99,713$ & $15,675.86 \pm 1,643.55$ \\
& \textsc{TextGrad} & $16.8 \pm 1.1$ & $16.0 \pm 1.4$ & $8.8 \pm 1.1$ & $-57.978 \pm 6.234$ & $-36.90 \pm 63.02$ & $73.94 \pm 40.79$ & $24.89 \pm 115.64$ & $305,201 \pm 27,870$ & $18,157.94 \pm 929.26$ \\
\bottomrule
\end{tabular}
\end{adjustbox}
\caption{Per-trial matrix-multiplication results for the large workload.}
\label{tab:matmul_large_trial_summary}
\end{table*}

\subsection{Mandelbrot Sets}
\begin{figure*}[ht]
    \centering
    \includegraphics[width=1.0\linewidth]{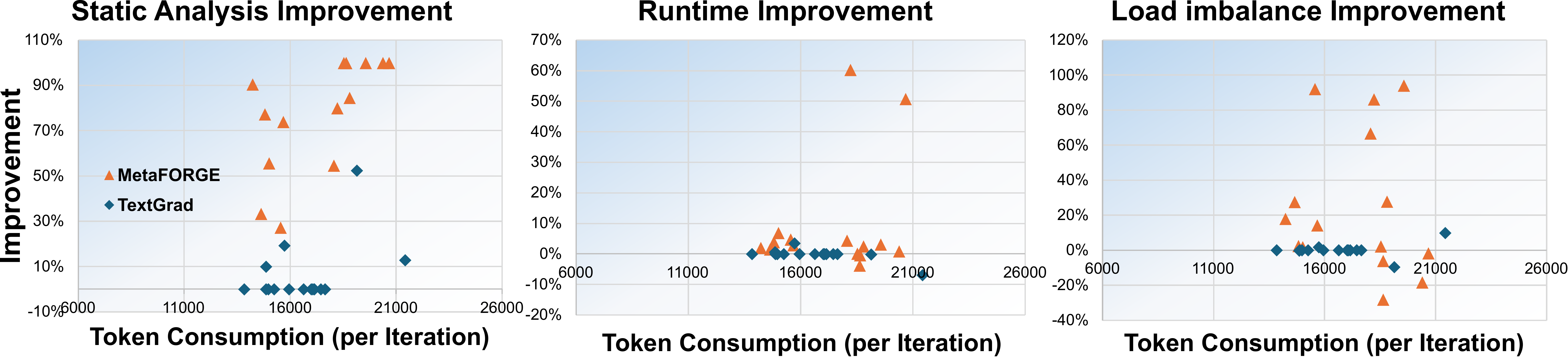}
    \caption{Mandelbrot-set generation results comparing \textsc{MetaFORGE} and \textsc{TextGrad} across 15 trials (five trials per workload). \textsc{MetaFORGE} achieves stronger static-analysis warning reduction, runtime improvement, and load-balance improvement, with modestly higher aggregate per-iteration token consumption. }

    \label{fig_Mandelbrotsets}
\end{figure*}

We selected Mandelbrot set generation because it is an irregular parallel workload in which each pixel can be computed independently, but the required iteration count varies across the complex plane. Points near the set boundary require more computation than points that diverge quickly, making static row- or block-based decomposition prone to rank-level imbalance. The task also requires correct reconstruction of a high-resolution color image from distributed pixel computations, making indexing, result gathering, and workload distribution central to correctness and performance. Figure~\ref{fig_Mandelbrotsets} summarizes the effectiveness--cost tradeoff, while Table~\ref{tab:overall_results} reports the aggregate Mandelbrot results.

MetaFORGE outperforms TextGrad across all aggregate optimization metrics. It improves static-analysis quality by 78.47\%, compared with 6.30\% for TextGrad, and improves runtime by 9.32\%, while TextGrad shows a 0.19\% runtime degradation. MetaFORGE also improves load-balance by 25.29\%, compared with only 0.15\% for TextGrad. Although MetaFORGE consumes modestly more tokens per iteration, 17,435.36 versus 16,604.08 for TextGrad, the additional cost yields larger gains in code quality, runtime performance, and utilization. Together, Figure~\ref{fig_Mandelbrotsets} and Table~\ref{tab:overall_results} show that MetaFORGE provides stronger optimization gains for MPI Mandelbrot set generation.

\subsubsection{Mandelbrot-Set Trial-Level Results}
\label{appendix:mandelbrot_trial_level_results}

Tables~\ref{tab:mandelbrot_small_trial_summary}--%
\ref{tab:mandelbrot_large_trial_summary} report the five independent trials for
each Mandelbrot-set workload. Static, runtime, and imbalance columns report
percentage improvement over the corresponding no-skill zero-shot baseline;
higher values are better, and negative values denote regression. The
``Exec. iter.'' column counts iterations for which the experimental record
contains a measured runtime. The best iteration and reward identify the
reward-maximizing candidate observed within each refinement trajectory. Total
tokens are summed across the complete trajectory. The final two rows report the
arithmetic mean and sample standard deviation of the five trial-level values.
Zero-valued and negative improvements are preserved exactly as recorded rather
than being treated as missing observations.

\begin{table*}[t]
\centering

\scriptsize
\setlength{\tabcolsep}{2.2pt}
\begin{adjustbox}{max width=\textwidth}
\begin{tabular}{ccccccccccc}
\toprule
\textbf{Trial} & \textbf{Method} & \makecell{\textbf{Iter.}} & \makecell{\textbf{Exec.}\\\textbf{iter.}} & \makecell{\textbf{Best}\\\textbf{iter.}} & \makecell{\textbf{Best}\\\textbf{reward}} & \makecell{\textbf{Static}\\\textbf{improv.} \textbf{(\%)}} & \makecell{\textbf{Runtime}\\\textbf{improv.} \textbf{(\%)}} & \makecell{\textbf{Balance}\\\textbf{improv.} \textbf{(\%)}} & \makecell{\textbf{Total}\\\textbf{tokens}} & \makecell{\textbf{Tokens/}\\\textbf{iteration}} \\
\midrule
\multirow{2}{*}{1} & \textsc{MetaFORGE} & 13 & 12 & 7 & -25.181 & 73.91 & 2.84 & 14.32 & 203,819 & 15,678.38 \\
 & \textsc{TextGrad} & 8 & 1 & 3 & -62.261 & 0.00 & 0.00 & 0.00 & 110,732 & 13,841.50 \\
\addlinespace[1.5pt]
\multirow{2}{*}{2} & \textsc{MetaFORGE} & 17 & 16 & 11 & -22.589 & 90.48 & 2.00 & 17.94 & 242,083 & 14,240.18 \\
 & \textsc{TextGrad} & 6 & 4 & 1 & -41.816 & 0.00 & 0.00 & 0.00 & 102,437 & 17,072.83 \\
\addlinespace[1.5pt]
\multirow{2}{*}{3} & \textsc{MetaFORGE} & 11 & 10 & 5 & -28.757 & 33.33 & 1.51 & 27.61 & 161,096 & 14,645.09 \\
 & \textsc{TextGrad} & 18 & 14 & 12 & -71.217 & 12.90 & -6.84 & 10.00 & 385,691 & 21,427.28 \\
\addlinespace[1.5pt]
\multirow{2}{*}{4} & \textsc{MetaFORGE} & 20 & 19 & 19 & -20.572 & 100.00 & -0.41 & -28.09 & 372,906 & 18,645.30 \\
 & \textsc{TextGrad} & 19 & 4 & 13 & -57.602 & 10.00 & 0.56 & -0.10 & 282,484 & 14,867.58 \\
\addlinespace[1.5pt]
\multirow{2}{*}{5} & \textsc{MetaFORGE} & 10 & 10 & 4 & -39.211 & 27.27 & 4.81 & 92.06 & 155,598 & 15,559.80 \\
 & \textsc{TextGrad} & 6 & 3 & 1 & -42.133 & 0.00 & 0.00 & 0.00 & 104,734 & 17,455.67 \\
\midrule
\multirow{2}{*}{\makecell{\textbf{Mean}\\$\boldsymbol{\pm}$ \textbf{SD}}}
& \textsc{MetaFORGE} & $14.2 \pm 4.2$ & $13.4 \pm 4.0$ & $9.2 \pm 6.1$ & $-27.262 \pm 7.348$ & $65.00 \pm 33.09$ & $2.15 \pm 1.91$ & $24.77 \pm 43.26$ & $227,100 \pm 88,734$ & $15,753.75 \pm 1,726.59$ \\
& \textsc{TextGrad} & $11.4 \pm 6.5$ & $5.2 \pm 5.1$ & $6.0 \pm 6.0$ & $-55.006 \pm 12.864$ & $4.58 \pm 6.36$ & $-1.26 \pm 3.13$ & $1.98 \pm 4.48$ & $197,216 \pm 130,201$ & $16,932.97 \pm 2,928.93$ \\
\bottomrule
\end{tabular}
\end{adjustbox}
\caption{Per-trial Mandelbrot-set results for the small workload.}
\label{tab:mandelbrot_small_trial_summary}
\end{table*}

\begin{table*}[t]
\centering

\scriptsize
\setlength{\tabcolsep}{2.2pt}
\begin{adjustbox}{max width=\textwidth}
\begin{tabular}{ccccccccccc}
\toprule
\textbf{Trial} & \textbf{Method} & \makecell{\textbf{Iter.}} & \makecell{\textbf{Exec.}\\\textbf{iter.}} & \makecell{\textbf{Best}\\\textbf{iter.}} & \makecell{\textbf{Best}\\\textbf{reward}} & \makecell{\textbf{Static}\\\textbf{improv.} \textbf{(\%)}} & \makecell{\textbf{Runtime}\\\textbf{improv.} \textbf{(\%)}} & \makecell{\textbf{Balance}\\\textbf{improv.} \textbf{(\%)}} & \makecell{\textbf{Total}\\\textbf{tokens}} & \makecell{\textbf{Tokens/}\\\textbf{iteration}} \\
\midrule
\multirow{2}{*}{1} & \textsc{MetaFORGE} & 12 & 11 & 4 & -11.328 & 80.00 & 60.24 & 86.18 & 218,732 & 18,227.67 \\
 & \textsc{TextGrad} & 20 & 16 & 18 & -31.637 & 52.38 & -0.07 & -9.49 & 382,966 & 19,148.30 \\
\addlinespace[1.5pt]
\multirow{2}{*}{2} & \textsc{MetaFORGE} & 18 & 18 & 10 & -18.957 & 55.56 & 6.85 & 1.72 & 270,328 & 15,018.22 \\
 & \textsc{TextGrad} & 8 & 4 & 1 & -21.224 & 0.00 & 0.00 & 0.00 & 133,101 & 16,637.62 \\
\addlinespace[1.5pt]
\multirow{2}{*}{3} & \textsc{MetaFORGE} & 13 & 12 & 5 & -16.544 & 77.27 & 3.88 & 2.50 & 192,676 & 14,821.23 \\
 & \textsc{TextGrad} & 8 & 5 & 1 & -32.459 & 0.00 & 0.00 & 0.00 & 127,596 & 15,949.50 \\
\addlinespace[1.5pt]
\multirow{2}{*}{4} & \textsc{MetaFORGE} & 15 & 13 & 7 & -12.035 & 100.00 & -3.74 & -6.05 & 279,616 & 18,641.07 \\
 & \textsc{TextGrad} & 10 & 4 & 3 & -34.587 & 0.00 & 0.00 & 0.00 & 149,736 & 14,973.60 \\
\addlinespace[1.5pt]
\multirow{2}{*}{5} & \textsc{MetaFORGE} & 14 & 9 & 6 & -26.242 & 54.55 & 4.34 & 66.67 & 252,908 & 18,064.86 \\
 & \textsc{TextGrad} & 8 & 2 & 1 & -33.633 & 0.00 & 0.00 & 0.00 & 119,035 & 14,879.38 \\
\midrule
\multirow{2}{*}{\makecell{\textbf{Mean}\\$\boldsymbol{\pm}$ \textbf{SD}}}
& \textsc{MetaFORGE} & $14.4 \pm 2.3$ & $12.6 \pm 3.4$ & $6.4 \pm 2.3$ & $-17.021 \pm 6.047$ & $73.47 \pm 18.97$ & $14.31 \pm 25.98$ & $30.20 \pm 42.88$ & $242,852 \pm 36,425$ & $16,954.61 \pm 1,870.72$ \\
& \textsc{TextGrad} & $10.8 \pm 5.2$ & $6.2 \pm 5.6$ & $4.8 \pm 7.4$ & $-30.708 \pm 5.419$ & $10.48 \pm 23.43$ & $-0.01 \pm 0.03$ & $-1.90 \pm 4.24$ & $182,487 \pm 112,631$ & $16,317.68 \pm 1,741.09$ \\
\bottomrule
\end{tabular}
\end{adjustbox}
\caption{Per-trial Mandelbrot-set results for the medium workload.}
\label{tab:mandelbrot_medium_trial_summary}
\end{table*}

\begin{table*}[t]
\centering

\scriptsize
\setlength{\tabcolsep}{2.2pt}
\begin{adjustbox}{max width=\textwidth}
\begin{tabular}{ccccccccccc}
\toprule
\textbf{Trial} & \textbf{Method} & \makecell{\textbf{Iter.}} & \makecell{\textbf{Exec.}\\\textbf{iter.}} & \makecell{\textbf{Best}\\\textbf{iter.}} & \makecell{\textbf{Best}\\\textbf{reward}} & \makecell{\textbf{Static}\\\textbf{improv.} \textbf{(\%)}} & \makecell{\textbf{Runtime}\\\textbf{improv.} \textbf{(\%)}} & \makecell{\textbf{Balance}\\\textbf{improv.} \textbf{(\%)}} & \makecell{\textbf{Total}\\\textbf{tokens}} & \makecell{\textbf{Tokens/}\\\textbf{iteration}} \\
\midrule
\multirow{2}{*}{1} & \textsc{MetaFORGE} & 20 & 18 & 15 & -21.537 & 100.00 & 0.83 & -18.18 & 407,872 & 20,393.60 \\
 & \textsc{TextGrad} & 10 & 6 & 3 & -44.490 & 0.00 & 0.00 & 0.00 & 170,111 & 17,011.10 \\
\addlinespace[1.5pt]
\multirow{2}{*}{2} & \textsc{MetaFORGE} & 20 & 19 & 20 & -21.091 & 100.00 & 3.10 & 94.00 & 391,483 & 19,574.15 \\
 & \textsc{TextGrad} & 14 & 5 & 6 & -44.473 & 19.23 & 3.52 & 1.78 & 220,271 & 15,733.64 \\
\addlinespace[1.5pt]
\multirow{2}{*}{3} & \textsc{MetaFORGE} & 20 & 19 & 17 & -11.024 & 100.00 & 50.77 & -1.58 & 413,642 & 20,682.10 \\
 & \textsc{TextGrad} & 10 & 0 & 2 & -10010.000 & 0.00 & 0.00 & 0.00 & 152,592 & 15,259.20 \\
\addlinespace[1.5pt]
\multirow{2}{*}{4} & \textsc{MetaFORGE} & 17 & 16 & 9 & -25.501 & 84.62 & 2.56 & 27.94 & 319,676 & 18,804.47 \\
 & \textsc{TextGrad} & 8 & 4 & 1 & -40.385 & 0.00 & 0.00 & 0.00 & 141,216 & 17,652.00 \\
\addlinespace[1.5pt]
\multirow{2}{*}{5} & \textsc{MetaFORGE} & 20 & 19 & 19 & -21.440 & 100.00 & 0.18 & 2.30 & 370,686 & 18,534.30 \\
 & \textsc{TextGrad} & 9 & 3 & 1 & -43.604 & 0.00 & 0.00 & 0.00 & 154,368 & 17,152.00 \\
\midrule
\multirow{2}{*}{\makecell{\textbf{Mean}\\$\boldsymbol{\pm}$ \textbf{SD}}}
& \textsc{MetaFORGE} & $19.4 \pm 1.3$ & $18.2 \pm 1.3$ & $16.0 \pm 4.4$ & $-20.119 \pm 5.394$ & $96.92 \pm 6.88$ & $11.49 \pm 21.99$ & $20.90 \pm 44.08$ & $380,672 \pm 37,960$ & $19,597.72 \pm 944.72$ \\
& \textsc{TextGrad} & $10.2 \pm 2.3$ & $3.6 \pm 2.3$ & $2.6 \pm 2.1$ & $-2,036.590 \pm 4,457.272$ & $3.85 \pm 8.60$ & $0.70 \pm 1.57$ & $0.36 \pm 0.79$ & $167,712 \pm 31,132$ & $16,561.59 \pm 1,015.06$ \\
\bottomrule
\end{tabular}
\end{adjustbox}
\caption{Per-trial Mandelbrot-set results for the large workload.}
\label{tab:mandelbrot_large_trial_summary}
\end{table*}

\subsection{Heat Transfer Using Partial Differential Equations}
We selected the heat-transfer PDE task because it stresses numerical correctness, domain decomposition, halo exchange, synchronization, and communication efficiency in MPI C. The task requires partitioning a computational grid across ranks, iteratively updating local subdomains, and exchanging boundary values with neighboring ranks. Because stencil solvers progress at the rate of the slowest rank, inefficient decomposition or synchronization can reduce performance or compromise correctness. In this task, Codex was directed to solve the heat-transfer PDE using a Poisson/Laplace iterative method with Jacobi relaxation and to visualize the resulting heat field.

\begin{figure*}[h]
    \centering
    \includegraphics[width=1.0\linewidth]{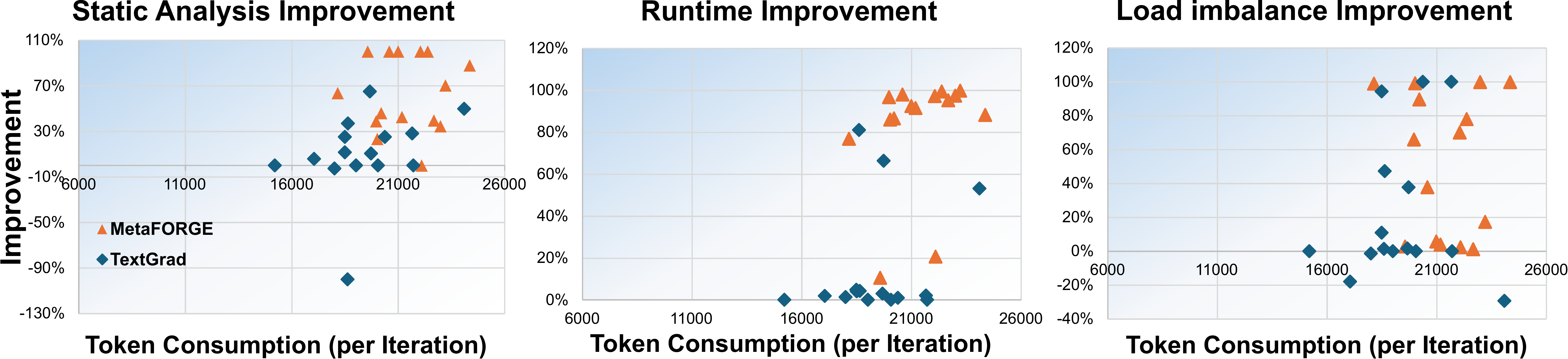}
    \caption{Heat-transfer PDE results comparing MetaFORGE and TextGrad. MetaFORGE achieves stronger code-quality, runtime, and load-balance improvements, with modestly higher token consumption across 15 trials (five trials per workload).}
    \label{fig:heattransfer_PDE}
\end{figure*}

Figure~\ref{fig:heattransfer_PDE} summarizes the effectiveness--cost tradeoff, while Table~\ref{tab:overall_results} reports the aggregate heat-transfer results. MetaFORGE outperforms TextGrad across all optimization metrics, improving static-analysis quality by 63.08\% compared with 10.38\%, runtime by 82.46\% compared with 14.87\%, and load-balance utilization by 51.41\% compared with 22.94\%. Although MetaFORGE consumes more tokens per iteration, 21,350.48 versus 19,375.34 for TextGrad, the modest additional cost yields much larger and more consistent gains in code quality, runtime performance, and utilization.

\subsubsection{Heat-Transfer PDE Trial-Level Results}
\label{appendix:heattransfer_trial_level_results}

Tables~\ref{tab:heattransfer_small_trial_summary}--%
\ref{tab:heattransfer_large_trial_summary} report the five independent trials
for each heat-transfer PDE workload. Static, runtime, and imbalance columns
report percentage improvement over the corresponding no-skill zero-shot
baseline; higher values are better, and negative values denote regression.
The ``Exec. iter.'' column counts iterations for which the experimental record
contains a measured runtime. The best iteration and reward identify the
reward-maximizing candidate observed within each refinement trajectory. Total
tokens are summed across the complete trajectory. The final two rows report the
arithmetic mean and sample standard deviation of the five trial-level values.

\begin{table*}[h]
\centering

\scriptsize
\setlength{\tabcolsep}{2.2pt}
\begin{adjustbox}{max width=\textwidth}
\begin{tabular}{ccccccccccc}
\toprule
\textbf{Trial} & \textbf{Method} & \makecell{\textbf{Iter.}} & \makecell{\textbf{Exec.}\\\textbf{iter.}} & \makecell{\textbf{Best}\\\textbf{iter.}} & \makecell{\textbf{Best}\\\textbf{reward}} & \makecell{\textbf{Static}\\\textbf{improv.} \textbf{(\%)}} & \makecell{\textbf{Runtime}\\\textbf{improv.} \textbf{(\%)}} & \makecell{\textbf{Balance}\\\textbf{improv.} \textbf{(\%)}} & \makecell{\textbf{Total}\\\textbf{tokens}} & \makecell{\textbf{Tokens/}\\\textbf{iteration}} \\
\midrule
\multirow{2}{*}{1} & \textsc{MetaFORGE} & 20 & 17 & 17 & -24.759 & 42.50 & 91.63 & 3.86 & 423,434 & 21,171.70 \\
 & \textsc{TextGrad} & 14 & 7 & 7 & -35.266 & 10.71 & 66.36 & 37.65 & 275,961 & 19,711.50 \\
\addlinespace[1.5pt]
\multirow{2}{*}{2} & \textsc{MetaFORGE} & 14 & 4 & 6 & -35.469 & 0.00 & 20.69 & 2.24 & 309,170 & 22,083.57 \\
 & \textsc{TextGrad} & 9 & 2 & 1 & -42.510 & 0.00 & 0.00 & 0.00 & 136,708 & 15,189.78 \\
\addlinespace[1.5pt]
\multirow{2}{*}{3} & \textsc{MetaFORGE} & 20 & 12 & 18 & -22.268 & 39.39 & 95.15 & 1.13 & 453,188 & 22,659.40 \\
 & \textsc{TextGrad} & 12 & 8 & 4 & -45.201 & 50.00 & 53.24 & -29.38 & 288,915 & 24,076.25 \\
\addlinespace[1.5pt]
\multirow{2}{*}{4} & \textsc{MetaFORGE} & 20 & 16 & 16 & -20.562 & 34.48 & 97.45 & 99.87 & 459,460 & 22,973.00 \\
 & \textsc{TextGrad} & 8 & 6 & 1 & -44.026 & 0.00 & 0.00 & 0.00 & 173,585 & 21,698.12 \\
\addlinespace[1.5pt]
\multirow{2}{*}{5} & \textsc{MetaFORGE} & 20 & 15 & 15 & -24.722 & 23.33 & 85.99 & 99.06 & 400,181 & 20,009.05 \\
 & \textsc{TextGrad} & 8 & 1 & 1 & -45.356 & 0.00 & 0.00 & 0.00 & 160,339 & 20,042.38 \\
\midrule
\multirow{2}{*}{\makecell{\textbf{Mean}\\$\boldsymbol{\pm}$ \textbf{SD}}}
& \textsc{MetaFORGE} & $18.8 \pm 2.7$ & $12.8 \pm 5.3$ & $14.4 \pm 4.8$ & $-25.556 \pm 5.817$ & $27.94 \pm 17.23$ & $78.18 \pm 32.43$ & $41.23 \pm 53.17$ & $409,087 \pm 60,727$ & $21,779.34 \pm 1,203.64$ \\
& \textsc{TextGrad} & $10.2 \pm 2.7$ & $4.8 \pm 3.1$ & $2.8 \pm 2.7$ & $-42.472 \pm 4.186$ & $12.14 \pm 21.67$ & $23.92 \pm 33.08$ & $1.65 \pm 23.81$ & $207,102 \pm 70,179$ & $20,143.61 \pm 3,264.18$ \\
\bottomrule
\end{tabular}
\end{adjustbox}
\caption{Per-trial heat-transfer PDE results for the small workload.}
\label{tab:heattransfer_small_trial_summary}
\end{table*}

\begin{table*}[h]
\centering

\scriptsize
\setlength{\tabcolsep}{2.2pt}
\begin{adjustbox}{max width=\textwidth}
\begin{tabular}{ccccccccccc}
\toprule
\textbf{Trial} & \textbf{Method} & \makecell{\textbf{Iter.}} & \makecell{\textbf{Exec.}\\\textbf{iter.}} & \makecell{\textbf{Best}\\\textbf{iter.}} & \makecell{\textbf{Best}\\\textbf{reward}} & \makecell{\textbf{Static}\\\textbf{improv.} \textbf{(\%)}} & \makecell{\textbf{Runtime}\\\textbf{improv.} \textbf{(\%)}} & \makecell{\textbf{Balance}\\\textbf{improv.} \textbf{(\%)}} & \makecell{\textbf{Total}\\\textbf{tokens}} & \makecell{\textbf{Tokens/}\\\textbf{iteration}} \\
\midrule
\multirow{2}{*}{1} & \textsc{MetaFORGE} & 20 & 16 & 12 & -26.912 & 38.89 & 96.75 & 65.80 & 399,218 & 19,960.90 \\
 & \textsc{TextGrad} & 15 & 4 & 8 & -113.356 & 5.88 & 1.86 & -17.99 & 255,631 & 17,042.07 \\
\addlinespace[1.5pt]
\multirow{2}{*}{2} & \textsc{MetaFORGE} & 20 & 18 & 19 & -1.788 & 100.00 & 92.52 & 5.62 & 419,707 & 20,985.35 \\
 & \textsc{TextGrad} & 17 & 10 & 9 & -108.301 & 65.12 & 2.97 & 1.59 & 334,238 & 19,661.06 \\
\addlinespace[1.5pt]
\multirow{2}{*}{3} & \textsc{MetaFORGE} & 20 & 17 & 17 & -1.409 & 100.00 & 99.50 & 77.85 & 447,410 & 22,370.50 \\
 & \textsc{TextGrad} & 19 & 11 & 12 & -105.967 & 28.21 & 2.10 & 99.98 & 411,419 & 21,653.63 \\
\addlinespace[1.5pt]
\multirow{2}{*}{4} & \textsc{MetaFORGE} & 20 & 16 & 13 & -3.951 & 100.00 & 97.92 & 37.75 & 411,451 & 20,572.55 \\
 & \textsc{TextGrad} & 8 & 3 & 1 & -110.317 & 0.00 & 0.00 & 0.00 & 152,006 & 19,000.75 \\
\addlinespace[1.5pt]
\multirow{2}{*}{5} & \textsc{MetaFORGE} & 20 & 15 & 18 & -3.056 & 100.00 & 97.33 & 70.00 & 440,831 & 22,041.55 \\
 & \textsc{TextGrad} & 20 & 6 & 13 & -108.138 & 25.00 & 1.06 & 99.98 & 407,325 & 20,366.25 \\
\midrule
\multirow{2}{*}{\makecell{\textbf{Mean}\\$\boldsymbol{\pm}$ \textbf{SD}}}
& \textsc{MetaFORGE} & $20.0 \pm 0.0$ & $16.4 \pm 1.1$ & $15.8 \pm 3.1$ & $-7.423 \pm 10.942$ & $87.78 \pm 27.33$ & $96.80 \pm 2.61$ & $51.40 \pm 29.73$ & $423,723 \pm 20,131$ & $21,186.17 \pm 1,006.54$ \\
& \textsc{TextGrad} & $15.8 \pm 4.8$ & $6.8 \pm 3.6$ & $8.6 \pm 4.7$ & $-109.216 \pm 2.779$ & $24.84 \pm 25.54$ & $1.60 \pm 1.12$ & $36.71 \pm 58.27$ & $312,124 \pm 109,819$ & $19,544.75 \pm 1,709.91$ \\
\bottomrule
\end{tabular}
\end{adjustbox}
\caption{Per-trial heat-transfer PDE results for the medium workload.}
\label{tab:heattransfer_medium_trial_summary}
\end{table*}

\begin{table*}[h]
\centering

\scriptsize
\setlength{\tabcolsep}{2.2pt}
\begin{adjustbox}{max width=\textwidth}
\begin{tabular}{ccccccccccc}
\toprule
\textbf{Trial} & \textbf{Method} & \makecell{\textbf{Iter.}} & \makecell{\textbf{Exec.}\\\textbf{iter.}} & \makecell{\textbf{Best}\\\textbf{iter.}} & \makecell{\textbf{Best}\\\textbf{reward}} & \makecell{\textbf{Static}\\\textbf{improv.} \textbf{(\%)}} & \makecell{\textbf{Runtime}\\\textbf{improv.} \textbf{(\%)}} & \makecell{\textbf{Balance}\\\textbf{improv.} \textbf{(\%)}} & \makecell{\textbf{Total}\\\textbf{tokens}} & \makecell{\textbf{Tokens/}\\\textbf{iteration}} \\
\midrule
\multirow{2}{*}{1} & \textsc{MetaFORGE} & 20 & 13 & 20 & -12.003 & 70.27 & 99.82 & 17.23 & 463,985 & 23,199.25 \\
 & \textsc{TextGrad} & 13 & 9 & 5 & -293.298 & 36.84 & 4.26 & 47.24 & 242,217 & 18,632.08 \\
\addlinespace[1.5pt]
\multirow{2}{*}{2} & \textsc{MetaFORGE} & 16 & 9 & 8 & -57.625 & 45.95 & 86.56 & 89.53 & 323,152 & 20,197.00 \\
 & \textsc{TextGrad} & 20 & 9 & 18 & -304.377 & 11.63 & 4.06 & 10.84 & 369,697 & 18,484.85 \\
\addlinespace[1.5pt]
\multirow{2}{*}{3} & \textsc{MetaFORGE} & 19 & 14 & 11 & -54.421 & 63.64 & 76.82 & 98.95 & 344,715 & 18,142.89 \\
 & \textsc{TextGrad} & 19 & 7 & 11 & -296.371 & 25.00 & 4.78 & 94.26 & 351,141 & 18,481.11 \\
\addlinespace[1.5pt]
\multirow{2}{*}{4} & \textsc{MetaFORGE} & 20 & 12 & 20 & -36.601 & 100.00 & 10.49 & 2.59 & 391,036 & 19,551.80 \\
 & \textsc{TextGrad} & 17 & 7 & 9 & -107.932 & -100.00 & 81.00 & 1.29 & 316,198 & 18,599.88 \\
\addlinespace[1.5pt]
\multirow{2}{*}{5} & \textsc{MetaFORGE} & 20 & 14 & 13 & -11.357 & 87.76 & 88.25 & 99.72 & 486,774 & 24,338.70 \\
 & \textsc{TextGrad} & 13 & 4 & 5 & -315.978 & -2.70 & 1.30 & -1.36 & 233,876 & 17,990.46 \\
\midrule
\multirow{2}{*}{\makecell{\textbf{Mean}\\$\boldsymbol{\pm}$ \textbf{SD}}}
& \textsc{MetaFORGE} & $19.0 \pm 1.7$ & $12.4 \pm 2.1$ & $14.4 \pm 5.4$ & $-34.401 \pm 22.236$ & $73.52 \pm 21.05$ & $72.39 \pm 35.56$ & $61.61 \pm 47.64$ & $401,932 \pm 71,846$ & $21,085.93 \pm 2,590.94$ \\
& \textsc{TextGrad} & $16.4 \pm 3.3$ & $7.2 \pm 2.0$ & $9.6 \pm 5.4$ & $-263.591 \pm 87.457$ & $-5.85 \pm 54.67$ & $19.08 \pm 34.64$ & $30.45 \pm 40.62$ & $302,626 \pm 62,073$ & $18,437.68 \pm 258.95$ \\
\bottomrule
\end{tabular}
\end{adjustbox}
\caption{Per-trial heat-transfer PDE results for the large workload.}
\label{tab:heattransfer_large_trial_summary}
\end{table*}

\subsection{Distributed Principal Component Analysis}
\label{appendix:pca_results}
We selected distributed principal component analysis (PCA) because it is a widely used machine-learning method for dimensionality reduction, feature extraction, noise reduction, and exploratory data analysis. Distributed PCA is also a strong MPI benchmark because it combines several performance-sensitive phases: data decomposition, local statistical computation, global aggregation of feature means and covariance contributions, and iterative eigensolver computation. These phases stress dense numerical computation, collective communication, memory layout, synchronization, and load-balance behavior commonly found in distributed machine-learning workloads.

\begin{figure*}[h]
    \centering
    \includegraphics[width=1.0\linewidth]{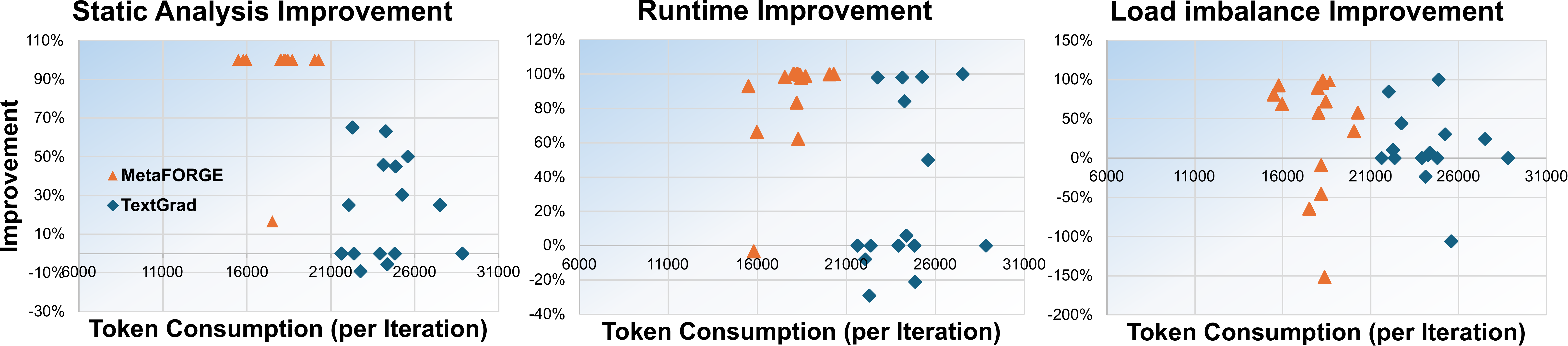}
    \caption{Distributed PCA results comparing MetaFORGE and TextGrad. MetaFORGE achieves higher code-quality, runtime, and load-balance improvements with lower per-iteration token consumption across 15 trials (five trials per workload).}
    \label{fig:princiaplcomponentanalysis}
\end{figure*}

Figure~\ref{fig:princiaplcomponentanalysis} summarizes the effectiveness--cost tradeoff, while Table~\ref{tab:overall_results} reports the aggregate distributed PCA results. MetaFORGE outperforms TextGrad across all optimization metrics. It achieves a 94.44\% static-analysis improvement compared with 22.32\% for TextGrad, indicating a much stronger reduction in static-analysis warnings. MetaFORGE also improves runtime by 86.36\%, more than twice TextGrad's 31.75\% improvement, demonstrating that its skill rewrites produce more efficient distributed PCA implementations.

MetaFORGE further improves load-balance utilization by 38.27\%, compared with 11.64\% for TextGrad, a result particularly important for distributed PCA because aggregation and eigensolver phases can amplify rank-level imbalance and synchronization delays. MetaFORGE also reduces per-iteration token consumption, requiring 17,965.56 tokens compared with 24,298.55 for TextGrad. Together, Figure~\ref{fig:princiaplcomponentanalysis} and Table~\ref{tab:overall_results} show that MetaFORGE provides a stronger effectiveness--cost profile for distributed PCA, producing higher-quality, faster, and better-balanced MPI code while also reducing token consumption.

\subsubsection{Distributed PCA Trial-Level Results}
\label{appendix:pca_trial_level_results}

Tables~\ref{tab:pca_small_trial_summary}--\ref{tab:pca_large_trial_summary}
report the five independent trials for each distributed PCA workload.
Static, runtime, and imbalance columns report percentage improvement over the corresponding no-skill zero-shot baseline; higher values are better, and
negative values denote regression. The best iteration and reward identify the reward-maximizing candidate observed within each refinement trajectory. Total tokens are the sum of the recorded token use across the trajectory.
The final two rows report the arithmetic mean and sample standard deviation of the trial-level values. Because the aggregate percentages 
are computed from pooled raw measurements, they may differ slightly from the arithmetic mean of the trial-level percentages reported here.

\begin{table*}[h]
\centering

\scriptsize
\setlength{\tabcolsep}{2.6pt}
\begin{adjustbox}{max width=\textwidth}
\begin{tabular}{cccccccccc}
\toprule
\textbf{Trial} & \textbf{Method} & \textbf{Iter.} & \makecell{\textbf{Best}\\\textbf{iter.}} & \makecell{\textbf{Best}\\\textbf{reward}} & \makecell{\textbf{Static}\\\textbf{improv.} \textbf{(\%)}} & \makecell{\textbf{Runtime}\\\textbf{improv.} \textbf{(\%)}} & \makecell{\textbf{Balance}\\\textbf{improv.} \textbf{(\%)}} & \makecell{\textbf{Total}\\\textbf{tokens}} & \makecell{\textbf{Tokens/}\\\textbf{iteration}} \\
\midrule
\multirow{2}{*}{1} & \textsc{MetaFORGE} & 19 & 11 & -2.665 & 100.00 & 66.21 & 68.63 & 303,246 & 15,960.32 \\
& \textsc{TextGrad} & 9 & 1 & -22.526 & 0.00 & 0.00 & 0.00 & 215,161 & 23,906.78 \\
\addlinespace[1.5pt]
\multirow{2}{*}{2} & \textsc{MetaFORGE} & 20 & 13 & -2.486 & 100.00 & -3.32 & 92.40 & 315,590 & 15,779.50 \\
& \textsc{TextGrad} & 20 & 13 & -14.850 & 45.00 & -21.23 & 99.76 & 497,195 & 24,859.75 \\
\addlinespace[1.5pt]
\multirow{2}{*}{3} & \textsc{MetaFORGE} & 20 & 16 & -1.341 & 100.00 & 83.24 & -45.98 & 363,650 & 18,182.50 \\
& \textsc{TextGrad} & 20 & 12 & -10.742 & 65.00 & -29.22 & 10.05 & 445,359 & 22,267.95 \\
\addlinespace[1.5pt]
\multirow{2}{*}{4} & \textsc{MetaFORGE} & 20 & 18 & -1.523 & 100.00 & 98.82 & 97.97 & 373,570 & 18,678.50 \\
& \textsc{TextGrad} & 20 & 15 & -9.795 & 50.00 & 49.85 & -106.12 & 511,755 & 25,587.75 \\
\addlinespace[1.5pt]
\multirow{2}{*}{5} & \textsc{MetaFORGE} & 20 & 20 & -1.366 & 100.00 & 99.59 & -152.02 & 367,740 & 18,387.00 \\
& \textsc{TextGrad} & 12 & 4 & -8.209 & 63.16 & 84.21 & 4.08 & 291,072 & 24,256.00 \\
\midrule
\multirow{2}{*}{\makecell{\textbf{Mean}\\$\boldsymbol{\pm}$ \textbf{SD}}}
& \textsc{MetaFORGE} & $19.8 \pm 0.4$ & $15.6 \pm 3.6$ & $-1.876 \pm 0.645$ & $100.00 \pm 0.00$ & $68.91 \pm 42.62$ & $12.20 \pm 108.77$ & $344,759 \pm 32,746$ & $17,397.56 \pm 1,407.10$ \\
& \textsc{TextGrad} & $16.2 \pm 5.3$ & $9.0 \pm 6.1$ & $-13.225 \pm 5.750$ & $44.63 \pm 26.36$ & $16.72 \pm 48.69$ & $1.55 \pm 73.01$ & $392,108 \pm 132,016$ & $24,175.65 \pm 1,242.99$ \\
\bottomrule
\end{tabular}
\end{adjustbox}
\caption{Per-trial distributed PCA results for the small workload.}
\label{tab:pca_small_trial_summary}
\end{table*}

\begin{table*}[h]
\centering

\scriptsize
\setlength{\tabcolsep}{2.6pt}
\begin{adjustbox}{max width=\textwidth}
\begin{tabular}{cccccccccc}
\toprule
\textbf{Trial} & \textbf{Method} & \textbf{Iter.} & \makecell{\textbf{Best}\\\textbf{iter.}} & \makecell{\textbf{Best}\\\textbf{reward}} & \makecell{\textbf{Static}\\\textbf{improv.} \textbf{(\%)}} & \makecell{\textbf{Runtime}\\\textbf{improv.} \textbf{(\%)}} & \makecell{\textbf{Balance}\\\textbf{improv.} \textbf{(\%)}} & \makecell{\textbf{Total}\\\textbf{tokens}} & \makecell{\textbf{Tokens/}\\\textbf{iteration}} \\
\midrule
\multirow{2}{*}{1} & \textsc{MetaFORGE} & 20 & 16 & -1.006 & 100.00 & 99.95 & 89.39 & 359,794 & 17,989.70 \\
& \textsc{TextGrad} & 11 & 3 & -20.848 & 25.00 & -7.88 & 84.89 & 242,432 & 22,039.27 \\
\addlinespace[1.5pt]
\multirow{2}{*}{2} & \textsc{MetaFORGE} & 20 & 20 & -1.034 & 100.00 & 100.00 & 57.63 & 360,697 & 18,034.85 \\
& \textsc{TextGrad} & 13 & 5 & -21.388 & -9.09 & 98.16 & 44.28 & 295,752 & 22,750.15 \\
\addlinespace[1.5pt]
\multirow{2}{*}{3} & \textsc{MetaFORGE} & 20 & 16 & -1.024 & 100.00 & 99.99 & -9.17 & 363,838 & 18,191.90 \\
& \textsc{TextGrad} & 9 & 1 & -20.858 & 0.00 & 0.00 & 0.00 & 194,512 & 21,612.44 \\
\addlinespace[1.5pt]
\multirow{2}{*}{4} & \textsc{MetaFORGE} & 20 & 16 & -1.140 & 100.00 & 99.95 & 57.83 & 405,276 & 20,263.80 \\
& \textsc{TextGrad} & 20 & 15 & -21.509 & 30.43 & 98.55 & 30.30 & 504,622 & 25,231.10 \\
\addlinespace[1.5pt]
\multirow{2}{*}{5} & \textsc{MetaFORGE} & 16 & 8 & -1.114 & 100.00 & 97.84 & 72.38 & 295,122 & 18,445.12 \\
& \textsc{TextGrad} & 10 & 2 & -18.551 & 45.83 & 98.06 & -23.84 & 241,191 & 24,119.10 \\
\midrule
\multirow{2}{*}{\makecell{\textbf{Mean}\\$\boldsymbol{\pm}$ \textbf{SD}}}
& \textsc{MetaFORGE} & $19.2 \pm 1.8$ & $15.2 \pm 4.4$ & $-1.064 \pm 0.059$ & $100.00 \pm 0.00$ & $99.55 \pm 0.95$ & $53.61 \pm 37.44$ & $356,945 \pm 39,458$ & $18,585.08 \pm 955.18$ \\
& \textsc{TextGrad} & $12.6 \pm 4.4$ & $5.2 \pm 5.7$ & $-20.631 \pm 1.201$ & $18.44 \pm 22.56$ & $57.38 \pm 56.05$ & $27.13 \pm 41.76$ & $295,702 \pm 122,164$ & $23,150.41 \pm 1,502.31$ \\
\bottomrule
\end{tabular}
\end{adjustbox}
\caption{Per-trial distributed PCA results for the medium workload.}
\label{tab:pca_medium_trial_summary}
\end{table*}

\begin{table*}[h]
\centering

\scriptsize
\setlength{\tabcolsep}{2.6pt}
\begin{adjustbox}{max width=\textwidth}
\begin{tabular}{cccccccccc}
\toprule
\textbf{Trial} & \textbf{Method} & \textbf{Iter.} & \makecell{\textbf{Best}\\\textbf{iter.}} & \makecell{\textbf{Best}\\\textbf{reward}} & \makecell{\textbf{Static}\\\textbf{improv.} \textbf{(\%)}} & \makecell{\textbf{Runtime}\\\textbf{improv.} \textbf{(\%)}} & \makecell{\textbf{Balance}\\\textbf{improv.} \textbf{(\%)}} & \makecell{\textbf{Total}\\\textbf{tokens}} & \makecell{\textbf{Tokens/}\\\textbf{iteration}} \\
\midrule
\multirow{2}{*}{1} & \textsc{MetaFORGE} & 14 & 6 & -6.050 & 16.67 & 98.18 & -65.00 & 245,178 & 17,512.71 \\
& \textsc{TextGrad} & 12 & 4 & -39.047 & -5.56 & 5.76 & 6.72 & 292,295 & 24,357.92 \\
\addlinespace[1.5pt]
\multirow{2}{*}{2} & \textsc{MetaFORGE} & 19 & 11 & -1.024 & 100.00 & 92.94 & 80.94 & 294,289 & 15,488.89 \\
& \textsc{TextGrad} & 9 & 1 & -68.156 & 0.00 & 0.00 & 0.00 & 201,220 & 22,357.78 \\
\addlinespace[1.5pt]
\multirow{2}{*}{3} & \textsc{MetaFORGE} & 20 & 13 & -1.004 & 100.00 & 100.00 & 95.99 & 365,050 & 18,252.50 \\
& \textsc{TextGrad} & 9 & 1 & -70.885 & 0.00 & 0.00 & 0.00 & 223,229 & 24,803.22 \\
\addlinespace[1.5pt]
\multirow{2}{*}{4} & \textsc{MetaFORGE} & 20 & 14 & -1.034 & 100.00 & 99.93 & 33.92 & 400,875 & 20,043.75 \\
& \textsc{TextGrad} & 9 & 1 & -2167.000 & 0.00 & 0.00 & 0.00 & 259,410 & 28,823.33 \\
\addlinespace[1.5pt]
\multirow{2}{*}{5} & \textsc{MetaFORGE} & 14 & 6 & -1.011 & 100.00 & 62.08 & 99.15 & 255,812 & 18,272.29 \\
& \textsc{TextGrad} & 15 & 7 & -13.075 & 25.00 & 99.99 & 24.44 & 412,586 & 27,505.73 \\
\midrule
\multirow{2}{*}{\makecell{\textbf{Mean}\\$\boldsymbol{\pm}$ \textbf{SD}}}
& \textsc{MetaFORGE} & $17.4 \pm 3.1$ & $10.0 \pm 3.8$ & $-2.024 \pm 2.251$ & $83.33 \pm 37.27$ & $90.63 \pm 16.21$ & $49.00 \pm 68.86$ & $312,241 \pm 68,280$ & $17,914.03 \pm 1,644.76$ \\
& \textsc{TextGrad} & $10.8 \pm 2.7$ & $2.8 \pm 2.7$ & $-471.633 \pm 948.033$ & $3.89 \pm 12.04$ & $21.15 \pm 44.15$ & $6.23 \pm 10.59$ & $277,748 \pm 83,003$ & $25,569.60 \pm 2,583.93$ \\
\bottomrule
\end{tabular}
\end{adjustbox}
\caption{Per-trial distributed PCA results for the large workload.}
\label{tab:pca_large_trial_summary}
\end{table*}

\begin{table*}[h]
\centering

\footnotesize
\setlength{\tabcolsep}{4pt}
\begin{tabular}{llcccccc}
\toprule
Held-Out Task & Skill & Compiler Warnings & Static Warnings & Runtime (s) & Imbalance & Reward & Wilcoxon $p$ \\
\midrule
Calculate $\pi$ & Initial     & 1.000 &6& 0.102 & 0.0441 & -7.27\\ 
                & Transferred & 1.000 &0& 0.090 & 0.0375 & -1.24 & 0.0039\\
\addlinespace
Matrix Multiplication & Initial & 1.000 & 21 & 30.133 & 0.1500 & -82.49\\ 
                & Transferred & 1.000 & 9 & 24.705 & 0.0500 & -59.49 & 0.0078\\
\addlinespace
Mandelbrot Sets & Initial      &  1.000 & 22 & 8.330 & 0.2000 & -39.96\\
               & Transferred  &  1.000 &  5 & 6.248 & 0.0800 & -18.62 & 0.0039\\
\bottomrule
\end{tabular}
\caption{Zero-update results averaged across trials. All generated programs passed task-specific correctness testing.}
\label{tab:zero_update_results}
\end{table*}

\subsection{Run-Level Token Consumption}
\label{appendix:run_level_token_cost}

Table~\ref{tab:run_level_token_cost} reports end-to-end token consumption for
each completed optimization run. Unlike the per-iteration token values reported
in the main results, these totals account for differences in refinement-trajectory
length and therefore provide a direct measure of the observed run-level inference
cost of each complete method. For every task, workload, and method, the table
reports the mean and sample standard deviation across five independent runs.
Prompt tokens contain the recorded input-token total, processing tokens contain
the recorded model-processing/output total, and total tokens are their sum.
Iterations report the number of completed outer-loop iterations, while
tokens per iteration retain the corresponding run-normalized measure.
Because \textsc{MetaFORGE} and the unmodified \textsc{TextGrad} implementation
use different native refinement pipelines and numbers of model calls per
iteration, the run-level totals should be interpreted as end-to-end method costs
rather than as a model-call-matched ablation.

\begin{table*}[h]
\centering

\scriptsize
\setlength{\tabcolsep}{2.5pt}
\begin{adjustbox}{max width=\textwidth}
\begin{tabular}{llcrrrrr}
\toprule
\textbf{Task} & \textbf{Workload} & \textbf{Method} & \makecell{\textbf{Iterations}\\\textbf{per run}} & \makecell{\textbf{Prompt}\\\textbf{tokens}} & \makecell{\textbf{Processing}\\\textbf{tokens}} & \makecell{\textbf{Total}\\\textbf{tokens}} & \makecell{\textbf{Tokens per}\\\textbf{iteration}} \\
\midrule
\multirow{6}{*}{Calculate $\pi$} & \multirow{2}{*}{small} & \textsc{MetaFORGE} & $15.4 \pm 4.9$ & $68,165 \pm 32,201$ & $66,986 \pm 26,256$ & $135,151 \pm 58,073$ & $8,548.65 \pm 1,184.41$ \\
 &  & \textsc{TextGrad} & $10.4 \pm 1.8$ & $89,999 \pm 18,527$ & $26,428 \pm 6,563$ & $116,427 \pm 24,729$ & $11,146.33 \pm 691.85$ \\
\addlinespace[1pt]
 & \multirow{2}{*}{medium} & \textsc{MetaFORGE} & $14.2 \pm 4.9$ & $61,614 \pm 34,457$ & $61,615 \pm 23,171$ & $123,229 \pm 57,588$ & $8,391.84 \pm 1,058.72$ \\
 &  & \textsc{TextGrad} & $12.4 \pm 3.6$ & $115,813 \pm 32,393$ & $33,647 \pm 9,239$ & $149,459 \pm 40,715$ & $12,154.55 \pm 660.07$ \\
\addlinespace[1pt]
 & \multirow{2}{*}{large} & \textsc{MetaFORGE} & $11.8 \pm 1.5$ & $44,926 \pm 10,514$ & $47,447 \pm 10,223$ & $92,374 \pm 20,290$ & $7,756.51 \pm 825.41$ \\
 &  & \textsc{TextGrad} & $9.6 \pm 5.8$ & $108,045 \pm 78,573$ & $31,601 \pm 24,903$ & $139,646 \pm 103,443$ & $13,926.57 \pm 1,590.87$ \\
\midrule
\multirow{6}{*}{Dot Product} & \multirow{2}{*}{small} & \textsc{MetaFORGE} & $13.0 \pm 4.6$ & $59,270 \pm 31,722$ & $65,774 \pm 21,640$ & $125,045 \pm 52,428$ & $9,506.86 \pm 784.85$ \\
 &  & \textsc{TextGrad} & $11.4 \pm 2.3$ & $118,035 \pm 26,672$ & $37,925 \pm 5,483$ & $155,960 \pm 31,972$ & $13,673.25 \pm 293.89$ \\
\addlinespace[1pt]
 & \multirow{2}{*}{medium} & \textsc{MetaFORGE} & $10.4 \pm 1.1$ & $38,420 \pm 8,424$ & $45,303 \pm 8,868$ & $83,723 \pm 15,079$ & $8,050.91 \pm 1,039.30$ \\
 &  & \textsc{TextGrad} & $11.0 \pm 3.1$ & $132,867 \pm 42,197$ & $40,152 \pm 8,844$ & $173,019 \pm 50,671$ & $15,714.14 \pm 837.25$ \\
\addlinespace[1pt]
 & \multirow{2}{*}{large} & \textsc{MetaFORGE} & $17.2 \pm 4.7$ & $80,225 \pm 28,251$ & $91,038 \pm 30,579$ & $171,262 \pm 57,765$ & $9,760.65 \pm 1,154.49$ \\
 &  & \textsc{TextGrad} & $9.8 \pm 1.8$ & $97,991 \pm 21,111$ & $36,640 \pm 8,021$ & $134,631 \pm 28,510$ & $13,698.97 \pm 773.50$ \\
\midrule
\multirow{6}{*}{Matrix multiplication} & \multirow{2}{*}{small} & \textsc{MetaFORGE} & $15.6 \pm 5.4$ & $98,021 \pm 47,199$ & $164,779 \pm 83,286$ & $262,800 \pm 130,405$ & $15,997.76 \pm 3,369.63$ \\
 &  & \textsc{TextGrad} & $14.2 \pm 4.7$ & $177,462 \pm 78,398$ & $87,150 \pm 33,408$ & $264,612 \pm 111,007$ & $18,127.86 \pm 2,164.62$ \\
\addlinespace[1pt]
 & \multirow{2}{*}{medium} & \textsc{MetaFORGE} & $13.2 \pm 5.2$ & $83,015 \pm 44,263$ & $119,744 \pm 42,115$ & $202,759 \pm 86,187$ & $15,250.04 \pm 616.59$ \\
 &  & \textsc{TextGrad} & $11.8 \pm 4.7$ & $142,465 \pm 73,524$ & $77,486 \pm 56,264$ & $219,951 \pm 129,524$ & $17,809.33 \pm 2,650.03$ \\
\addlinespace[1pt]
 & \multirow{2}{*}{large} & \textsc{MetaFORGE} & $11.6 \pm 4.8$ & $72,736 \pm 48,289$ & $114,108 \pm 51,734$ & $186,844 \pm 99,713$ & $15,675.86 \pm 1,643.55$ \\
 &  & \textsc{TextGrad} & $16.8 \pm 1.1$ & $208,605 \pm 22,330$ & $96,596 \pm 10,820$ & $305,201 \pm 27,870$ & $18,157.94 \pm 929.26$ \\
\midrule
\multirow{6}{*}{Mandelbrot set} & \multirow{2}{*}{small} & \textsc{MetaFORGE} & $14.2 \pm 4.2$ & $94,302 \pm 43,642$ & $132,798 \pm 45,934$ & $227,100 \pm 88,734$ & $15,753.75 \pm 1,726.59$ \\
 &  & \textsc{TextGrad} & $11.4 \pm 6.5$ & $125,950 \pm 78,644$ & $71,266 \pm 53,146$ & $197,216 \pm 130,201$ & $16,932.97 \pm 2,928.93$ \\
\addlinespace[1pt]
 & \multirow{2}{*}{medium} & \textsc{MetaFORGE} & $14.4 \pm 2.3$ & $93,944 \pm 19,429$ & $148,908 \pm 22,976$ & $242,852 \pm 36,425$ & $16,954.61 \pm 1,870.72$ \\
 &  & \textsc{TextGrad} & $10.8 \pm 5.2$ & $127,125 \pm 78,074$ & $55,362 \pm 34,682$ & $182,487 \pm 112,631$ & $16,317.68 \pm 1,741.09$ \\
\addlinespace[1pt]
 & \multirow{2}{*}{large} & \textsc{MetaFORGE} & $19.4 \pm 1.3$ & $157,616 \pm 20,455$ & $223,055 \pm 27,376$ & $380,672 \pm 37,960$ & $19,597.72 \pm 944.72$ \\
 &  & \textsc{TextGrad} & $10.2 \pm 2.3$ & $106,788 \pm 28,969$ & $60,924 \pm 10,317$ & $167,712 \pm 31,132$ & $16,561.59 \pm 1,015.06$ \\
\midrule
\multirow{6}{*}{Heat-transfer PDE} & \multirow{2}{*}{small} & \textsc{MetaFORGE} & $18.8 \pm 2.7$ & $150,460 \pm 33,844$ & $258,627 \pm 35,237$ & $409,087 \pm 60,727$ & $21,779.34 \pm 1,203.64$ \\
 &  & \textsc{TextGrad} & $10.2 \pm 2.7$ & $125,426 \pm 50,359$ & $81,676 \pm 26,602$ & $207,102 \pm 70,179$ & $20,143.61 \pm 3,264.18$ \\
\addlinespace[1pt]
 & \multirow{2}{*}{medium} & \textsc{MetaFORGE} & $20.0 \pm 0.0$ & $161,722 \pm 12,341$ & $262,002 \pm 17,711$ & $423,723 \pm 20,131$ & $21,186.17 \pm 1,006.54$ \\
 &  & \textsc{TextGrad} & $15.8 \pm 4.8$ & $195,237 \pm 68,833$ & $116,887 \pm 42,802$ & $312,124 \pm 109,819$ & $19,544.75 \pm 1,709.91$ \\
\addlinespace[1pt]
 & \multirow{2}{*}{large} & \textsc{MetaFORGE} & $19.0 \pm 1.7$ & $150,887 \pm 31,678$ & $251,045 \pm 44,066$ & $401,932 \pm 71,846$ & $21,085.93 \pm 2,590.94$ \\
 &  & \textsc{TextGrad} & $16.4 \pm 3.3$ & $186,865 \pm 42,367$ & $115,760 \pm 21,548$ & $302,626 \pm 62,073$ & $18,437.68 \pm 258.95$ \\
\midrule
\multirow{6}{*}{Distributed PCA} & \multirow{2}{*}{small} & \textsc{MetaFORGE} & $19.8 \pm 0.4$ & $143,817 \pm 14,195$ & $200,942 \pm 19,032$ & $344,759 \pm 32,746$ & $17,397.56 \pm 1,407.10$ \\
 &  & \textsc{TextGrad} & $16.2 \pm 5.3$ & $281,317 \pm 100,264$ & $110,791 \pm 34,134$ & $392,108 \pm 132,016$ & $24,175.65 \pm 1,242.99$ \\
\addlinespace[1pt]
 & \multirow{2}{*}{medium} & \textsc{MetaFORGE} & $19.2 \pm 1.8$ & $148,068 \pm 20,098$ & $208,877 \pm 26,819$ & $356,945 \pm 39,458$ & $18,585.08 \pm 955.18$ \\
 &  & \textsc{TextGrad} & $12.6 \pm 4.4$ & $205,426 \pm 90,007$ & $90,275 \pm 32,855$ & $295,702 \pm 122,164$ & $23,150.41 \pm 1,502.31$ \\
\addlinespace[1pt]
 & \multirow{2}{*}{large} & \textsc{MetaFORGE} & $17.4 \pm 3.1$ & $127,305 \pm 37,114$ & $184,936 \pm 32,306$ & $312,241 \pm 68,280$ & $17,914.03 \pm 1,644.76$ \\
 &  & \textsc{TextGrad} & $10.8 \pm 2.7$ & $183,786 \pm 58,028$ & $93,962 \pm 27,931$ & $277,748 \pm 83,003$ & $25,569.60 \pm 2,583.93$ \\
\bottomrule
\end{tabular}
\end{adjustbox}
\caption{Run-level optimization cost across five independent trials per workload.
Values are mean $\pm$ sample standard deviation. Prompt and processing tokens
sum to total tokens. Lower token counts indicate lower measured token consumption.}
\label{tab:run_level_token_cost}
\end{table*}

These run-level totals complement the per-iteration results by exposing the
combined effects of native model-call count, early stopping, and trajectory
length. Accordingly, claims concerning inference efficiency should distinguish
between per-iteration token consumption and total tokens required to complete an
optimization run.

\subsection{Ablation Study---MetaFORGE with and without Context Retrieval}
\label{appendix:retrieval_ablation_all}

Retrieval can expose the skill-refinement process to previously useful
optimization knowledge, but it can also introduce irrelevant context or
additional prompt cost. We therefore ablated retrieval while leaving the
remaining MetaFORGE optimization procedure unchanged. The ablation covers all
six MPI tasks and all three workload sizes, yielding 18 task--workload
configurations with five independent trials per configuration.

\begin{table*}[t]
\centering
\scriptsize
\setlength{\tabcolsep}{3.0pt}
\renewcommand{\arraystretch}{1.08}
\begin{adjustbox}{max width=\textwidth}
\begin{tabular}{llrrrr|rrrr}
\toprule
& & \multicolumn{4}{c|}{\textbf{With Context}} & \multicolumn{4}{c}{\textbf{Without Context}} \\
\cmidrule(lr){3-6}\cmidrule(lr){7-10}
\textbf{Task} & \textbf{Workload} & \textbf{Static} & \textbf{Runtime} & \textbf{Imbalance} & \textbf{Tokens} & \textbf{Static} & \textbf{Runtime} & \textbf{Imbalance} & \textbf{Tokens} \\
\midrule
\multirow{3}{*}{$\pi$ estimation}
& small  & \textbf{100.00\%} & \textbf{20.77\%} & \textbf{97.49\%} & 8549 & \textbf{100.00\%} & 8.22\% & 91.52\% & \textbf{8085} \\
& medium & \textbf{100.00\%} & \textbf{15.60\%} & 93.00\% & \textbf{8392} & \textbf{100.00\%} & 9.38\% & \textbf{99.14\%} & 9547 \\
& large  & \textbf{100.00\%} & \textbf{12.98\%} & 98.75\% & \textbf{7757} & \textbf{100.00\%} & 8.22\% & \textbf{99.76\%} & 8949 \\
\addlinespace
\multirow{3}{*}{Dot Product}
& small  & \textbf{100.00\%} & \textbf{19.27\%} & \textbf{46.47\%} & 9506 & 73.22\% & 13.68\% & 40.89\% & \textbf{9030} \\
& medium & \textbf{100.00\%} & \textbf{21.65\%} & \textbf{45.67\%} & \textbf{8051} & 65.15\% & 15.80\% & 34.71\% & 7890 \\
& large  & \textbf{100.00\%} & \textbf{40.08\%} & \textbf{60.08\%} & \textbf{9760} & 71.86\% & 10.42\% & 19.82\% & 11712 \\
\addlinespace
\multirow{3}{*}{Matrix Multiplication}
& small  & \textbf{90.96\%} & \textbf{93.49\%} & \textbf{60.37\%} & \textbf{15997} & 54.58\% & 57.96\% & 47.09\% & 17179 \\
& medium & \textbf{29.91\%} & \textbf{71.95\%} & \textbf{63.41\%} & \textbf{15250} & 22.43\% & 48.93\% & 45.66\% & 16241 \\
& large  & \textbf{50.60\%} & \textbf{98.52\%} & \textbf{17.34\%} & \textbf{15675} & 36.44\% & 67.98\% & 13.35\% & 16850 \\
\addlinespace
\multirow{3}{*}{Mandelbrot Set}
& small  & \textbf{65.00\%} & \textbf{2.15\%} & \textbf{24.77\%} & 15753 & 44.20\% & 1.52\% & 21.79\% & \textbf{14964} \\
& medium & \textbf{73.47\%} & \textbf{14.31\%} & \textbf{30.20\%} & 16954 & 56.57\% & 10.44\% & 22.95\% & \textbf{16276} \\
& large  & \textbf{96.92\%} & \textbf{11.49\%} & \textbf{20.90\%} & \textbf{19597} & 71.74\% & 8.27\% & 17.14\% & 20380 \\
\addlinespace
\multirow{3}{*}{Heat Transfer}
& small  & \textbf{27.94\%} & \textbf{78.18\%} & \textbf{41.23\%} & 21779 & 20.12\% & 56.29\% & 33.81\% & \textbf{21125} \\
& medium & \textbf{87.78\%} & \textbf{96.80\%} & \textbf{51.40\%} & 21186 & 67.59\% & 74.53\% & 39.58\% & \textbf{20762} \\
& large  & \textbf{73.52\%} & \textbf{72.39\%} & \textbf{61.61\%} & \textbf{21085} & 55.14\% & 52.84\% & 48.67\% & 21295 \\
\addlinespace
\multirow{3}{*}{Distributed PCA}
& small  & \textbf{100.00\%} & \textbf{68.91\%} & \textbf{12.20\%} & 17397 & 68.00\% & 51.14\% & 10.01\% & \textbf{16962} \\
& medium & \textbf{100.00\%} & \textbf{99.45\%} & \textbf{44.67\%} & 18585 & 71.25\% & 80.55\% & 35.28\% & \textbf{18213} \\
& large  & \textbf{83.33\%} & \textbf{90.63\%} & \textbf{49.00\%} & \textbf{17914} & 61.83\% & 71.59\% & 39.69\% & 18593 \\
\bottomrule
\end{tabular}
\end{adjustbox}
\caption{Effect of retrieval context across all six MPI tasks and three workload
sizes. Values are means over five independent trials per task--workload
configuration. Higher values are better for static-analysis, runtime, and
load-imbalance improvement; lower values are better for per-iteration token
consumption. Bold indicates the better value within each pair; ties are bolded
for both conditions.}
\label{tab:context_ablation_all}
\end{table*}

Across the 18 configurations, retrieval increases the unweighted mean
static-analysis improvement from 63.34\% to 82.19\% ($+18.85$ percentage
points), runtime improvement from 35.99\% to 51.59\% ($+15.60$ points), and
load-imbalance improvement from 42.27\% to 51.03\% ($+8.76$ points). Runtime
improves in all 18 configurations, static quality improves in 15 with three
ties, and load balance improves in 16. Mean token consumption decreases from
15,225 to 14,955 tokens per iteration, a 1.8\% reduction, with retrieval using
fewer tokens in 9 of 18 configurations. Because these entries are workload-
level means rather than independent trial-level observations, these summaries
are descriptive rather than inferential tests.

\begin{figure*}[h]
    \centering
    \includegraphics[width=1.0\linewidth]{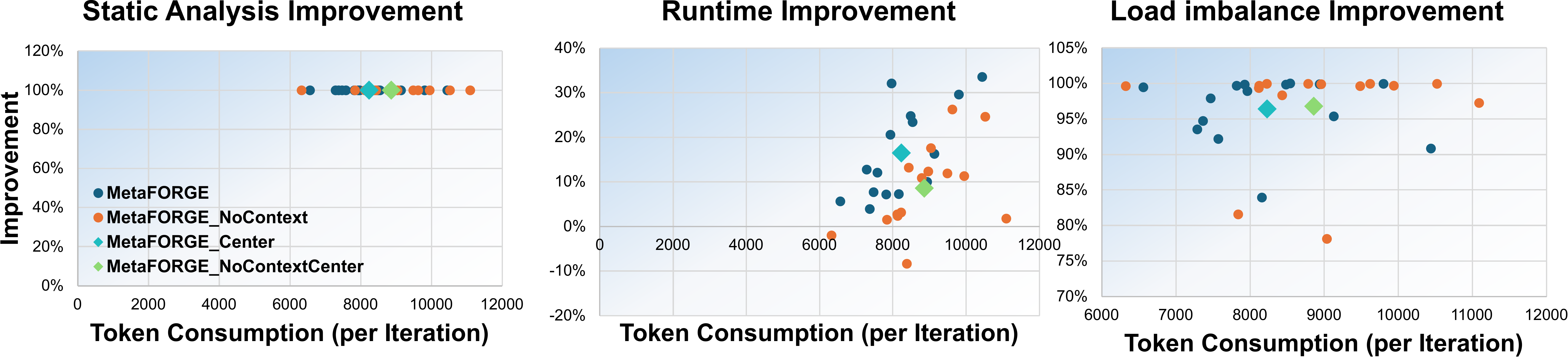}
    \caption{Comparison of MetaFORGE with and without retrieval context memory on the $\pi$ calculation task. Each point represents one optimization trial across all iterations, while diamonds indicate the mean value. The three subfigures report (left) static analysis improvement, (center) runtime improvement, and (right) load imbalance improvement. Incorporating retrieval context consistently improves runtime optimization while maintaining comparable static analysis and load balancing performance.}
    \label{fig:calculatepi_ablation}
\end{figure*}

The $\pi$ task illustrates the task dependence of the effect: both conditions
reach 100\% static-analysis improvement, while retrieval increases task-average
runtime improvement from 8.61\% to 16.45\%. Figure~\ref{fig:calculatepi_ablation}
visualizes this representative case; Table~\ref{tab:context_ablation_all}
shows that the runtime benefit extends across the complete benchmark suite.

\begin{table*}[h]
\centering

\footnotesize
\setlength{\tabcolsep}{8pt}
\renewcommand{\arraystretch}{1.05}

\begin{adjustbox}{width=\textwidth}
\begin{tabular}{llccccc}
\toprule
\textbf{Held-Out Task}
& \textbf{Cross-Task Memory}
& \textbf{Warnings $\downarrow$}
& \textbf{Runtime (s) $\downarrow$}
& \textbf{Load Imb. $\downarrow$}
& \textbf{Reward $\uparrow$}
& \textbf{Wilcoxon $p$} \\
\midrule

\multirow{2}{*}{$\pi$}
& Without
& 6
& 0.089
& 0.0441
& -7.24
& \multirow{2}{*}{$\mathbf{0.0039}$} \\
& With
& \textbf{0}
& \textbf{0.079}
& \textbf{0.0375}
& \textbf{-1.21}
& \\

\addlinespace

\multirow{2}{*}{Matrix Multiplication}
& Without
& 21
& 59.643
& 0.15
& -141.73
& \multirow{2}{*}{$\mathbf{0.0039}$} \\
& With
& \textbf{9}
& \textbf{48.167}
& \textbf{0.05}
& \textbf{-106.63}
& \\

\addlinespace

\multirow{2}{*}{Mandelbrot Set}
& Without
& 22
& 7.487
& 0.20
& -38.27
& \multirow{2}{*}{$\mathbf{0.0039}$} \\
& With
& \textbf{5}
& \textbf{5.615}
& \textbf{0.08}
& \textbf{-17.35}
& \\

\bottomrule
\end{tabular}
\end{adjustbox}
\caption{
Held-out cross-task adaptation with and without retrieval from memory
learned exclusively from the two non-held-out training tasks.
All other settings are identical.
Lower is better for warnings, runtime, and load imbalance;
higher is better for reward.
}

\label{tab:heldout}
\vspace{1mm}
\begin{minipage}{0.98\textwidth}
\footnotesize
\textit{Note.}
OpenAI discontinued access to \texttt{gpt-5.1-codex-max} after the
main MetaFORGE experiments but before the held-out cross-task transfer
study began. We therefore conducted the transfer experiment using
\texttt{gpt-5.3-codex}.
\end{minipage}
\end{table*}

\subsection{Held-Out Cross-Task Transfer}
\label{held_out_cross_task_transfer_study}

We used a leave-one-task-out design to determine whether
\textsc{MetaFORGE} learns optimization knowledge that transfers to tasks
excluded from skill and memory construction. The transfer pool contained
Calculate~$\pi$, Matrix Multiplication, and Mandelbrot Sets. In each fold, one
task and all three of its workload sizes were held out, while the remaining
two tasks were processed sequentially during Phase~1. Training began from the
same initial skill specification and an empty cross-task memory store.
Optimization episodes accumulated in the same memory throughout the fixed
training curriculum. The held-out task did not appear in any Phase~1 task
specification, skill update, or memory record.

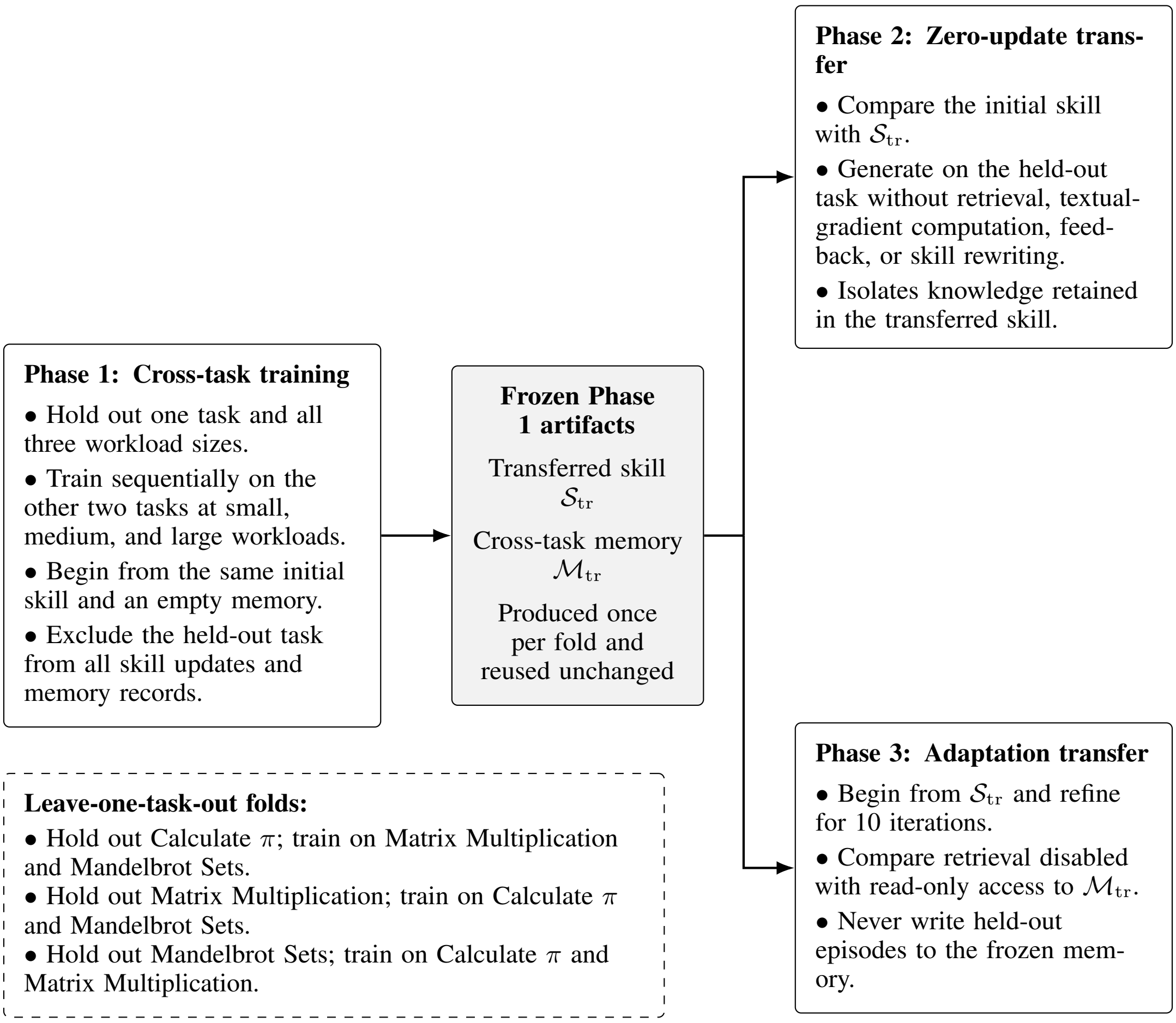
\begin{figure*}[t]
\centering
\begin{tikzpicture}[
    font=\footnotesize,
    protocol/.style={
        draw,
        rounded corners=2pt,
        align=left,
        text width=0.255\textwidth,
        inner sep=7pt,
        minimum height=3.0cm
    },
    artifact/.style={
        draw,
        rounded corners=2pt,
        fill=gray!10,
        align=center,
        text width=0.16\textwidth,
        inner sep=7pt,
        minimum height=2.65cm
    },
    evaluation/.style={
        draw,
        rounded corners=2pt,
        align=left,
        text width=0.255\textwidth,
        inner sep=7pt,
        minimum height=2.55cm
    },
    folds/.style={
        draw,
        dashed,
        rounded corners=2pt,
        align=left,
        text width=0.47\textwidth,
        inner sep=7pt
    },
    flow/.style={
        -{Latex[length=2.2mm]},
        thick
    }
]

\node[protocol] (phase1) {
    \textbf{Phase 1: Cross-task training}\\[4pt]
    $\bullet$ Hold out one task and all three workload sizes.\\[2pt]
    $\bullet$ Train sequentially on the other two tasks at small,
    medium, and large workloads.\\[2pt]
    $\bullet$ Begin from the same initial skill and an empty memory.\\[2pt]
    $\bullet$ Exclude the held-out task from all skill updates and
    memory records.
};

\node[artifact, right=0.85cm of phase1] (frozen) {
    \textbf{Frozen Phase 1 artifacts}\\[5pt]
    Transferred skill\\
    $\mathcal{S}_{\mathrm{tr}}$\\[5pt]
    Cross-task memory\\
    $\mathcal{M}_{\mathrm{tr}}$\\[5pt]
    Produced once per fold and reused unchanged
};

\node[
    evaluation,
    above right=0.20cm and 1.10cm of frozen
] (phase2) {
    \textbf{Phase 2: Zero-update transfer}\\[4pt]
    $\bullet$ Compare the initial skill with
    $\mathcal{S}_{\mathrm{tr}}$.\\[2pt]
    $\bullet$ Generate on the held-out task without retrieval,
    textual-gradient computation, feedback, or skill rewriting.\\[2pt]
    $\bullet$ Isolates knowledge retained in the transferred skill.
};

\node[
    evaluation,
    below right=0.20cm and 1.10cm of frozen
] (phase3) {
    \textbf{Phase 3: Adaptation transfer}\\[4pt]
    $\bullet$ Begin from $\mathcal{S}_{\mathrm{tr}}$ and refine for
    10 iterations.\\[2pt]
    $\bullet$ Compare retrieval disabled with read-only access to
    $\mathcal{M}_{\mathrm{tr}}$.\\[2pt]
    $\bullet$ Never write held-out episodes to the frozen memory.
};

\draw[flow] (phase1.east) -- (frozen.west);

\path (frozen.east) ++(0.48cm,0) coordinate (split);
\draw[thick] (frozen.east) -- (split);
\draw[flow] (split) |- (phase2.west);
\draw[flow] (split) |- (phase3.west);

\node[
    folds,
    anchor=north west
] (foldlist) at ([yshift=-0.55cm]phase1.south west) {
    \textbf{Leave-one-task-out folds:}\\[2pt]
    $\bullet$ Hold out Calculate~$\pi$; train on Matrix Multiplication
    and Mandelbrot Sets.\\
    $\bullet$ Hold out Matrix Multiplication; train on Calculate~$\pi$
    and Mandelbrot Sets.\\
    $\bullet$ Hold out Mandelbrot Sets; train on Calculate~$\pi$ and
    Matrix Multiplication.
};

\end{tikzpicture}

\caption{\textbf{Held-out cross-task transfer protocol.}
For each leave-one-task-out fold, \textsc{MetaFORGE} is trained sequentially
on the two non-held-out tasks across all workload sizes. The resulting
transferred skill and cross-task memory are frozen before evaluation.
Zero-update transfer isolates knowledge retained in the transferred skill,
whereas adaptation transfer evaluates refinement with and without read-only
access to the frozen cross-task memory. The held-out task is excluded from
all Phase~1 skill updates and memory construction.}
\label{fig:heldout_transfer_protocol}
\end{figure*}

Phase~1 produced one transferred skill and one populated cross-task memory for
each fold. The transferred skill was the best skill retained after the final
training stage. Phase~1 was executed once per fold rather than once per
evaluation trial. Upon completion, the transferred skill and memory were
frozen and reused as the common starting point for all held-out evaluation
trials. Figure~\ref{fig:heldout_transfer_protocol} summarizes the training,
freezing, and evaluation procedure.

We evaluated two complementary forms of transfer. In Phase~2, the
\emph{zero-update} experiment isolated knowledge retained in the transferred
skill by comparing it with the initial skill without retrieval, textual-gradient
computation, feedback, or rewriting. In Phase~3, the
\emph{adaptation-transfer} experiment refined the transferred skill for a
fixed budget of 10 iterations, either with retrieval disabled or with
read-only access to the frozen Phase~1 memory. The memory remained unchanged
during held-out evaluation, preventing held-out episodes from contaminating
subsequent iterations or trials.

For each held-out task, we conducted three stochastic trials at each of the three workload sizes, yielding nine paired observations per task and 27 observations across the complete experiment. The Phase~1 transferred skill and memory were trained once for each fold and reused across its trials. Thus, the trials represent independent stochastic code-generation and rewriting trajectories from a shared training outcome rather than replications based upon independently retrained skills and memories.   Furthermore, the per-task analysis pools observations across three workload sizes, despite differences in their computational difficulty. The resulting statistical evidence therefore should be interpreted  as conditional on the shared Phase~1 training outcome and the pooled workload design. 

Each paired observation compared the retrieval-enabled and retrieval-disabled run for the same held-out task, trial number, and workload size. Both members of a pair therefore shared the same task configuration and differed only in whether cross-task memory retrieval was available. Statistical testing was performed exclusively on the best reward obtained during each run. The held-out study used the same scalar reward definition and weights as Eq.~\eqref{eq:reward}. The maximum reward achieved across the 10-iteration adaptation trajectory constituted one run-level observation. Individual iterations were not treated as independent samples because successive skill rewrites were sequentially dependent. This run-level aggregation avoids pseudo-replication, although an unusually strong early iteration may substantially influence the best reward recorded for a run.

We compared paired best rewards using exact, two-sided Wilcoxon signed-rank tests. Runtime and load-imbalance measurements were included only for programs that passed correctness testing; these component metrics were reported descriptively and were not tested separately for statistical significance.

As shown in Table~\ref{tab:heldout}, optimization knowledge learned during Phase~1 transferred effectively to previously unseen tasks, and frozen cross-task memory further improved adaptation outcomes. For Calculate~$\pi$, retrieval eliminated the six reported static-analysis warnings, reduced runtime from (0.089) to (0.079) seconds, and reduced load imbalance from (0.0441) to (0.0375). For Matrix Multiplication, retrieval reduced static-analysis warnings from 21 to 9, runtime from (59.643) to (48.167) seconds, and load imbalance from (0.15) to (0.05). The largest static-analysis reduction occurred for Mandelbrot Sets, for which retrieval reduced warnings from 22 to 5, runtime from (7.487) to (5.615) seconds, and load imbalance from (0.20) to (0.08). Compiler diagnostics were unchanged, indicating that the reward improvements primarily resulted from better static quality, runtime performance, and workload balance. Correspondingly, the best reward became substantially less negative for every held-out task. The paired reward differences were statistically significant for each task under exact two-sided Wilcoxon signed-rank tests ($p = 0.0039$ for all three tasks), providing evidence that optimization knowledge retrieved from other tasks improved held-out adaptation under the evaluated conditions.

OpenAI discontinued access to \texttt{gpt-5.1-codex-max} after the primary \textsc{MetaFORGE} experiments but before the held-out cross-task transfer study began. Consequently, the held-out experiment was conducted using \texttt{gpt-5.3-codex} \cite{openai2026gpt53codex}.

\subsection{Zero-Update}
\label{zero_update_study}
We conducted a zero-update experiment to isolate cross-task skill transfer from \textsc{MetaFORGE}'s iterative adaptation process. For each held-out task, MPI C programs were generated using either the initial skill or the transferred skill, without textual-gradient rewriting, retrieval, feedback, or subsequent skill updates. This setting therefore measures whether knowledge acquired from other tasks improves generation under fixed-skill, zero-shot conditions.

As shown in Table~\ref{tab:zero_update_results}, the transferred skill produced better rewards for all three held-out tasks (p $\leq 0.0078$), while all generated programs passed correctness testing. Compiler warnings remained unchanged, indicating that the gains resulted from reductions in static-analysis warnings, runtime, and load imbalance. For Calculate~$\pi$, the transferred skill eliminated all six static-analysis warnings, reduced runtime from (0.102) to (0.090) seconds, and reduced imbalance from (0.0441) to (0.0375). For Matrix Multiplication, it reduced static warnings from 21 to 9, runtime from (30.133) to (24.705) seconds, and imbalance from (0.15) to (0.05). For Mandelbrot Sets, static warnings decreased from 22 to 5, runtime from (8.330) to (6.248) seconds, and imbalance from (0.20) to (0.08).

The runtime reductions were approximately (11.8\%), (18.0\%), and (25.0\%) for Calculate~$\pi$, Matrix Multiplication, and Mandelbrot Sets, respectively. Correspondingly, the composite reward improved from (-7.27) to (-1.24), from (-82.49) to (-59.49), and from (-39.96) to (-18.62). These results provide evidence under the evaluated three-task setting that the transferred skill retains useful cross-task optimization knowledge and improves held-out code generation even without retrieval or task-specific adaptation.

\subsection{Ablation Study---MetaFORGE with Claude vs. Codex}

Table~\ref{tab:model_ablation_pca} reports the cross-model ablation comparing MetaFORGE with \texttt{gpt-5.1-codex-max} and Claude Sonnet~5 (\texttt{claude-sonnet-5}) on the distributed PCA task. Both models benefit from MetaFORGE's reward-guided skill-refinement process, but they exhibit different optimization profiles. Codex provides stronger runtime optimization and lower token consumption, whereas Claude achieves stronger load-balance improvement and slightly better static-analysis improvement on the largest workload.

\begin{table*}[h]
\centering

\begin{adjustbox}{width=2.0\columnwidth}
\scriptsize
\setlength{\tabcolsep}{2.5pt}
\renewcommand{\arraystretch}{0.92}

\begin{tabular}{ll|cccc|cccc}
\Xhline{1pt}
\multirow{2}{*}{\textbf{Task}}
& \multirow{2}{*}{\textbf{Workload}}
& \multicolumn{4}{c|}{\textbf{MetaFORGE (Codex)}}
& \multicolumn{4}{c}{\textbf{MetaFORGE (Claude)}} \\
\cline{3-10}
&
& \textbf{Static}
& \textbf{Runtime}
& \textbf{Imbalance}
& \textbf{Tokens}
& \textbf{Static}
& \textbf{Runtime}
& \textbf{Imbalance}
& \textbf{Tokens} \\
\Xhline{0.6pt}

\multirow{3}{*}{Distributed PCA}
& small
& \textbf{100.00\%} & \textbf{68.91\%} & 12.20\% & \textbf{17397}
& \textbf{100.00\%} & 29.25\% & \textbf{60.54\%} & 38352 \\

& medium
& \textbf{100.00\%} & \textbf{99.45\%} & 44.67\% & \textbf{18585}
& \textbf{100.00\%} & 63.53\% & \textbf{85.50\%} & 40210 \\

& large
& 83.33\% & \textbf{90.63\%} & 49.00\% & \textbf{17914}
& \textbf{100.00\%} & 73.64\% & \textbf{75.25\%} & 42635 \\

\Xhline{1pt}
\end{tabular}
\end{adjustbox}
\caption{Cross-model ablation on distributed PCA over five trials per workload. Values are mean improvements and per-iteration token consumption. \textbf{Bold} indicates the better result.}

\label{tab:model_ablation_pca}
\end{table*}

For code quality, both models achieve 100.00\% static-analysis improvement on the small and medium PCA workloads. On the large workload, Claude maintains 100.00\% static improvement, while Codex reaches 83.33\%. This result suggests that Claude is more consistent at eliminating static-analysis warnings for the most complex PCA configuration. However, the static-analysis advantage does not translate into better runtime performance.

Runtime improvement is higher when MetaFORGE uses Codex on all three
Distributed PCA workload sizes. Codex improves runtime by 68.91\%, 99.45\%,
and 90.63\% on the small, medium, and large workloads, respectively, compared
with 29.25\%, 63.53\%, and 73.64\% for Claude. Thus, within this Distributed
PCA experiment, Codex produces the larger runtime improvement at every
evaluated workload size.

Claude produces larger load-balance improvements on all three PCA workload
sizes. MetaFORGE with Claude improves imbalance by 60.54\%, 85.50\%, and
75.25\%, compared with 12.20\%, 44.67\%, and 49.00\% for Codex. The observed
tradeoff is therefore task specific: Codex produces larger runtime gains,
whereas Claude produces larger rank-balance gains.

Per-iteration token consumption is also lower for Codex. Across the three
workload sizes, Codex requires 17,397--18,585 tokens per iteration, whereas
Claude requires 38,352--42,635. On this task, Codex therefore provides the
stronger runtime/token tradeoff, while Claude provides stronger imbalance
reduction and static-analysis improvement. Overall,
Table~\ref{tab:model_ablation_pca} shows optimization gains with both evaluated
model families on Distributed PCA. This single-task cross-provider ablation
does not establish model independence.

\subsection{Refinement-Model Strength Ablation}
\label{appendix:refiner_strength_ablation}

The primary comparison uses different native refinement models: MetaFORGE uses
a stronger reasoning model for critique and rewriting, whereas the published
TextGrad configuration uses \texttt{gpt-4o}. To isolate this confound, we
performed a model-matched ablation using the same \texttt{gpt-5.3-codex} code
generator, seed skill, evaluator, reward function, stopping rule, and frozen
read-only memory snapshot in all conditions. \textsc{MetaFORGE}-Strong uses
\texttt{gpt-5.3-codex} for refinement, \textsc{MetaFORGE}-Weak uses
\texttt{gpt-4o}, and \textsc{TextGrad}-Native retains its published
\texttt{gpt-4o} refinement procedure. Each task--method combination was
evaluated in five independent trials, yielding 90 runs; all selected programs
passed correctness testing.

\begin{table*}[t]
\centering

\scriptsize
\setlength{\tabcolsep}{3.2pt}
\renewcommand{\arraystretch}{1.08}
\resizebox{\textwidth}{!}{%
\begin{tabular}{llrrrrr}
\toprule
\textbf{Task} & \textbf{Method} & \textbf{Static warnings} & \textbf{Runtime (s)} & \textbf{Load imbalance} & \textbf{Best reward $r^*$} & \textbf{Total tokens} \\
\midrule
$\pi$ & MetaFORGE-Strong & $2.40 \pm 2.30$ & $0.0684 \pm 0.0083$ & $(3.54 \pm 5.26)\times10^{-3}$ & $-2.542 \pm 2.317$ & $362{,}964 \pm 47{,}371$ \\
$\pi$ & MetaFORGE-Weak & $3.00 \pm 1.58$ & $0.0839 \pm 0.0053$ & $(4.55 \pm 5.06)\times10^{-2}$ & $-3.236 \pm 1.549$ & $80{,}272 \pm 15{,}303$ \\
$\pi$ & TextGrad-Native & $6.80 \pm 1.30$ & $0.0900 \pm 0.0232$ & $(2.89 \pm 5.47)\times10^{-2}$ & $-7.023 \pm 1.369$ & $246{,}120 \pm 95{,}703$ \\
\midrule
Dot Product & MetaFORGE-Strong & $3.60 \pm 2.30$ & $0.0017 \pm 0.0010$ & $8.39\times10^{-2} \pm 1.67\times10^{-1}$ & $-3.729 \pm 2.349$ & $279{,}069 \pm 58{,}870$ \\
Dot Product & MetaFORGE-Weak & $3.40 \pm 0.89$ & $0.0200 \pm 0.0099$ & $4.91\times10^{-2} \pm 7.20\times10^{-2}$ & $-3.514 \pm 0.879$ & $99{,}469 \pm 34{,}758$ \\
Dot Product & TextGrad-Native & $8.00 \pm 0.71$ & $0.0293 \pm 0.0074$ & $4.96\times10^{-2} \pm 1.40\times10^{-2}$ & $-8.133 \pm 0.705$ & $166{,}654 \pm 21{,}436$ \\
\midrule
Matrix Multiplication & MetaFORGE-Strong & $9.40 \pm 11.44$ & $8.5931 \pm 3.1626$ & $7.25\times10^{-3} \pm 7.77\times10^{-3}$ & $-26.597 \pm 15.253$ & $454{,}206 \pm 158{,}701$ \\
Matrix Multiplication & MetaFORGE-Weak & $9.80 \pm 7.40$ & $8.3543 \pm 1.6077$ & $6.30\times10^{-3} \pm 1.02\times10^{-2}$ & $-26.518 \pm 7.449$ & $185{,}205 \pm 68{,}023$ \\
Matrix Multiplication & TextGrad-Native & $21.60 \pm 7.57$ & $8.7189 \pm 1.2054$ & $5.14\times10^{-3} \pm 3.49\times10^{-3}$ & $-39.045 \pm 9.585$ & $220{,}272 \pm 58{,}010$ \\
\midrule
Mandelbrot & MetaFORGE-Strong & $11.20 \pm 7.63$ & $3.2669 \pm 1.3092$ & $1.36\times10^{-2} \pm 1.46\times10^{-2}$ & $-17.754 \pm 6.876$ & $340{,}144 \pm 178{,}604$ \\
Mandelbrot & MetaFORGE-Weak & $13.60 \pm 5.55$ & $4.1869 \pm 1.4921$ & $1.19\times10^{-2} \pm 9.88\times10^{-3}$ & $-21.992 \pm 7.772$ & $187{,}872 \pm 109{,}963$ \\
Mandelbrot & TextGrad-Native & $16.00 \pm 6.20$ & $5.6415 \pm 0.1285$ & $1.14\times10^{-2} \pm 1.13\times10^{-3}$ & $-27.300 \pm 6.084$ & $328{,}812 \pm 141{,}164$ \\
\midrule
Heat Transfer & MetaFORGE-Strong & $19.00 \pm 9.11$ & $47.5152 \pm 1.4676$ & $6.32\times10^{-3} \pm 8.14\times10^{-3}$ & $-114.240 \pm 7.066$ & $374{,}454 \pm 189{,}099$ \\
Heat Transfer & MetaFORGE-Weak & $19.60 \pm 16.41$ & $47.7822 \pm 27.3321$ & $0.763 \pm 1.68$ & $-116.309 \pm 37.961$ & $237{,}836 \pm 66{,}739$ \\
Heat Transfer & TextGrad-Native & $23.40 \pm 3.05$ & $49.2273 \pm 1.5946$ & $1.25\times10^{-2} \pm 6.34\times10^{-3}$ & $-122.073 \pm 4.769$ & $319{,}632 \pm 110{,}104$ \\
\midrule
Distributed PCA & MetaFORGE-Strong & $10.20 \pm 4.55$ & $50.4811 \pm 6.4115$ & $1.93\times10^{-6} \pm 9.85\times10^{-7}$ & $-111.162 \pm 16.976$ & $461{,}760 \pm 159{,}016$ \\
Distributed PCA & MetaFORGE-Weak & $12.20 \pm 3.11$ & $72.3796 \pm 13.5163$ & $1.53\times10^{-4} \pm 1.70\times10^{-4}$ & $-157.159 \pm 28.533$ & $140{,}183 \pm 19{,}748$ \\
Distributed PCA & TextGrad-Native & $20.60 \pm 5.46$ & $77.7888 \pm 24.4503$ & $1.28\times10^{-4} \pm 1.37\times10^{-4}$ & $-176.378 \pm 47.177$ & $364{,}280 \pm 103{,}025$ \\
\bottomrule
\end{tabular}%
}

\caption{Refinement-model sensitivity across six MPI tasks. Values are means
$\pm$ sample standard deviations over five independent trials. Higher
(less negative) best reward $r^*$ is better; lower values are better for
static-analysis warnings, runtime, load imbalance, and total token consumption.}

\label{tab:refiner_ablation_results}
\end{table*}

Both MetaFORGE conditions achieve higher mean best reward than
\textsc{TextGrad}-Native on all six tasks. Ranking the 15 observations within
each task gives mean ranks of 5.30 for Strong, 7.30 for Weak, and 11.40 for
TextGrad-Native, where lower is better. Because trials are independent rather
than paired by index, inference uses exact task-stratified permutation tests.
The corresponding pairwise test statistics, including exact and Holm-adjusted
$p$-values, are reported in Table~\ref{tab:refiner_posthoc_main}.

Both MetaFORGE conditions retain statistically significant best-reward
advantages over TextGrad-Native after multiplicity correction. Strong versus
Weak favors the stronger refiner numerically but does not reach the corrected
$\alpha=0.05$ threshold. Token use shows the opposite tradeoff: mean total token
consumption across the 30 runs per method is 378,766 for Strong, 155,139 for
Weak, and 274,295 for TextGrad-Native. Thus, the model-matched Weak condition
uses 43.4\% fewer tokens than TextGrad-Native while retaining its significant
best-reward advantage. The ablation supports the conclusion that
MetaFORGE remains effective when refinement-model capability is reduced, however it
does not establish equivalence or model-independence.

\clearpage
\raggedbottom

\section{Memory Algorithms}
\label{appendix:memory_algorithms}

The following algorithms provide implementation details for context-augmented iteration, tuple extraction, episode compression, lesson distillation, and weighted retrieval. These details are summarized in the main paper in Section~\ref{subsection:long_term_context_memory_system}.

For tuple extraction, $\Pi(p)$ is the keyword set for parallel pattern $p$,
$\mathcal{W}$ is the task vocabulary, $S_{\mathrm{perf}}$ and
$S_{\mathrm{corr}}$ are the performance- and correctness-symptom sets,
$\Phi_s$ is the metric predicate for symptom $s$, and $K_s$ contains its
diagnostic keywords. During compression, $E$ denotes all recorded episodes,
$E^+$ the reward-improving subset, and $E^{\mathrm{sel}}$ the episodes used
to construct memory. The aggregated tags are $T_c$, the ordered textual
gradients are $G$, and $h$, $A$, and $\ell$ denote the distilled headline,
actions, and lesson. The resulting memory has reward improvement $\Delta r$
and identifier $\mu$. When $E^+$ is empty, all episodes are used so that the
memory remains defined.
For Algorithm~\ref{alg:distill_lesson_inline}, $L_{\min}$ is the minimum
candidate-line length, $A_{\max}$ is the maximum number of unique candidate
actions retained before summarization, and $A_{\mathrm{ret}}$ is the maximum
number of actions returned with the distilled lesson.
$\operatorname{UniqueFirst}$ removes duplicates while preserving first
occurrence order.

For Algorithm~\ref{alg:compress_and_Index}, $E$ denotes the episodes loaded
from the episode store, and $e.\mathrm{improved}$ is the Boolean flag recorded
for episode $e$ indicating whether that episode improved the best-observed
reward. Accordingly, $E^+$ denotes the reward-improving subset and
$E^{\mathrm{sel}}$ the episodes selected for memory construction. For each
tuple category $c$, $T_c$ is the union of values in the stored tuple field
$e.\mathbf{q}_c$ contributed by the selected episodes. The collection
$\{T_c\}_c$ forms the aggregated tag bundle written to memory. The ordered
list $G$ contains the episodes' textual gradients, while $h$, $A$, and $\ell$
denote the distilled headline, action list, and lesson. The value $\Delta r$
is the reward improvement over the first episode, and $\mu$ is the identifier
assigned to the resulting memory record.

\begin{algorithm}[H]     
\caption{Tuple Extraction}
\label{alg:extract}
\footnotesize
\begin{algorithmic}[1]
\Require Source $c$, task $\mathcal{T}$, metrics $\mathbf{m}$,
previous textual gradient $g_{\mathrm{prev}}$, diagnostics $\ell$

\Ensure Tuple bundle $\mathbf{q}$

\State $\qset{mpi}\gets
 \{x \mid x \text{ matches } \mpipat
 \text{ in } c\cup g_{\mathrm{prev}}\}$

\State $\qset{pat}\gets
 \{p \mid \exists kw\in\Pi(p): kw\in\oplow(c\cup g_{\mathrm{prev}})\}$

\State $\qset{task}\gets
 \{w\in\mathcal{W} \mid w\in\oplow(\mathcal{T})\}$

\State $\qset{perf}\gets
 \{s\in S_{\mathrm{perf}} \mid \Phi_s(\mathbf{m})\}$

\State Identify additional performance symptoms in $S_{\mathrm{perf}}$ that are not already in $\qset{perf}$ and whose keywords appear in $\oplow(g_{\mathrm{prev}})$.
\State Add the identified symptoms to $\qset{perf}$.

\State $\qset{corr}\gets
 \{s\in S_{\mathrm{corr}} \mid
   \exists kw\in K_s:
   kw\in\oplow(\ell\cup g_{\mathrm{prev}})\}$

\State $\mathbf{q}\gets
 \{\qset{mpi},\qset{pat},\qset{task},\qset{perf},\qset{corr}\}$

\State \Return $\mathbf{q}$

\end{algorithmic}
\end{algorithm}

\begin{algorithm}[H]
\caption{Compress and Index}
\label{alg:compress_and_Index}
\footnotesize
\begin{algorithmic}[1]
\Require Episode IDs $\mathbf{ids}=[\mathrm{id}_1,\ldots,\mathrm{id}_n]$
\Ensure Memory ID $\mu$; return $\bot$ if empty

\State Exit and return $\bot$ if $\mathbf{ids}$ is empty.

\State $E\gets[\mathcal{E}.\mathrm{load}(\mathrm{id}) \mid
\mathrm{id}\in\mathbf{ids}]$

\State $E^+\gets\{e\in E \mid e.\mathrm{improved}\}$

\State $E^{\mathrm{sel}}\gets
\begin{cases}
E^+, & E^+\neq\emptyset,\\
E,   & E^+=\emptyset
\end{cases}$

\State For each tuple category
$c\in\{\mathrm{mpi},\mathrm{pat},\mathrm{task},\mathrm{perf},
\mathrm{corr}\}$, set
$T_c\gets\bigcup_{e\in E^{\mathrm{sel}}}e.\mathbf{q}_c$

\State $G\gets
[e.g\mid e\in E^{\mathrm{sel}}]\mathbin{+\!\!+}
[e.g\mid e\in E\setminus E^{\mathrm{sel}}]$

\State $(h,A,\ell)\gets
\mathrm{DistillLesson}(G,E,E^{\mathrm{sel}})$
\Comment{Alg.~\ref{alg:distill_lesson_inline}}
\State $\Delta r\gets \max_{e\in E}e.r - E[0].r$

\State $\mu\gets\mathrm{``mem\_''}+\mathrm{timestamp}()$
\State $\mathcal{M}.\mathrm{write}
(\mu,\{\mu,\{T_c\}_c,h,A,\ell,\Delta r,|E^+|,|E|\})$

\State For each tuple category $c$ and value $v\in T_c$, update $\mathcal{I}[c][v]\gets\mathcal{I}[c][v]\cup\{\mu\}$

\State $\mathcal{I}.\mathrm{flush}()$
\State \Return $\mu$
\end{algorithmic}
\end{algorithm}

\begin{algorithm}[H]
\caption{Lesson Distillation}
\label{alg:distill_lesson_inline}
\footnotesize
\begin{algorithmic}[1]
\Require Gradients $G=[g_1,g_2,\ldots]$ with selected episodes
first; all episodes $E$; selected episodes $E^{\mathrm{sel}}$;
action limits $(A_{\max},A_{\mathrm{ret}})$; minimum candidate-line
length $L_{\min}$

\Ensure Headline $h$, actions $A$, lesson $\ell$

\State $A\gets[\,]$

\State $L\gets
\{\operatorname{strip}(\lambda)\mid
g\in G,\;\lambda\in\operatorname{lines}(g)\}$

\State $A\gets
\operatorname{UniqueFirst}_{A_{\max}}\!\left(
\{\lambda\in L\mid |\lambda|\geq L_{\min}\right.$

\Statex \hspace{\algorithmicindent}
$\left.\land\;\exists v\in\mathcal{V}:
v\in\oplow(\lambda)\}\right)$

\State $e^*\gets
\arg\max_{e\in E^{\mathrm{sel}}}e.r$

\State $h\gets
\operatorname{firstSubstantiveLine}(e^*.g)$

\If{$h=\varepsilon$}
  \State $h\gets
  \begin{cases}
    A[0], & A\neq[\,],\\
    \textit{(no lesson)}, & \text{otherwise}
  \end{cases}$
\EndIf

\State $e_b\gets\arg\max_{e\in E}e.r$

\State $\textit{hdr}\gets
\langle\text{best metrics of }e_b\rangle$

\State $\ell\gets
\textsc{SummarizeRecommendations}(\textit{hdr},A)$
\Comment{Prompt~\ref{distillandcompress}}

\State \Return $h,\;A[0{:}A_{\mathrm{ret}}],\;\ell$

\end{algorithmic}
\end{algorithm}

\begin{algorithm}[H]
\caption{Weighted Context Retrieval}
\label{alg:retrieve_inline}
\footnotesize
\begin{algorithmic}[1]
\Require Query tags $q$, retrieval budget $k$
\Ensure Ranked memories $\mathbf{R}$

\State $\mathcal{C}\gets \bigcup_{c,\,v\in q_c}\mathcal{I}[c][v]$ \Comment{tag-overlap candidates}
\State $\mathbf{R}\gets[\,]$

\State $\mathbf{R}\gets \{(m,\sigma)\mid \mu\in\mathcal{C},\; m=\mathcal{M}.\mathrm{load}(\mu),\; \sigma=\mathrm{WeightedScore}(q,m),\; \sigma>0\}$

\State \Return $\topk(\mathbf{R},k)$ sorted by descending $\sigma$
\Statex
\end{algorithmic}
\end{algorithm}


\section{Prompts Used in MetaFORGE}
\label{appendix:prompts}
\subsection{Code Generation Prompt}
\begin{tcolorbox}[compactprompt,label=prompt_code_generation]
\footnotesize
\# Skill Specification

\{Skill Text\}

\# Task

\{Task Text\}

Generate a complete MPI C program implementing the task per the skill specification.

\end{tcolorbox}

\subsection{Textual Gradient Prompt}

\begin{tcolorbox}[compactprompt,label=prompt_text_gradient_inline]
\footnotesize

\{Previous Skill Specification\}

\{Task Specification\}


\{Metric Summary: current and best values for correctness status, compiler warnings, static-analysis warnings, runtime, load imbalance, and reward\}

\[
\begin{aligned}
\max\quad & r \\
\text{s.t.}\quad
& r = -(\alpha_c n_c + \alpha_s n_s + \alpha_t t_{\mathrm{job}} \\
&\qquad\quad + \alpha_\lambda \lambda + \alpha_f f).
\end{aligned}
\]

\noindent
where $f=1$ if correctness testing fails and $f=0$ otherwise.

\medskip
\noindent
\{Compiler Warnings Text\}

\{Static Analysis Warnings\}

\{Generated Code\}

Produce the textual gradient for the skill specification above. Prioritize changes that restore correctness when $f=1$; otherwise, prioritize changes that reduce the largest weighted reward terms.
\end{tcolorbox}

\subsection{Skill Rewrite Prompt}

\begin{tcolorbox}[compactprompt,label=skillrewrite]
\footnotesize
\{Previous Skill Definition\}

\{Computed Textual Gradient\}

Rewrite the skill specification applying the gradient's guidance. 
Increment the 'version' field in the formatter by 1.
Output ONLY the updated skill Markdown.
\end{tcolorbox}

\subsection{Knowledge Distillation Prompt}

\begin{tcolorbox}[compactprompt,label=distillandcompress]
\footnotesize

The following action lines were extracted from MPI optimization gradients across \{n\} candidate lines.

Best result achieved in this run:

\{Best Candidate reward, compiler stats, static results, and runtime\}

Action Lines:

\{Action Lines\}

Produce a concise summary of actionable recommendations.

Rules:

- Preserve specific MPI function names, flags, and data types.

- Merge near-duplicates into a single, precise statement.

- Resolve contradictions in favor of actions from improving iterations.

- Remove vague or non-actionable lines.

- Output only the recommendations, one per line, with no bullets, numbering, or extra commentary.

\end{tcolorbox}
\clearpage
\balance
\section{Notation \& Tables}

\begin{center}
\begin{minipage}{\columnwidth}
\centering

\label{tab:notation}
\begin{adjustbox}{width=\columnwidth}
\begin{tabular}{@{}ll@{}}
\toprule
Symbol & Description \\
\midrule
$\mathcal{S}$ & Skill specification \\
$\mathcal{T}$ & Fixed task specification \\
$C_i$ & MPI C program at iteration $i$ \\
$n_c$ & Compiler-warning count \\
$n_s$ & Static-analysis-warning count \\
$t_{\mathrm{job}}$ & Mean externally measured whole-job runtime over 20 executions \\
$\boldsymbol{\tau}$ & Per-rank elapsed times in $\mathbb{R}^K$ used for load imbalance \\
$\lambda$ & Load imbalance, $(\max_k \tau_k-\min_k \tau_k)/\bar{\tau}$ \\
$r$ & Scalar reward; closer to zero is better \\
$r^*$ & Best observed reward \\
$\nabla_{\!\text{text}}$ & Textual gradient for skill rewriting \\
$p$ & Non-improvement counter \\
$P$ & No-improvement limit \\
$\rho$ & Rewriter backend \\
\bottomrule
\end{tabular}
\end{adjustbox}
\captionof{table}{MetaFORGE notation.}
\end{minipage}
\end{center}

\begin{center}
\begin{minipage}{\columnwidth}
\centering

\label{tab:math_notations_used}
\begin{adjustbox}{width=\columnwidth}
\begin{tabular}{@{}ll@{}}
\toprule
Symbol & Meaning \\
\midrule
$\mathbf{q}$ & Tuple bundle: $\qset{mpi},\qset{pat},\qset{task},\qset{perf},\qset{corr}$ \\
$q,m$ & Query tuple and stored memory entry \\
$\mathcal{E},\mathcal{M}$ & Episode and memory stores \\
$\mathcal{I}$ & Inverted index from tags to memory IDs \\
$\jac(A,B)$ & Jaccard similarity; $\jac(\emptyset,\emptyset)=0$ \\
$\Delta r_m$ & Reward improvement for memory $m$ \\
$\age(m)$ & Age of memory $m$ in days \\
$\Pi$ & MPI-pattern keyword dictionary \\
$\Phi_s$ & Predicate for performance symptom $s$ \\
$\mathcal{W}$ & MPI task vocabulary \\
$\mathcal{V}$ & Action-verb lexicon \\
$S_{\mathrm{perf}},S_{\mathrm{corr}}$ & Performance and correctness symptom sets \\
$K_s$ & Diagnostic keywords for symptom $s$ \\
\bottomrule
\end{tabular}
\end{adjustbox}
\captionof{table}{Context-retrieval notation.}
\end{minipage}
\end{center}

\begin{center}
\begin{minipage}{\columnwidth}
\centering

\label{tab:weights}
\renewcommand{\arraystretch}{0.92}
\begin{adjustbox}{width=\columnwidth}
\begin{tabular}{@{}lcp{3.25cm}@{}}
\toprule
Component & Weight & Score term \\
\midrule
MPI construct
& 0.25
& $\jac(q_{\mathrm{mpi}},m_{\mathrm{mpi}})$ \\

Parallel pattern
& 0.25
& $\jac(q_{\mathrm{pat}},m_{\mathrm{pat}})$ \\

Performance symptom
& 0.20
& $\jac(q_{\mathrm{perf}},m_{\mathrm{perf}})$ \\

Task keyword
& 0.15
& $\jac(q_{\mathrm{task}},m_{\mathrm{task}})$ \\

Reward confidence
& 0.10
& $\clamp(\Delta r_m/2,0,1)$ \\

Recency
& 0.05
& $e^{-\age(m)/30}$ \\
\bottomrule
\end{tabular}
\end{adjustbox}
\captionof{table}{Context-retrieval scoring weights.}
\end{minipage}
\end{center}

\end{document}